\documentclass{article}

\usepackage[final]{neurips_2025}

\usepackage[utf8]{inputenc} 
\usepackage[T1]{fontenc}    
\usepackage[pagebackref=true,breaklinks=true,colorlinks,citecolor=blue,bookmarks=false]{hyperref}
\usepackage{url}            
\usepackage{booktabs}       
\usepackage{amsfonts}       
\usepackage{nicefrac}       
\usepackage{microtype}      
\usepackage{xcolor}         

\usepackage{amsmath}
\usepackage{multirow}
\usepackage{graphicx} 
\usepackage{float}
\usepackage{natbib}
\usepackage{wrapfig}
\usepackage{tikz}
\usepackage{subcaption} 

\usepackage{multirow}
\usepackage[normalem]{ulem}
\useunder{\uline}{\ul}{}
\usepackage{longtable}
\usepackage{booktabs}
\usepackage{array}
\usepackage{url}
\usepackage{tabularx}
\usepackage{ragged2e}
\usepackage{hyperref}
\usepackage{multirow, makecell}
\usepackage{colortbl}

\usepackage{placeins}
\usepackage{needspace}

\usepackage{authblk}

\newcommand{\eg}{\textit{e.g.}~}
\newcommand{\ie}{\textit{i.e.}~}

\newcommand{\circledgray}[1]{%
  \tikz[baseline=(char.base)]{%
    \node[shape=circle,fill=gray!40,inner sep=0.5pt,minimum size=6pt] (char) {\textcolor{black}{#1}};
  }%
}

\newcommand{\circledblack}[1]{%
  \tikz[baseline=(char.base)]{%
    \node[shape=circle,fill=black,inner sep=0.5pt,minimum size=6pt] (char) {\textcolor{white}{#1}};
  }%
}

\definecolor{lightgrey}{RGB}{224, 224, 224}
\newcommand{\lightgrey}{\cellcolor{lightgrey}}

\title{GreenHyperSpectra: A multi-source hyperspectral dataset for global vegetation trait prediction}

\author[   1,2,3]{Eya Cherif\thanks{Core Team}$^{*}$\thanks{Corresponding author: eya.cherif@uni-leipzig.de}$^{\dagger}$}
\author[3,4]{Arthur Ouaknine$^{*}$}
\author[5]{Luke A. Brown}
\author[6,7,8]{Phuong D. Dao}
\author[9]{Kyle R. Kovach}
\author[10]{Bing Lu}
\author[1]{Daniel Mederer}
\author[1,2,12,13]{Hannes Feilhauer$^{*}$}
\author[11,12]{Teja Kattenborn$^{*}$}
\author[3,4]{David Rolnick$^{*}$}

\affil[1]{\tiny Institute for Earth System Science and Remote Sensing, Leipzig University, Germany}
\affil[2]{Center for Scalable Data Analytics and Artificial Intelligence (ScaDS.AI), Leipzig University, Germany}
\affil[3]{Mila – Québec AI Institute, Canada}
\affil[4]{McGill University, Canada}
\affil[5]{School of Science, Engineering \& Environment, University of Salford, UK}
\affil[6]{Department of Agricultural Biology, Colorado State University, USA}
\affil[7]{Graduate Degree Program in Ecology, Colorado State University, USA}
\affil[8]{School of Global Environmental Sustainability, Colorado State University, USA}
\affil[9]{Department of Forest and Wildlife Ecology, University of Wisconsin, USA}
\affil[10]{Department of Geography, Simon Fraser University, Canada}
\affil[11]{Chair of Sensor-based Geoinformatics (geosense), University of Freiburg, Germany}
\affil[12]{German Centre for Integrative Biodiversity Research (iDiv), Halle-Jena-Leipzig, Germany}
\affil[13]{Helmholtz-Centre for Environmental Research (UFZ), Leipzig, Germany}

\begin{document}
\maketitle

\begin{abstract}
Plant traits such as leaf carbon content and leaf mass are essential variables in the study of biodiversity and climate change. However, conventional field sampling cannot feasibly cover trait variation at ecologically meaningful spatial scales. 
Machine learning represents a valuable solution for plant trait prediction across ecosystems, leveraging hyperspectral data from remote sensing.
Nevertheless, trait prediction from hyperspectral data is challenged by label scarcity and substantial domain shifts (\eg across sensors, ecological distributions), requiring robust cross-domain methods.
Here, we present GreenHyperSpectra, a pretraining dataset encompassing real-world cross-sensor and cross-ecosystem samples designed to benchmark trait prediction with semi- and self-supervised methods. We adopt an evaluation framework encompassing in-distribution and out-of-distribution scenarios. 
We successfully leverage GreenHyperSpectra to pretrain label-efficient multi-output regression models that outperform the state-of-the-art supervised baseline. 
Our empirical analyses demonstrate substantial improvements in learning spectral representations for trait prediction, establishing a comprehensive methodological framework to catalyze research at the intersection of representation learning and plant functional traits assessment. We also share the dataset\footnotemark[1], code and pretrained model objects 
for this study \href{https://github.com/echerif18/HyspectraSSL}{here}.

\footnotetext[1]{GreenHyperSpectra dataset: \href{https://huggingface.co/datasets/Avatarr05/GreenHyperSpectra}{https://huggingface.co/datasets/Avatarr05/GreenHyperSpectra}}
\end{abstract}

\section{Introduction}
\label{sec:intro}
Plant functional traits are a fundamental component of biodiversity assessment, offering insights into plant productivity, ecological interactions, resilience, and adaptation to environmental change \citep{cavender2020remote, funk2017revisiting, skidmore2021priority, yan2023essential}. Leaf traits such as leaf mass per area, as well as chlorophyll, nitrogen, and carbon content, are key to understanding plant growth dynamics and ecosystem processes such as carbon cycling and productivity \citep{berger2022multi, bongers2021functional, damm2018remote, roscher2012using, zarco2018previsual, zarco2019chlorophyll}. 
The monitoring of these traits is thus crucial for understanding ecosystem function and guiding biodiversity conservation strategies \citep{cavender2022integrating, pettorelli2021time, xu2021ensuring}. 
\begin{wrapfigure}{r}{0.5\textwidth}
\centering
\includegraphics[width=0.50\textwidth]{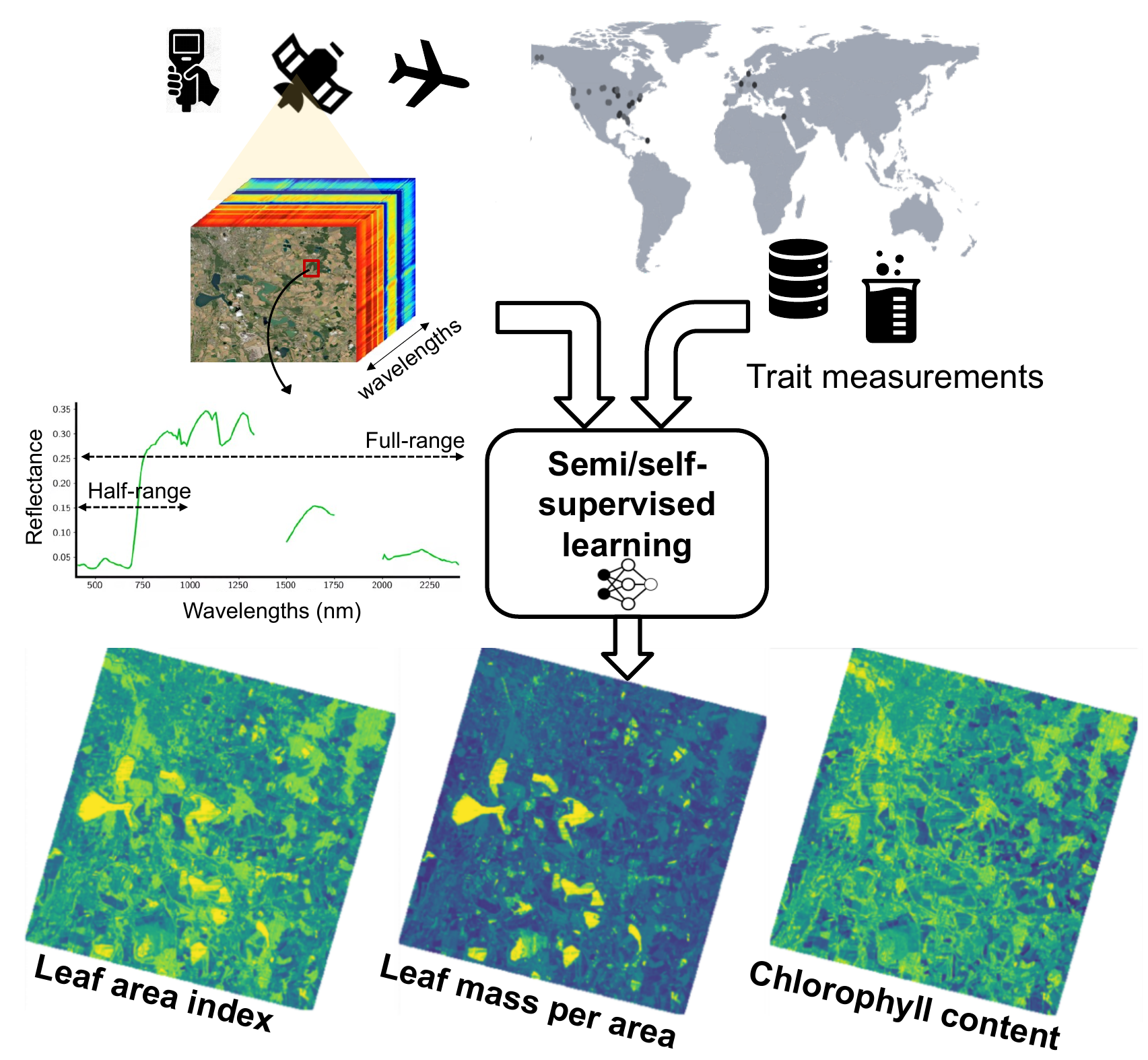}
\caption{Overview of the semi/self-supervised framework for multi-trait regression task.}
  \label{fig:teaser}
\vspace{-5mm}
\end{wrapfigure}
Initiatives such as the Intergovernmental Science-Policy Platform on Biodiversity and Ecosystem Services (IPBES; \citep{diaz2015ipbes, ipbes2019global}) have raised global awareness around the urgent need for monitoring functional traits and their diversity across spatial scales. However, we still lack efficient tools to track these functional traits in space and time. Hyperspectral remote sensing from airborne and satellite systems has emerged as a promising tool to bridge this gap, enabling non-destructive, scalable, and repeatable reflectance measurements that can be used to predict these traits \citep{berger2020crop, cherif2023spectra, jetz2016monitoring, serbin2019arctic, van2018functional}. Hyperspectral sensors measure radiation reflected from the ground across hundreds of narrow, contiguous spectral bands, spanning the visible to shortwave infrared (VNIR + SWIR) domains. These measurements are informative for trait prediction, as they are directly influenced by the chemical and structural characteristics of plant leaves and canopies \citep{jacquemoud2019leaf}. 
Plant trait prediction from hyperspectral data is inherently a regression problem, and was initially often explored with Partial Least Squares Regression (PLSR, \citep{geladi1986partial}) to link hyperspectral observations to individual traits \citep{feilhauer2010brightness, serbin2015remotely, helsen2021evaluating,ustin2009retrieval}. 
However, non-parametric machine learning methods, in particular deep learning, have recently been explored to offer greater flexibility in modeling complex, non-linear trait-spectral relationships and trait-trait interactions \citep{cherif2023spectra, tsakiridis2020simultaneous, he2016novel}. In this context, trait prediction is framed as a multi-output regression task within a multi-task learning framework, where the outputs, representing multiple plant traits, are inherently correlated.

\begin{wrapfigure}{r}{0.5\textwidth}

  \centering
    \includegraphics[width=0.99\linewidth, keepaspectratio]{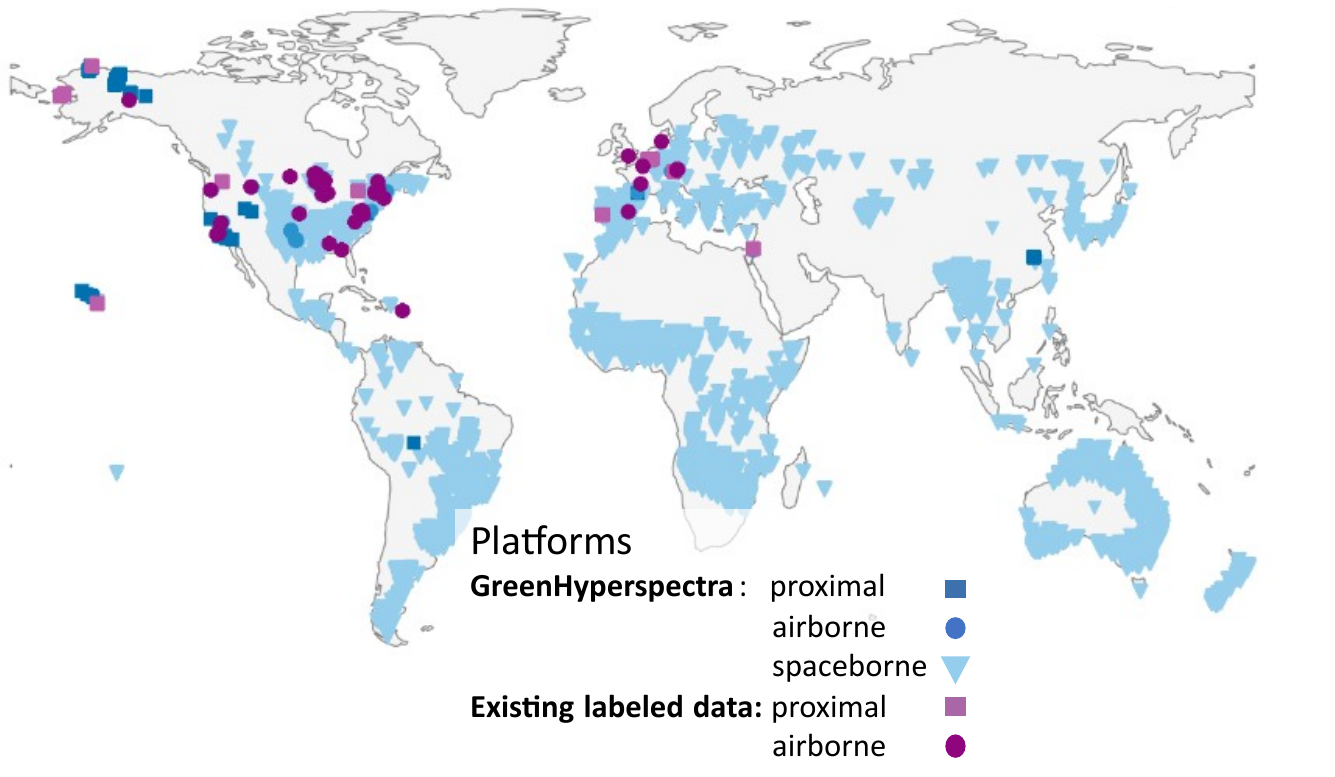}

  \vspace{1ex} 
  
  \caption{\textbf{Spatial coverage of the datasets.} Points represent sample locations of GreenHyperSpectra compared to the existing labeled dataset. GreenHyperSpectra data span diverse vegetation type and acquisition conditions. 
  }
  \label{fig:coverage}
  \vspace{-5mm}
\end{wrapfigure}

Trait prediction poses fundamental machine learning challenges including heterogeneous target distributions requiring specialized multi-task methods \cite{cherif2023spectra}, extreme label scarcity, and significant distributional biases (\eg spatial, Figure~\ref{fig:coverage}). Current supervised approaches demonstrate limited cross-domain generalization due to training on sparse, non-representative datasets \citep{danilevicz2022plant, pichler2020machine}. 
Hyperspectral data further compound these challenges 
exhibiting substantial covariate shifts across acquisition conditions, sensor configurations and resolutions, radiometric calibrations and variable input modalities. 
To address these limitations, we introduce GreenHyperSpectra, a large spectral dataset designed to improve representation robustness against domain adaptation challenges while being collected from multiple ecosystems, instruments, spatial resolutions, and acquisition conditions. This dataset enables semi- and self-supervised learning applications, which take advantage of vast unlabeled spectral data, providing extensive coverage and variability to facilitate benchmarking.

Our contributions include:
\circledgray{1} Building GreenHyperSpectra, a dataset for pretraining consisting of cross-domain samples and substantially expanding available datasets for representation learning; 
\circledgray{2} framing a suite of semi- and self-supervised methods for multi-output regression with one dimensional (1D) hyperspectral data;
\circledgray{3} comparing these methods with fully supervised baseline, highlighting the superior performance of the former, particularly in scenarios with limited labeled data;
\circledgray{4} Testing how well such methods generalize across variable inputs, representing the diversity of sensor settings (full-range (VNIR+SWIR) vs. half-range (VNIR-only), see Figure~\ref{fig:teaser}).

\section{Related work}
\label{sec:relatedwork}
Despite increasing hyperspectral data availability, plant trait prediction is still largely constrained by the lack of large annotated datasets \citep{danilevicz2022plant, todman2023small}. Trait labels are costly and time-consuming to obtain, often requiring field sampling and laboratory analysis ~\citep{cornelissen2003handbook, baraloto2010functional}. Available datasets lack harmonization and often differ in sampling strategy, measurement assumptions and protocols. As a result, most labeled datasets are geographically and ecologically limited, with sparse coverage across space and time, as well as across ecosystems and acquisition conditions.
To address this, previous studies explored synthetic datasets generated from Radiative Transfer Models (RTMs, \citep{feret2008prospect}) that simulate canopy spectral responses under diverse conditions. In this context, it motivated hybrid approaches that combine RTM-based simulations with machine learning \citep{danner2021efficient, mederer2025plant, parigi2024towards, tagliabue2022hybrid, verrelst2015optical, wang2021airborne}.

%
However, several comparisons reveal important limitations, with models trained on real, multi-site datasets consistently outperforming those using only synthetic spectra \citep{mederer2025plant, parigi2024towards}, highlighting a persistent domain gap between simulated and field data.

In this regard, there is a need of large-scale unlabeled spectral datasets from real measurements to pretrain models. 
To support this, a number of benchmarking efforts have been introduced \citep{fuchs2023hyspecnet, wang2025hypersigma,braham2024spectralearth}.
However, most of these initiatives are built from a single sensor type, restricting spectral diversity and limiting generalization to new acquisition conditions. Moreover, these datasets often consist of full hyperspectral imagery, where spatially contiguous pixels are subject to spatial autocorrelation. This spatial redundancy reduces the spectral variability necessary for training models that generalize well in trait prediction task, which depend on spectral rather than spatial information \citep{thoreau2024toulouse}. 
They also typically include a broad mix of land cover types, including non-vegetated surfaces. This introduces inefficiencies, as vegetated pixels must be sampled through additional preprocessing.

Large-scale unlabeled datasets have been leveraged for self-supervised learning to improve natural image representations \cite{Chen_2021_ICCV, Assran2023Self, oquab2024dinov} as well as in remote sensing with multispectral data \cite{ tseng2023lightweight, NEURIPS2023_11822e84, xiong2024neuralplasticityinspiredmultimodalfoundation, szwarcman2025prithvieo20versatilemultitemporalfoundation, bountos_fomo_2025, astruc2025anysatearthobservationmodel, tseng2025galileolearninggloballocal}.
Semi- and self-supervised learning techniques are increasingly being explored to exploit hyperspectral data for image classification, segmentation, and super-resolution \citep{ahmad2021hyperspectral, ranjan2023unlocking, wang2022self}.
These methods are designed to better learn the spectral representation of hyperspectral remote sensing data to reduce the need for reference labels.
Approaches such as masked autoencoders (MAE) \citep{cong2022satmae, hong2021spectralformer, reed2023scale, scheibenreif2023masked, thoreau2024toulouse}, contrastive learning frameworks \citep{chen2020simple, fuller2023croma, he2020momentum}, generative networks (GAN) \citep{alipour2020structure, he2017generative, kwak2023semi, roy2021generative, zhan2018semi, zhu2018generative}, and autoencoders (AE) \citep{ahmad2019segmented, gallo2023self, thoreau2024toulouse, zhao2017spectral} have been successfully applied for land cover classification. 
Whereas most attempts at trait prediction using hyperspectral data relied on fully supervised pipelines, notably PLSR
\citep{serbin2015remotely, singh2015imaging, ustin2009retrieval, wang2020foliar}, Gaussian Process Regression (GPR) \citep{tagliabue2022hybrid, verrelst2013gaussian, verrelst2012gaussian}, ANN \citep{schlerf2006inversion} and deep learning methods \citep{cherif2023spectra, pullanagari2021field, shi2022convolution}, they are cardinally constrained by label scarcity, hampering their ability to reliably generalize across ecosystems, sensor platforms, and acquisition conditions.
Applying semi- and self-supervised methods to trait prediction remains largely unexplored and offers a compelling direction for investigation.
Recent semi- and self-supervised applications in trait prediction include vision transformers for nitrogen estimation from simulated data \citep{gallo2023self} and Long short-term memory (LSTM) models for chlorophyll prediction with limited spectral bands \citep{zhao2023improving} (\ie VNIR-only).
While semi- and self-supervised methods show promise for trait prediction, existing models remain constrained to single traits. Moreover, existing models are typically limited to specific sensor configurations and experimental conditions, limiting generalization across sensor modalities (\eg full-range vs. half-range spectrometers), acquisition geometries, and vegetation types. This underscores the need for flexible approaches that can handle heterogeneous inputs while supporting transferable predictions in diverse real-world ecological scenarios. To the best of our knowledge, no semi- or self-supervised method addresses trait prediction via multi-output regression.

\section{The GreenHyperSpectra dataset}
\label{sec:GreenSpecdata}
We introduce a large-scale, multi-source hyperspectral dataset comprising over $140,000$ vegetation canopy surface reflectance spectra captured across diverse continents, ecosystems, sensor platforms, spatial resolutions, and measurement geometries. Unlike existing benchmarks of hyperspectral data limited to single sensors or narrow ecological domains, our dataset features a substantially larger pretraining spectral dataset supporting semi- and self-supervised learning approaches.

\paragraph{Acquisition platforms and sensor diversity.}
We curated spectral data from multiple instruments across three primary platforms: proximal, airborne, and spaceborne (Figure \ref{fig:coverage}). 
All data were processed to the level of at-surface reflectance. 
Proximal measurements were obtained using field spectrometers such as the ASD FieldSpec and SVC HR-1024i, typically positioned in a close range in nadir orientation to record top-of-canopy reflectance.
Airborne data were acquired using high-spectral resolution sensors, including the AVIRIS-Next Generation, AVIRIS-Classic, NEON Airborne Observation Platform (AOP) and Specim AISAFenix instruments, which cover landscape-level vegetation scenes with variable viewing geometries and meter-scale spatial resolutions. 
Spaceborne acquisitions were collected from missions such as PRISMA, Hyperion, EMIT, and EnMAP, offering a larger scale 
\begin{wraptable}{r}{0.35\textwidth}
\centering
\vspace{-2mm}
\resizebox{.99\linewidth}{!}{
  
  \begin{tabular}{l c c l l}
    \toprule
    Platform & GSD & Spectral res. & \#Samples \\
    \midrule
    Proximal   & <1 m    & 1--4 nm    & 5620   \\
    Airborne   & 1--20 m & 3--7 nm    & 96699  \\
    Spaceborne & 30--60 m & 6--12 nm  & 36059  \\
    \bottomrule
  \end{tabular}
  }
  \caption{Specifications of spectroscopy instruments with different platforms.}
  \label{tab:instrument-specs}
  \vspace{-7mm}
\end{wraptable}
observations at 30-60 m resolution with varying viewing geometry.
Table \ref{tab:instrument-specs} summarizes the platforms, spectral properties, and scene-level characteristics associated with each acquisition (more details see Appendix~\ref{sec:datasetinfo}). The pre-processing of spectra harmoinization is described in Appendix ~\ref{sec:datasetinfo}. 

The multi-platform nature of our dataset introduces valuable reflectance signal variability through differences in spatial and spectral resolution, sun-sensor geometry, scene heterogeneity, background conditions and pre-processing from radiance to reflectance. 
This variability, often lacking in single-platform or synthetic datasets, is essential to develop generalizable models capable of scaling across diverse remote sensing contexts \citep{cherif2023spectra}.
While satellite-based datasets such as SpectralEarth \citep{braham2024spectralearth} provide temporally rich but sensor-specific imagery, our dataset uniquely incorporates multi-sensor observations across spaceborne, airborne, and proximal platforms.

\paragraph{Spatial and temporal coverage.}
The dataset includes samples from diverse biomes, with acquisitions spanning from 1992 to 2024, capturing broad ecological and climatic 
variability across a wide range of environments. Figure \ref{fig:coverage} maps the global spatial distribution of GreenHyperSpectra and a pool of previously aggregated datasets for plant trait prediction (see details in \S\ \ref{sec:methods} ). While the compiled labeled dataset is spatially limited, GreenHyperSpectra encompasses substantially broader spatial coverage and environmental heterogeneity, better representing real-world remote sensing operational conditions (more details in Appendix \ref{sec:datasetinfo}).

\section{Benchmarking methods and protocols}
\label{sec:methods}

\paragraph{Trait-annotated dataset.}
\label{sec:LbSection}
For benchmarking different semi- and self-supervised methods, we use an existing aggregated dataset \citep{cherif2023spectra} comprising $7,900$ canopy reflectance spectra with co-located measurements of seven functional plant traits: leaf mass per area (Cm) $[\text{g}/\text{cm}^2]$, leaf protein content (Cp) $[\text{g}/\text{cm}^2]$, equivalent water thickness (Cw) $[\text{cm}]$, leaf total chlorophyll (Cab)$[\mu\text{g}/\text{cm}^2]$, carotenoids (Car)$[\mu\text{g}/\text{cm}^2]$ and anthocyanins (Anth) $[\mu\text{g}/\text{cm}^2]$ content, and leaf area index (LAI) $[\text{m}^2/\text{m}^2]$. Trait values were obtained either through direct field measurements or via community-weighted means assigned at the pixel level based on ground-measured species composition.
For the analysis, we treat Cp and nitrogen as equivalent due to their strong correlation, while acknowledging that they are not strictly the same. Additionally, we introduce a derived trait, carbon-based constituents (cbc), which is computed as the difference between Cm and Cp. These data were aggregated from 50 experiments and campaigns \citep{gravel2024mapping, zheng2024variability, chadwick2023shift, brodrick2023shift, rogers2019leaf, dao2021mapping, burnett2021source, chlus2020mapping, brown2024hyperspectral, van2019novel, kattenborn2019differentiating, cerasoli2018estimating, ewald2020assessing, ewald2018analyzing, wocher2018physically, wang2016robust, wang2020foliar, herrmann2011lai, pottier2014modelling, hank2015neusling, hank2016neusling, singh2015imaging, dao2025imaging}, covering diverse vegetation types such as forests, croplands, tundra, and pastures. Trait values were harmonized by converting mass-based traits to area-based units \citep{kattenborn2019advantages}. The pre-processing of the spectra is similar to that described in Appendix ~\ref{sec:datasetinfo}.
This aggregated labeled dataset serves as the reference to train and evaluate the regression models.
To enhance training stability and better capture inter-trait correlations, we applied box-cox transformation \citep{10.1111/j.2517-6161.1964.tb00553.x} to the trait values \citep{cherif2023spectra} for all methods. 

\paragraph{Data splitting and evaluation protocol.}
\label{splitSection}
To ensure consistent representation of all contributing data sources during training, we divide GreenHyperSpectra into 20 non-overlapping subsets. In each split, the proportion of samples from any given data source matches that dataset’s overall contribution to the full merged dataset. This stratified splitting strategy maintains the natural diversity of vegetation types, sensors, and acquisition conditions, while preventing bias from individual sources by creating consistent and representative subsets suitable for semi- and self-supervised methods.
For the labeled dataset, we define standardized train and validation splits using a 80/20 hold-out strategy. 
The 80\%  portion is combined with the pretraining spectral dataset for calibration, while the remaining 20\% is fixed for all experiments and used to evaluate all methods. 
For out of distribution (OOD) evaluation experiments (detailed in \S\ref{sec:expe_settings}), we perform cross validation across the 50 labeled datasets. Specifically, from the 50 annotated sub-datasets, we hold out five datasets at a time for testing. The remaining datasets are used for training, with their data further split into 80\% for training and 20\% for validation.

\begin{figure}[t!] 
    \centering
    \small
    \label{fig:archi_all}
    \begin{subfigure}[c]{0.48\textwidth} 
        \centering
        \includegraphics[width=\linewidth, keepaspectratio]{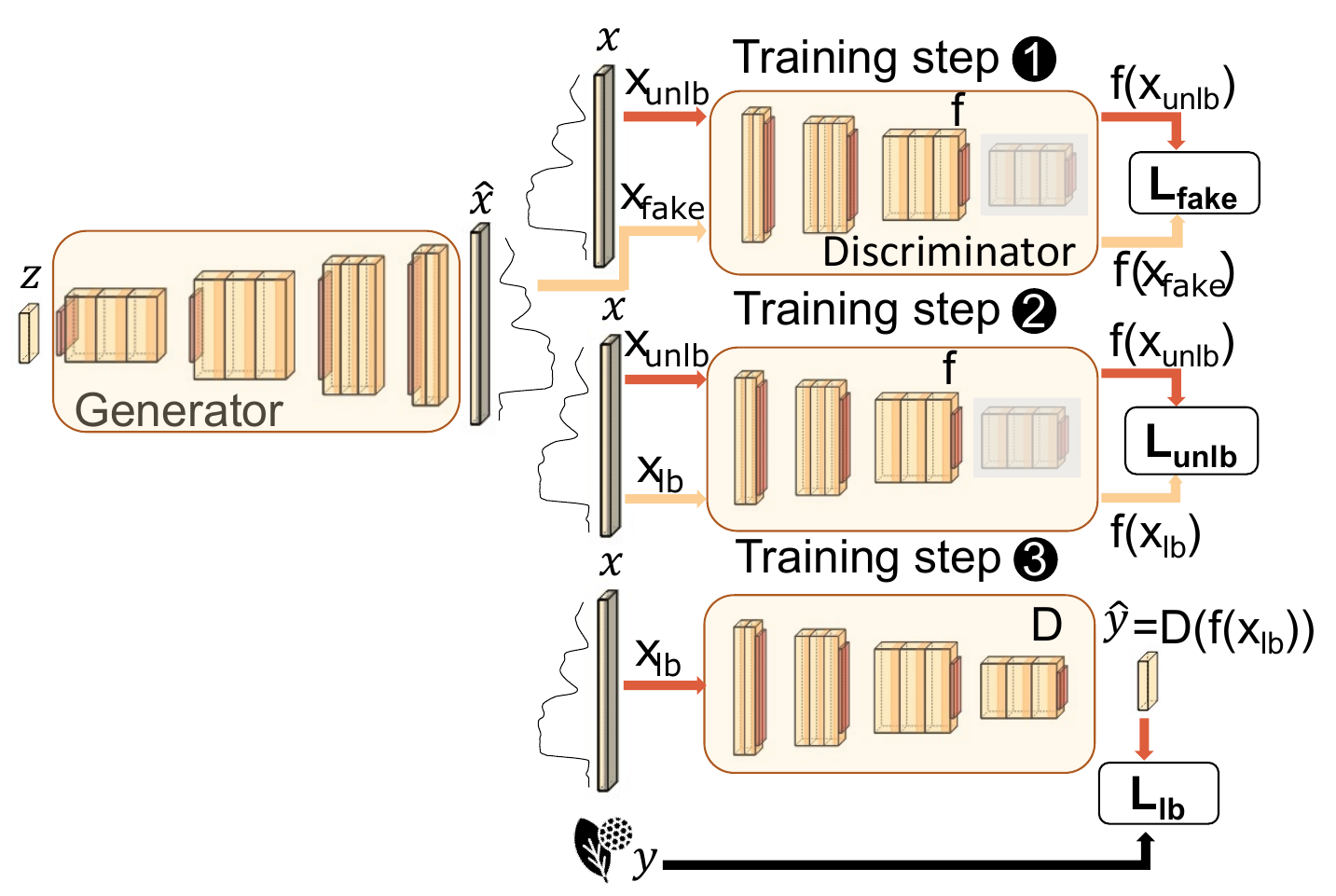}
        \caption{Semi-supervised generative adversarial network framework (SR-GAN).}
        \label{fig:archi_gan}
    \end{subfigure}
    \hfill
    \begin{subfigure}[c]{0.48\textwidth} 
        \centering
        \includegraphics[width=\linewidth, keepaspectratio]{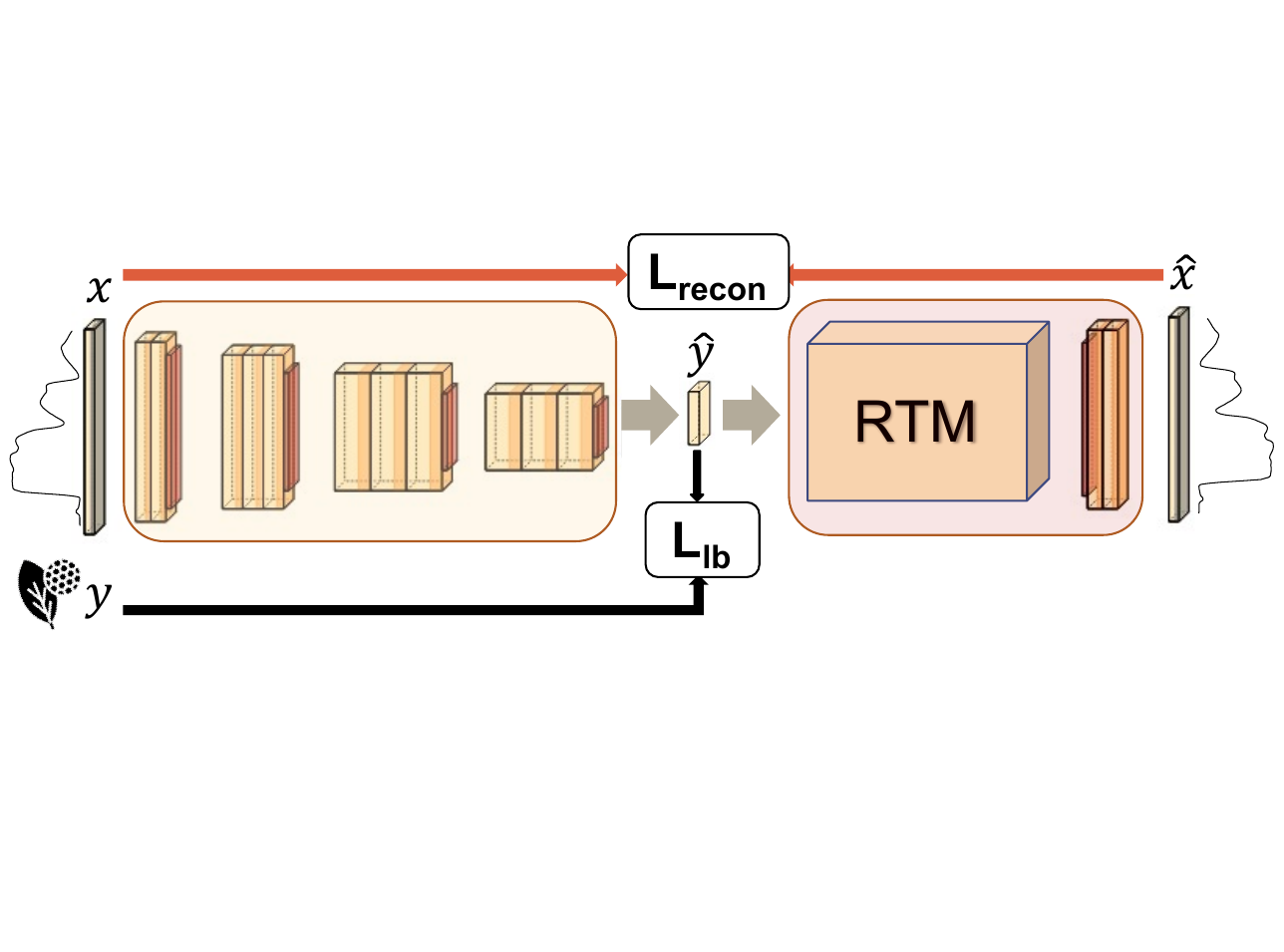}
        \vspace{-1.5cm}
        \caption{RTM-based autoencoder framework (RTM-AE).}
        \label{fig:archi_ae_rtm}
    \end{subfigure}
    \hfill
    \begin{subfigure}[c]{0.46\textwidth} %
        \centering
        \includegraphics[width=\linewidth, keepaspectratio]{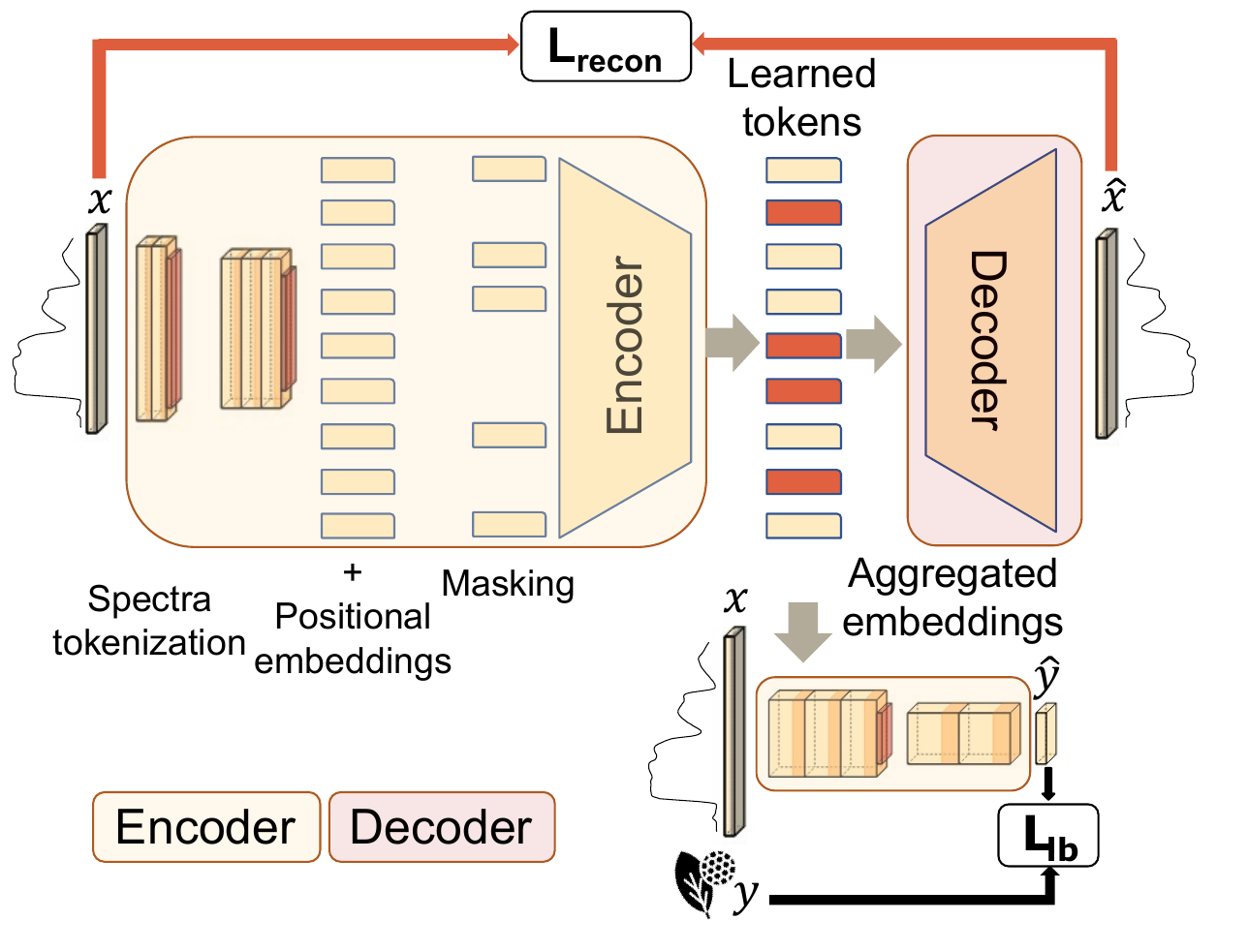}
        \caption{Masked-autoencoder framework (MAE).}  
        \label{fig:archi_mae}
    \end{subfigure}
    \caption{\textbf{Overview of the semi- and self-supervised frameworks.} (\ref{fig:archi_gan}) The semi-supervised regression GAN framework (SR-GAN): the generator maps a random noise \( z \) to synthetic samples \( \hat{x} \), while the discriminator processes \protect\circledblack{1} fake samples (\( x_{\text{fake}} \)), \protect\circledblack{2} unlabeled real samples (\( x_{\text{unlb}} \)), and \protect\circledblack{3} labeled real data samples (\( x_{\text{lb}} \)) with associated traits (y), optimizing fake (\( L_{\text{fake}} \)),  unlabeled (\( L_{\text{unlb}} \)), and labeled (\( L_{\text{lb}} \)) losses respectively. (\ref{fig:archi_ae_rtm}) The RTM-based autoencoder (RTM-AE) predicts traits from labeled embeddings while reconstructing spectra (\( x \to \hat{x} \), (\( L_{\text{recon}} \))). (\ref{fig:archi_mae}). The 1D masked autoencoder framework (1D-MAE) reconstructs masked spectra through tokenization, (\( L_{\text{recon}} \)); the learned representations are then used for trait prediction (\( L_{\text{lb}} \)). \textbf{Abbreviations}: \( x_{\text{fake}} \): generated fake spectra from the generator; \( x_{\text{unlb}} \): unlabeled sample from GreenHyperSpectra; \( x_{\text{lb}} \): spectra sample from the labeled data; \( L_{\text{unlb}} \): unlabeled loss; \( L_{\text{lb}} \): labeled loss; \( L_{\text{recon}} \): reconstruction loss; \( L_{\text{fake}} \): feature contrasting loss; RTM: radiative transfer model; AE: autoencoder; MAE: masked autoencoder.
    }
    \vspace{-5mm}
\end{figure}

\paragraph{Supervised baseline method.}
We consider a supervised CNN-based method \citep{cherif2023spectra, mederer2025plant} as a baseline, selected for its state-of-the-art performance in multi-trait plant prediction from a sparse annotated dataset.
It is built upon EfficientNet-B0 \citep{tan2019efficientnet} specifically framed for 1D feature extraction. The network employs multi-output regression to simultaneously predict the seven plant traits, a strategy that demonstrably outperforms single-trait modeling approaches \citep{cherif2023spectra, scutari2014multiple}. 

\paragraph{Semi-supervised regression generative adversarial network (SR-GAN).}
We frame the SR-GAN framework \citep{olmschenk2019generalizing} to address hyperspectral plant trait prediction. Our implementation employs a 1D convolutional GAN architecture designed specifically for spectral data processing. 
In this setup, the generator learns to produce synthetic reflectance spectra, while the discriminator simultaneously performs trait regression and learns discriminative feature representations. The training objective is formulated as a composite loss that encourages the discriminator to pull real spectral samples closer in the feature space, while pushing representations of generated (synthetic) samples further apart. This contrastive learning approach allows the model to leverage unlabeled data by learning informative spectral embeddings. The overall architecture of the SR-GAN framework is illustrated in Figure~\ref{fig:archi_gan} and detailed formulations of the loss components are provided in Appendix~\ref{sec:GAN_details}. 

\paragraph{Radiative transfer model based autoencoder (RTM-AE).}
We introduce a version of the autoencoder framework proposed by \citep{she2024spectra}, which replaces the decoder with a non-learnable RTM module to reconstruct spectra, thereby integrating physical constraints into the modeling process.
Specifically, our implementation employs PROSAIL-PRO \citep{feret2021prospect}, constraining the latent space to correspond directly to plant traits. PROSAIL-PRO is an RTM that combines the leaf reflectance model (PROSPECT,\citep{jacquemoud1990prospect}) with the canopy reflectance model (4SAIL,\citep{verhoef2007unified}). PROSPECT simulates leaf reflectance and transmittance based on biochemical composition and internal structure, while 4SAIL models the propagation of light through a vegetation canopy. Together, they simulate canopy spectral reflectance in the 400–2500 nm range using inputs such as chlorophyll content, leaf area index, and leaf angle.
As previously mentioned regarding the gap between RTM-simulated and real-world spectra (\S~\ref{sec:relatedwork}), we address the inherent discrepancies between RTM-generated and observed spectra, primarily resulting from simplified geometric assumptions within the model, by implementing a learnable correction layer that refines the simulated output \citep{she2024spectra}.
Our enhanced framework introduces three key improvements over the original design \citep{she2024spectra}: (1) incorporation of PROSAIL-PRO, (2) application of a supervised loss component targeting trait predictions, and (3) implementation of a composite reconstruction loss combining cosine similarity and mean absolute error to capture both spectral shape characteristics and amplitude information. The overall architecture is illustrated in Figure~\ref{fig:archi_ae_rtm} and the specifications are detailed in Appendix~\ref{sec:RTM_details}  
.
\paragraph{Masked autoencoder (MAE).}
We adopt a MAE framework, originally designed for land cover classification \citep{thoreau2024toulouse}, to predict plant traits with hyperspectral data. 
The model leverages self-supervised learning by reconstructing randomly masked spectral regions, enabling the extraction of meaningful representations from unlabeled hyperspectral signatures. 
Similarly to the RTM-AE, our adaptation incorporates a modified reconstruction objective that combines cosine similarity and mean squared error (MSE) with appropriate weighting, allowing the model to capture both spectral shape characteristics and amplitude information.
For downstream trait prediction, we attach a multi-output regression head to the latent features and fine-tune the model using labeled data. The overall architecture is illustrated in Figure~\ref{fig:archi_mae} and ablation studies for the MAE architecture are provided in Appendix~\ref{sec:MAE_details}.

\begin{figure}[t!]
  \centering
    \includegraphics[width=0.75\linewidth, height=0.75\textheight, keepaspectratio]{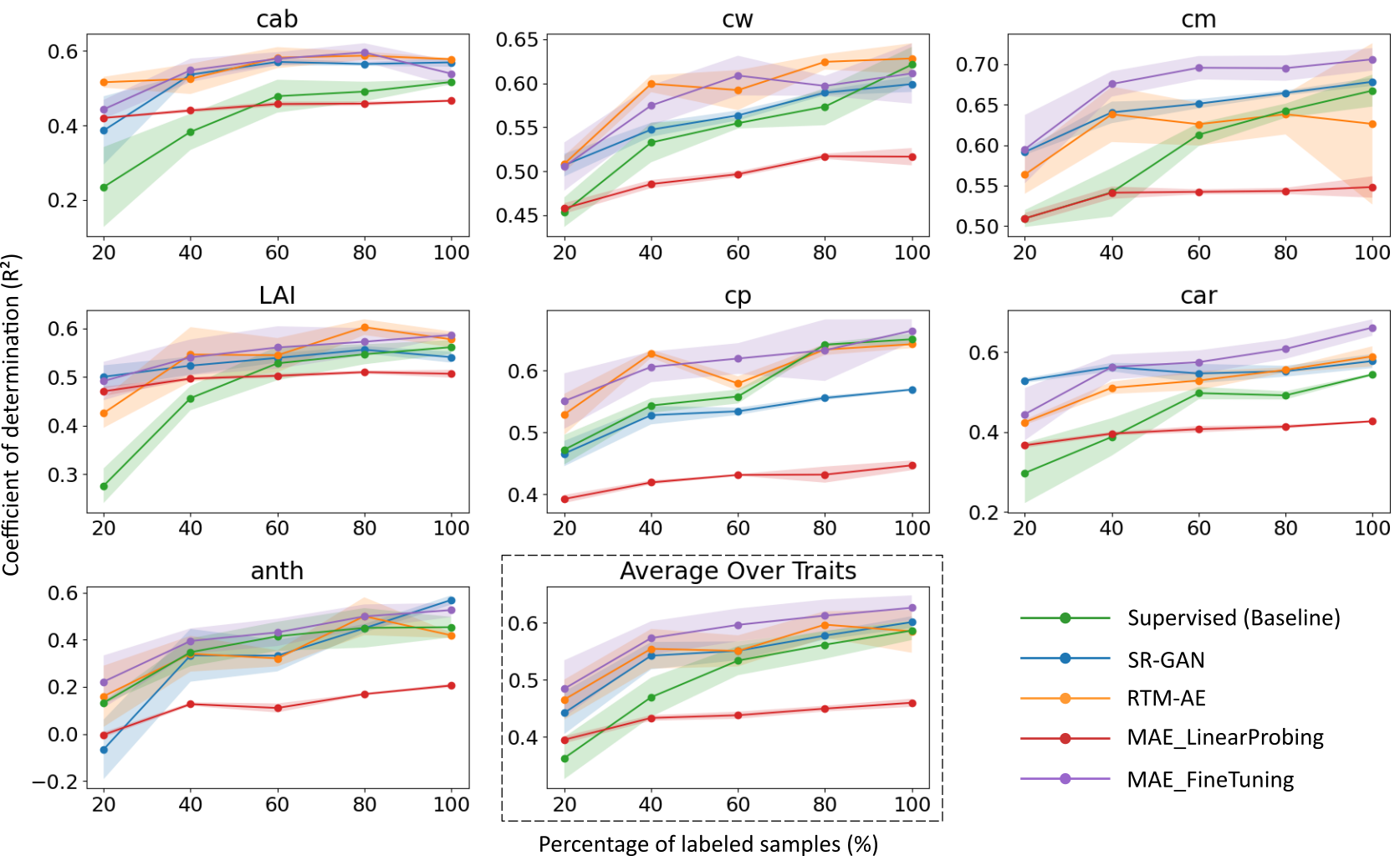}

  \vspace{1ex} 
  
  \caption{\textbf{Evaluation of trait prediction with variable-size labeled sets.} Validation performance ($R^2$) as a function of labeled data percentage used for training. The average $R^2$ performance across all traits is indicated by the dashed box. The higher $R^2$, the better. 
  For trait abbreviations, see Sec.~\ref{sec:LbSection}.
  }
  \label{fig:res_labeled}
  \vspace{-5mm}
\end{figure}

\section{Experimental settings}
\label{sec:expe_settings}
This section describes experimental setups used to benchmark and evaluate the performance of models trained on semi- and self-supervised learning fashion for multi-trait plant prediction detailed in \S~\ref{sec:methods}. 
Our experimental framework is structured into four principal components to test the capabilities of models across a range of scenarios reflecting critical use cases: comprehensive benchmarking using full-range (FR) spectra, benchmarking with half-range (HR) spectra, and assessment of OOD generalization capabilities, along with an ablation study on the design of the MAE models.
Throughout these experiments, we maintain standardized data splits for both labeled and unlabeled datasets as described in \S~\ref{splitSection}. Complete specifications regarding model hyperparameters, optimization settings, and implementation details are provided in Appendix~\ref{sec:model_details}.

\paragraph{Full-range trait prediction.}
We assess all benchmark models using the full-range spectra spanning 400–2450 nm (1721 bands), encompassing visible through shortwave infrared wavelengths. 
Results are presented in \S~\ref{sec:results} with Table~\ref{tab:fr_results}.

\begin{figure}[t!]
  \centering
    \includegraphics[width=0.75\linewidth, height=0.75\textheight, keepaspectratio]{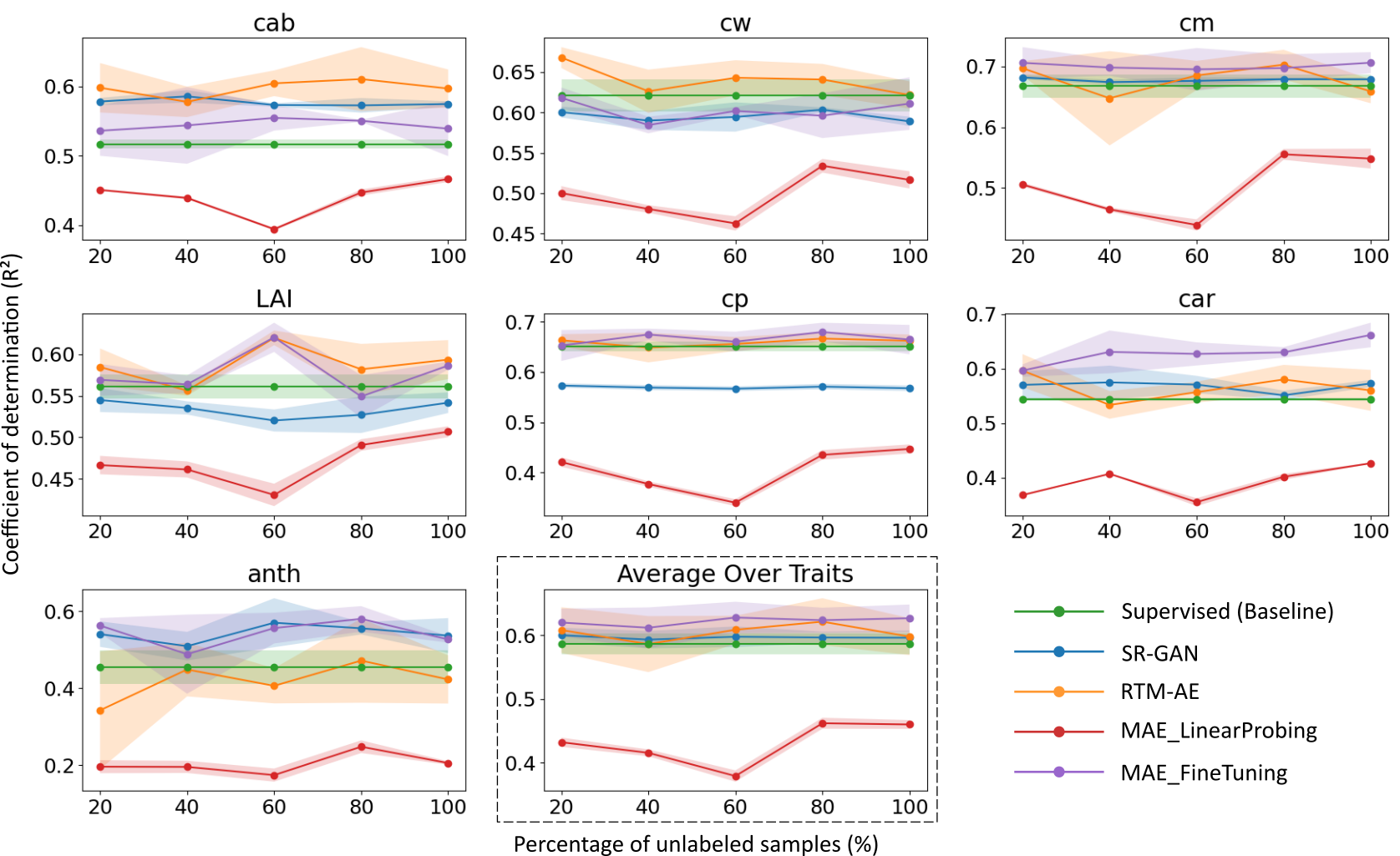}%

  \vspace{1ex} 
  
  \caption{\textbf{Evaluation of trait prediction with variable-size unlabeled sets.} Validation performance ($R^2$) as a function of the percentage of unlabeled data used for training. The average $R^2$ performance is indicated by the dashed box. The higher $R^2$, the better. For trait abbreviations, see Sec.~\ref{sec:LbSection}.
  }
    \label{fig:res_unlabeled}
    \vspace{-5mm}
\end{figure}

\paragraph{Sample sensitivity analysis.}
We examine the impact of label availability by simulating different levels of supervision and varying the amount of labeled data used for training from 20\% to 100\% while maintaining a consistent unlabeled dataset. Complementing this approach, we conduct experiments varying the quantity of unlabeled training data while maintaining fixed labeled data proportions to determine how unlabeled data volume influences model performance; note that in these experiments, we use only a subset of the full GreenHyperSpectra dataset ($80,000$ samples).
Results are presented in \S~\ref{sec:results}, with Figure~\ref{fig:res_labeled} and ~\ref{fig:res_unlabeled}.

\paragraph{Half-range trait prediction.}
A common constraint faced with satellite-based Earth observations is that many sensors do not cover the full spectrum. To evaluate model performance in this scenario, we replicate our benchmark procedure using only the half-range spectral subset spanning 400–900nm (500 bands). All models are trained on this spectral subset. Additionally, we implement an evaluation for the MAE architecture, where a model pretrained on full-range spectra is applied to half-range spectra inputs (this is possible only for the MAE models as the masking procedure means that they can accommodate variable input sizes). 
Results are presented in \S~\ref{sec:results} and Table~\ref{tab:hr_results}.

\paragraph{OOD evaluation.}
To assess each model's robustness to real-world distribution shifts, we perform a cross-dataset evaluation as described in \S~\ref{splitSection}. 
We compute a macro-level performance metric by aggregating predictions across all held-out datasets.
This setup reflects practical challenges in ecological monitoring applications, where spectral variability arises from differences in acquisition conditions, sensor platforms, or environmental contexts. 
Additionally, this approach ensures a broader coverage of trait value ranges, which often remain underrepresented when test sets are randomly sampled. 
Due to computational constraints, we conduct this evaluation using a single training run. To reduce the sensitivity of $R^2$ to unbalanced number of samples across the 50 aggregated datasets, we compute the macro-average over five random subsamples within each dataset, each constrained to the maximum number of 30 samples allowed per set, and report the mean and standard deviation of the resulting metrics. Results are presented in \S~\ref{sec:results} and Table~\ref{tab:ood_results}. 

\paragraph{Ablation studies on MAE.}
We conduct comprehensive ablation experiments across several dimensions to consistently evaluate design trade-offs of the MAE models in spectral representation learning.
First, we explore architectural complexity through a grid search spanning transformer configurations with varying numbers of layers $\{6, 8, 10\}$ and attention heads $\{4, 8, 16\}$. We select the configuration demonstrating optimal performance on the downstream trait prediction task for subsequent experiments. 
Considering this optimal architecture, we investigate alternative loss formulations for spectral reconstruction.
Beyond conventional MSE, we examine hybrid approaches which incorporate cosine similarity loss weighted by a coefficient $\alpha \in \{1, 0.1, 0.01, 0\}$ to enhance capture of spectral shape specificity. 
Finally, we assess the effect of token granularity by varying patch sizes (10, 20, 40, and 430) used during spectral masking and reconstruction in the full-range scenario. These targeted ablations informed the final MAE configuration used in our other benchmark evaluations.
All results of the aforementioned ablation studies are presented in Tables~\ref{tab:mae_ablation}, \ref{tab:mae_loss_ablation} and \ref{tab:mae_patchsize_ablation} in Appendix~\ref{sec:MAE_details}.

\paragraph{Evaluation metrics.}
For each experimental setting, we report the performance metrics averaged across three random seeds to measure the variability related to stochastic training effects. Our evaluation framework employs two complementary metrics: the coefficient of determination ($R^2$) and the normalized root mean square error (nRMSE). The nRMSE (in \%) is computed by normalizing the root mean square error by the
range of the traits observations (1–99\% quantile), providing a scale-invariant measure of prediction error.

\section{Results and discussions}
\label{sec:results}
\paragraph{Labeled and unlabeled data regimes.} 
To assess each model's sensitivity to the quantity of annotated samples, we analyze $R^2$ as a function of the proportion of available labeled and unlabeled data, as shown in Figures~\ref{fig:res_labeled} and \ref{fig:res_unlabeled} respectively. The corresponding trends for nRMSE are presented in Appendix~\ref{sec:resComplements}.
We observed that models leveraging unlabeled data through semi- and self-supervised methods consistently outperformed the fully supervised baseline, particularly in low-data regimes (20–40\% labeled data). Notably, semi- and self-supervised methods achieved higher average $R^2$ and lower nRMSE scores across most traits as labeled data availability decreased (Fig.~\ref{fig:res_labeled}).
This demonstrates that access to a large set of unlabeled spectra through GreenHyperSpectra substantially enhances model performance, leading to improved trait prediction accuracy. 
Interestingly, varying the size of the dataset for pretraining did not substantially impact performance. This suggests that the stratified splitting protocol (\S~\ref{splitSection}), ensuring consistent coverage across spectral sources, vegetation types, and acquisition conditions, plays a critical role in efficiently exploiting the available unlabeled data, even when subsampled.

\vspace{5mm}
\begin{table}[t!]
  \centering
  \resizebox{\textwidth}{!}{%
  \begin{tabular}{lccccccccc}
    \toprule
    & cab & cw & cm & LAI & cp & cbc & car & anth & \lightgrey average \\
    \midrule
    \multicolumn{10}{c}{$R^2$ $(\uparrow)$} \\
    \midrule
    Supervised & 0.517 \scriptsize (± 0.012) & 0.621 \scriptsize (± 0.009) & 0.667 \scriptsize (± 0.005) & \underline{0.561 \scriptsize (± 0.011)} & 0.651 \scriptsize (± 0.002) & 0.679 \scriptsize (± 0.006) & 0.544 \scriptsize ± (0.018) & 0.454 \scriptsize (± 0.020) & \lightgrey 0.587 \scriptsize (± 0.010) \\
    SR\_GAN & \underline{0.574 \scriptsize(± 0.008)} & 0.572 \scriptsize(± 0.013) & 0.669 \scriptsize(± 0.008) & 0.538 \scriptsize(± 0.005) & 0.558 \scriptsize(± 0.005) & \underline{0.704 \scriptsize(± 0.004)} & \underline{0.578 \scriptsize(± 0.006)} & \underline{0.541 \scriptsize(± 0.041)} & \lightgrey \underline{0.592 \scriptsize(± 0.011)} \\
    RTM\_AE & \textbf{0.584 \scriptsize(± 0.023)} & \textbf{0.658 \scriptsize(± 0.011)} & \underline{0.679 \scriptsize(± 0.021)} & 0.552 \scriptsize(± 0.029) & \underline{0.671 \scriptsize(± 0.015)} & 0.689 \scriptsize(± 0.021) & 0.566 \scriptsize(± 0.038) & 0.337 \scriptsize(± 0.125) & \lightgrey \underline{0.592 \scriptsize(± 0.035)} \\
    MAE\_FR\_LP & 0.462 \scriptsize(± 0.002) & 0.514 \scriptsize(± 0.009) & 0.577 \scriptsize(± 0.010) & 0.470 \scriptsize(± 0.007) & 0.464 \scriptsize(± 0.007) & 0.611 \scriptsize(± 0.008) & 0.412 \scriptsize(± 0.002) & 0.217 \scriptsize(± 0.015) & \lightgrey 0.466 \scriptsize(± 0.008) \\
    MAE\_FR\_FT & 0.515 \scriptsize(± 0.026) & \underline{0.634 \scriptsize(± 0.014)} & \textbf{0.716 \scriptsize(± 0.027)} & \textbf{0.615 \scriptsize(± 0.016)} & \textbf{0.676 \scriptsize(± 0.010)} & \textbf{0.727 \scriptsize(± 0.034)} & \textbf{0.649 \scriptsize(± 0.012)} & \textbf{0.598 \scriptsize(± 0.025)} & \lightgrey \textbf{0.641 \scriptsize(± 0.020)} \\
    \midrule
    \multicolumn{10}{c}{nRMSE $(\downarrow)$} \\
    \midrule
    Supervised & 17.341 \scriptsize(± 0.221) & 13.103 \scriptsize(± 0.150) & 10.671 \scriptsize(± 0.083) & 17.487 \scriptsize(± 0.218) & 10.229 \scriptsize(± 0.021) & 10.634 \scriptsize(± 0.099) & 13.647 \scriptsize(± 0.274) & 16.465 \scriptsize(± 0.333) & \lightgrey 13.697 \scriptsize(± 0.175) \\
    SR\_GAN & \underline{16.277 \scriptsize(± 0.161)} & 13.918 \scriptsize(± 0.208) & 10.658 \scriptsize(± 0.127) & 17.829 \scriptsize(± 0.091) & 11.565 \scriptsize(± 0.074) & \underline{10.258 \scriptsize(± 0.076)} & \underline{13.123 \scriptsize(± 0.095)} & \underline{15.084 \scriptsize(± 0.662)} & \lightgrey 13.589 \scriptsize(± 0.187) \\
    RTM\_AE & \textbf{16.096} \scriptsize(± 0.437) & \underline{12.443 \scriptsize(± 0.201)} & \underline{10.492 \scriptsize(± 0.350)} & \underline{17.556 \scriptsize(± 0.573)} & \underline{9.989 \scriptsize(± 0.224)} & 10.495 \scriptsize(± 0.350) & 13.297 \scriptsize(± 0.596) & 18.091 \scriptsize(± 1.761) & \lightgrey \underline{13.557 \scriptsize(± 0.561)} \\
    MAE\_FR\_LP & 18.297 \scriptsize(± 0.034) & 14.830 \scriptsize(± 0.135) & 12.052 \scriptsize(± 0.146) & 19.094 \scriptsize(± 0.133) & 12.746 \scriptsize(± 0.080) & 11.747 \scriptsize(± 0.120) & 15.501 \scriptsize(± 0.030) & 19.727 \scriptsize(± 0.185) & \lightgrey 15.499 \scriptsize(± 0.108) \\
    MAE\_FR\_FT & 17.374 \scriptsize(± 0.456) & \underline{12.861 \scriptsize(± 0.256)} & \textbf{9.856 \scriptsize(± 0.462)} & \textbf{16.285 \scriptsize(± 0.333)} & \textbf{9.916 \scriptsize(± 0.155)} & \textbf{9.833 \scriptsize(± 0.599)} & \textbf{11.975 \scriptsize(± 0.202)} & \textbf{14.117 \scriptsize(± 0.438)} & \lightgrey \textbf{12.777 \scriptsize(± 0.363)} \\
    \bottomrule
  \end{tabular}
  }
\vspace{1ex}
\caption{
\textbf{Evaluation of trait prediction with full-range (FR) samples.}
Trait-wise performance (mean ± standard deviation) includes $R^2$ $(\uparrow)$ and nRMSE $(\downarrow)$ metrics. Competing methods are: fully supervised baseline (`Supervised'); MAE with full-range training and linear probing (`MAE\_FR\_LP'); MAE with full-range training and fine-tuning (`MAE\_FR\_FT'); RTM-based autoencoder (`RTM\_AE'); and semi-supervised regression GAN (`SR-GAN'). In RTM-AE, cbc is not directly predicted but is derived from cm and cp estimates (cm – cp). We \textbf{bold} and \underline{underline} best and second best scores respectively. Trait abbreviations are detailed in Sec.~\ref{sec:LbSection}.
}
\label{tab:fr_results}
\vspace{-3mm}
\end{table}

\begin{table}[t!]
  \centering
  \resizebox{\textwidth}{!}{%
  \begin{tabular}{lccccccccc}
    \toprule
    & cab & cw & cm & LAI & cp & cbc & car & anth & \lightgrey average \\
    \midrule
    \multicolumn{10}{c}{$R^2$ $(\uparrow)$} \\
    \midrule
    Sup\_HR & 0.277 {\scriptsize(± 0.105)} & 0.072 {\scriptsize(± 0.032)} & 0.197 {\scriptsize(± 0.082)} & 0.048 {\scriptsize(± 0.110)} & 0.197 {\scriptsize(± 0.080)} & 0.219 {\scriptsize(± 0.074)} & 0.126 {\scriptsize(± 0.135)} & 0.166 {\scriptsize(± 0.052)} & \lightgrey 0.163 {\scriptsize(± 0.084)} \\
    SR-GAN\_HR & 0.496 {\scriptsize(± 0.017)} & 0.336 {\scriptsize(± 0.006)} & 0.356 {\scriptsize(± 0.011)} & 0.428 {\scriptsize(± 0.010)} & 0.371 {\scriptsize(± 0.010)} & 0.381 {\scriptsize(± 0.008)} & 0.455 {\scriptsize(± 0.015)} & \textbf{0.598 {\scriptsize(± 0.020)}} & \lightgrey 0.427 {\scriptsize(± 0.012)} \\
    RTM-AE\_HR & \textbf{0.582 {\scriptsize(± 0.024)}} & \underline{0.450 {\scriptsize(± 0.014)}} & \underline{0.472 {\scriptsize(± 0.031)}} & \underline{0.541 {\scriptsize(± 0.019)}} & \underline{0.546 {\scriptsize(± 0.019)}} & \underline{0.471 {\scriptsize(± 0.043)}} & \underline{0.491 {\scriptsize(± 0.029)}} & 0.538 {\scriptsize(± 0.010)} & \lightgrey \underline{0.511 {\scriptsize(± 0.023)}} \\
    MAE\_FR\_HR\_LP & 0.466 {\scriptsize(± 0.008)} & 0.220 {\scriptsize(± 0.009)} & 0.271 {\scriptsize(± 0.006)} & 0.378 {\scriptsize(± 0.004)} & 0.253 {\scriptsize(± 0.008)} & 0.274 {\scriptsize(± 0.003)} & 0.434 {\scriptsize(± 0.004)} & 0.234 {\scriptsize(± 0.013)} & \lightgrey 0.316 {\scriptsize(± 0.007)} \\
    MAE\_FR\_HR\_FT & \underline{0.578 {\scriptsize(± 0.011)}} & \textbf{0.553 {\scriptsize(± 0.009)}} & \textbf{0.655 {\scriptsize(± 0.012)}} & 0.540 {\scriptsize(± 0.012)} & \textbf{0.612 {\scriptsize(± 0.022)}} & \textbf{0.642 {\scriptsize(± 0.009)}} & \textbf{0.512 {\scriptsize(± 0.018)}} & 0.433 {\scriptsize(± 0.032)} & \lightgrey \textbf{0.566 {\scriptsize(± 0.015)}} \\
    MAE\_HR\_LP & 0.493 {\scriptsize(± 0.004)} & 0.221 {\scriptsize(± 0.014)} & 0.247 {\scriptsize(± 0.007)} & 0.435 {\scriptsize(± 0.003)} & 0.280 {\scriptsize(± 0.006)} & 0.279 {\scriptsize(± 0.003)} & 0.375 {\scriptsize(± 0.010)} & 0.376 {\scriptsize(± 0.019)} & \lightgrey 0.338 {\scriptsize(± 0.008)} \\
    MAE\_HR\_FT & 0.518 {\scriptsize(± 0.038)} & 0.392 {\scriptsize(± 0.045)} & 0.397 {\scriptsize(± 0.028)} & \textbf{0.567 {\scriptsize(± 0.016)}} & 0.478 {\scriptsize(± 0.057)} & 0.402 {\scriptsize(± 0.048)} & 0.418 {\scriptsize(± 0.034)} & \underline{0.547 {\scriptsize(± 0.056)}} & \lightgrey 0.465 {\scriptsize(± 0.040)} \\
    \midrule
    \multicolumn{10}{c}{nRMSE $(\downarrow)$} \\
    \midrule
    Sup\_HR & 21.177 {\scriptsize(± 1.575)} & 20.501 {\scriptsize(± 0.333)} & 16.560 {\scriptsize(± 0.831)} & 25.575 {\scriptsize(± 1.465)} & 15.479 {\scriptsize(± 0.757)} & 16.601 {\scriptsize(± 0.774)} & 18.856 {\scriptsize(± 1.492)} & 20.346 {\scriptsize(± 0.634)} & \lightgrey 19.387 {\scriptsize(± 0.982)} \\
    SR-GAN\_HR & 17.706 {\scriptsize(± 0.297)} & 17.339 {\scriptsize(± 0.073)} & 14.869 {\scriptsize(± 0.131)} & 20.007 {\scriptsize(± 0.175)} & 13.809 {\scriptsize(± 0.108)} & 14.827 {\scriptsize(± 0.092)} & 14.914 {\scriptsize(± 0.211)} & \textbf{14.133 {\scriptsize(± 0.355)}} & \lightgrey 15.950 {\scriptsize(± 0.180)} \\
    RTM-AE\_HR & \underline{16.125 {\scriptsize(± 0.460)}} & \underline{15.772 {\scriptsize(± 0.199)}} & \underline{13.460 {\scriptsize(± 0.395)}} & 17.777 {\scriptsize(± 0.370)} & \textbf{11.725 {\scriptsize(± 0.242)}} & \underline{13.698 {\scriptsize(± 0.558)}} & \underline{14.415 {\scriptsize(± 0.413)}} & 15.155 {\scriptsize(± 0.163)} & \lightgrey \underline{14.766 {\scriptsize(± 0.350)}} \\
    MAE\_FR\_HR\_LP & 17.742 {\scriptsize(± 0.140)} & 17.242 {\scriptsize(± 0.102)} & 16.930 {\scriptsize(± 0.065)} & 19.352 {\scriptsize(± 0.056)} & 18.377 {\scriptsize(± 0.095)} & 17.136 {\scriptsize(± 0.040)} & 15.197 {\scriptsize(± 0.052)} & 19.545 {\scriptsize(± 0.165)} & \lightgrey 17.690 {\scriptsize(± 0.089)} \\
    MAE\_FR\_HR\_FT & \textbf{15.771 {\scriptsize(± 0.202)}} & \textbf{13.055 {\scriptsize(± 0.126)}} & \textbf{11.652 {\scriptsize(± 0.205)}} & \textbf{16.647 {\scriptsize(± 0.213)}} & 13.242 {\scriptsize(± 0.370)} & \textbf{12.037 {\scriptsize(± 0.154)}} & \textbf{14.109 {\scriptsize(± 0.255)}} & 16.805 {\scriptsize(± 0.469)} & \lightgrey \textbf{14.165 {\scriptsize(± 0.249)}} \\
    MAE\_HR\_LP & 17.768 {\scriptsize(± 0.070)} & 18.769 {\scriptsize(± 0.163)} & 16.072 {\scriptsize(± 0.071)} & 19.729 {\scriptsize(± 0.050)} & 14.764 {\scriptsize(± 0.065)} & 15.995 {\scriptsize(± 0.036)} & 15.976 {\scriptsize(± 0.125)} & 17.609 {\scriptsize(± 0.262)} & \lightgrey 17.085 {\scriptsize(± 0.106)} \\
    MAE\_HR\_FT & 17.313 {\scriptsize(± 0.692)} & 16.578 {\scriptsize(± 0.621)} & 14.381 {\scriptsize(± 0.337)} & \underline{17.263 {\scriptsize(± 0.314)}} & \underline{12.561 {\scriptsize(± 0.692)}} & 14.563 {\scriptsize(± 0.589)} & 15.415 {\scriptsize(± 0.453)} & \underline{14.973 {\scriptsize(± 0.948)}} & \lightgrey 15.381 {\scriptsize(± 0.581)} \\
    \bottomrule
  \end{tabular}
  }
  \vspace{1ex}
  \caption{
\textbf{Evaluation of trait prediction with half-range (HR) samples.}
Trait-wise performance (mean ± standard deviation) includes $R^2$ $(\uparrow)$ and nRMSE $(\downarrow)$ metrics. Competing methods are: supervised baseline with HR settings (`Sup\_HR'); semi-supervised SR-GAN with HR settings (`SR-GAN\_HR'); RTM-based autoencoder with HR settings (`RTM-AE\_HR'); MAE pretrained on full-range and fine-tuned with linear probing (`MAE\_FR\_HR\_LP'); MAE pretrained on full-range and fine-tuned (`MAE\_FR\_HR\_FT'); MAE pretrained on HR and fine-tuned with linear probing (`MAE\_HR\_LP'); and MAE pretrained on HR and fine-tuned (`MAE\_HR\_FT'). In RTM-AE, cbc is not directly predicted but is derived from cm and cp estimates (cm – cp). We \textbf{bold} and \underline{underline} best and second best scores respectively. Trait abbreviations are detailed in Sec.~\ref{sec:LbSection}.
  }
  \label{tab:hr_results}
  \vspace{-3mm}
\end{table}

\paragraph{Full and half range spectra analyses.} Trait-specific results, reported with $R^2$ and nRMSE scores, are summarized in Tables~\ref{tab:fr_results} and~\ref{tab:hr_results} for the full- and half-range experiments, respectively. 
Among all competing methods, the fine-tuned MAE (MAE-FR-FT) outperformed all other methods on most traits when trained and tested on full-range spectra, recording the highest $R^2$ values and lowest nRMSE scores.
Compared to the fully supervised baseline, MAE-FR-FT led to an average improvement of 9\% in $R^2$ and 6\% in nRMSE. 
These results underscore the effectiveness of MAEs in learning meaningful spectral representations through masked spectral reconstruction.
Pretrained MAE models also exhibited strong cross-spectral generalization, performing competitively on half-range data even when pretrained on full-range spectra and applied to half-range data (MAE-FR-HR-FT). 
It indicates a good feature transferability and adaptability across heterogeneous sensor configurations, particularly valuable for operational deployment with multi-source data streams.

The RTM-AE model, which introduces physical interpretability into the learned latent space, underperformed compared to MAE but consistently achieved the second best results for both full- and half-range experiments.
It demonstrates that aligning latent representations with RTMs to enforce physically-constrained embeddings yields promising performance while simultaneously enhancing model explainability through semantically meaningful feature disentanglement and physics-informed representation learning.

To further assess robustness of the approaches under sensor noise such as illumination differences, or sensor-specific signal-to-noise characteristics, we additionally evaluated models’ performances under additive Gaussian noise at inference time (details in Tables \ref{tab:noise_results_sup}-\ref{tab:mae_ft_noise_results}). Zero-mean noise with standard deviations of 0.01, 0.03, and 0.05 was added across all spectral bands. Results show that MAE-FR-FT and RTM-AE are substantially more robust than the supervised baseline and GAN. For instance, at $\sigma = 0.05$, MAE-FR-FT retains an $R^2$ of 0.331 compared to the baseline dropping to –0.065, and both MAE-FR-FT and RTM-AE exhibit smaller increases in nRMSE under spectral corruption, highlighting their resilience.


In the half-range setting, semi- and self-supervised methods also clearly outperformed the supervised baseline. Gains ranged between 100–200\% in $R^2$ and 8–27\% in nRMSE, reinforcing the value of leveraging spectral variability from GreenHyperSpectra even under reduced spectral coverage (Figures ~\ref{fig:heat_r2} and ~\ref{fig:heat_nrmse}).

\begin{table}[t!]
  \centering
  \resizebox{\textwidth}{!}{%
  \begin{tabular}{lccccccccc}
    \toprule
    & cab & cw & cm & LAI & cp & cbc & car & anth & \lightgrey average \\
    \midrule
    \multicolumn{10}{c}{$R^2$ $(\uparrow)$} \\
    \midrule
    Supervised & \textbf{0.362 {\scriptsize(± 0.048)}} & 0.193 {\scriptsize(± 0.053)} & 0.446 {\scriptsize(± 0.049)} & 0.074 {\scriptsize(± 0.031)} & 0.183 {\scriptsize(± 0.041)} & 0.449 {\scriptsize(± 0.052)} & 0.181 {\scriptsize(± 0.045)} & 0.055 {\scriptsize(± 0.079)} & \lightgrey 0.243 {\scriptsize(± 0.050)} \\
    SR\_GAN & \underline{0.300 {\scriptsize(± 0.023)}} & \textbf{0.350 {\scriptsize(± 0.032)}} & \underline{0.507 {\scriptsize(± 0.029)}} & -0.199 {\scriptsize(± 0.111)} & \underline{0.273 {\scriptsize(± 0.037)}} & \underline{0.548 {\scriptsize(± 0.026)}} & 0.221 {\scriptsize(± 0.064)} & \textbf{0.197 {\scriptsize(± 0.175)}} & \lightgrey \underline{0.275 {\scriptsize(± 0.062)}} \\
    RTM\_AE & 0.272 {\scriptsize(± 0.033)} & 0.193 {\scriptsize(± 0.096)} & 0.453 {\scriptsize(± 0.067)} & 0.019 {\scriptsize(± 0.054)} & 0.192 {\scriptsize(± 0.056)} & -0.075 {\scriptsize(± 0.008)} & \textbf{0.266 {\scriptsize(± 0.056)}} & 0.067 {\scriptsize(± 0.252)} & \lightgrey 0.173 {\scriptsize(± 0.078)} \\
    MAE\_FR\_LP & 0.116 {\scriptsize(± 0.028)} & \underline{0.298 {\scriptsize(± 0.028)}} & 0.442 {\scriptsize(± 0.039)} & \underline{0.182 {\scriptsize(± 0.059)}} & 0.211 {\scriptsize(± 0.032)} & 0.478 {\scriptsize(± 0.044)} & \underline{0.232 {\scriptsize(± 0.020)}} & \underline{0.142 {\scriptsize(± 0.153)}} & \lightgrey 0.263 {\scriptsize(± 0.050)} \\
    MAE\_FR\_FT & 0.271 {\scriptsize(± 0.030)} & 0.28 {\scriptsize(± 0.102)} & \textbf{0.575 {\scriptsize(± 0.041)}} & \textbf{0.229 {\scriptsize(± 0.041)}} & \textbf{0.275 {\scriptsize(± 0.068)}} & \textbf{0.582 {\scriptsize(± 0.044)}} & 0.165 {\scriptsize(± 0.044)} & 0.112 {\scriptsize(± 0.234)} & \lightgrey \textbf{0.311 {\scriptsize(± 0.076)}} \\
        \midrule
    \multicolumn{10}{c}{nRMSE $(\downarrow)$} \\
    \midrule
    Supervised & \textbf{19.173 {\scriptsize(± 0.695)}} & 25.223 {\scriptsize(± 13.109)} & 14.238 {\scriptsize(± 0.496)} & 22.984 {\scriptsize(± 0.475)} & 17.072 {\scriptsize(± 0.669)} & 14.818 {\scriptsize(± 0.560)} & 19.185 {\scriptsize(± 0.581)} & 23.159 {\scriptsize(± 1.644)} & \lightgrey 19.482 {\scriptsize(± 2.278)}\\
    SR\_GAN & \underline{20.098 {\scriptsize(± 0.373)}} & \textbf{22.394 {\scriptsize(± 10.712)}} & \underline{13.445 {\scriptsize(± 0.404)}} & 26.075 {\scriptsize(± 1.018)} & \underline{16.108 {\scriptsize(± 0.662)}} & \underline{{13.438 {\scriptsize(± 0.578)}}} & 18.698 {\scriptsize(± 0.568)} & \textbf{21.360 {\scriptsize(± 3.596)}} & \lightgrey \underline{18.952 {\scriptsize(± 2.239)}} \\
    RTM-AE & 20.493 {\scriptsize(± 0.443)} & 25.137 {\scriptsize(± 13.001)} & 14.138 {\scriptsize(± 0.616)} & 23.652 {\scriptsize(± 0.668)} & 16.978 {\scriptsize(± 0.802)} & 20.742 {\scriptsize(± 0.741)} & \textbf{18.155 {\scriptsize(± 0.670)}} & 22.874 {\scriptsize(± 3.780)} & \lightgrey 20.271 {\scriptsize(± 2.590)} \\
    MAE\_FR\_LP & 22.589 {\scriptsize(± 0.406)} & \underline{23.406 {\scriptsize(± 11.682)}} & 14.296 {\scriptsize(± 0.474)} & \underline{21.585 {\scriptsize(± 0.763)}} & 16.781 {\scriptsize(± 0.677)} & 14.440 {\scriptsize(± 0.641)} & \underline{18.582 {\scriptsize(± 0.322)}} & \underline{22.147 {\scriptsize(± 3.725)}} & \lightgrey 19.228 {\scriptsize(± 2.336)} \\
    MAE\_FR\_FT & 20.505 {\scriptsize(± 0.334)} & 24.018 {\scriptsize(± 13.502)} & \textbf{12.466 {\scriptsize(± 0.425)}} & \textbf{20.842 {\scriptsize(± 0.548)}} & \textbf{16.069 {\scriptsize(± 0.860)}} & \textbf{12.907 {\scriptsize(± 0.426)}} & 19.371 {\scriptsize(± 0.414)} & 22.422 {\scriptsize(± 4.166)} & \lightgrey \textbf{18.575 {\scriptsize(± 2.584)}} \\
    \bottomrule
  \end{tabular}
  }
\vspace{1ex}
\caption{\textbf{Cross-dataset generalization performance.}
Models are trained on labeled data from all but five datasets (see Sec~\ref{splitSection}), and evaluated on held-out datasets to assess OOD generalization. Trait-wise performance includes $R^2$ $(\uparrow)$ and nRMSE $(\downarrow)$ metrics. In RTM-AE, cbc is not directly predicted but is derived from cm and cp estimates (cm – cp). We \textbf{bold} and \underline{underline} best and second best scores respectively.
}
\label{tab:ood_results}
\vspace{-7mm}
\end{table}

\paragraph{OOD evaluation.}
As shown in Table~\ref{tab:ood_results}, the fine-tuned MAE (MAE-FR-FT) had the highest performance over all other methods across traits, achieving a slight improvement in $R^2$ relative to the supervised baseline ($0.31$ vs. $0.24$), along with the lowest average nRMSE. 
Since many traits are not associated to single-band features but instead arise from complex interactions across multiple regions of the spectrum, MAE provides a strong prior: it enforces the learning of localized correlations and long-range dependencies within hyperspectral signals, by reconstructing both across adjacent tokens and distant tokens. This prior knowledge facilitates better generalization and more efficient fine-tuning with MAE-FR-FT for the downstream regression. However, this prior alone is not sufficient. When only linear probing is applied (MAE-FR-LP), the model retains general spectral trends leading to underperformance. The necessity of fine-tuning becomes evident in our feature attribution analysis (Fig.~\ref{fig:featureXai_MAE}), where we compared gradient amplitudes across spectral bands for MAE-FR-LP, last block fine-tuning, and full fine-tuning MAE-FR-FT models. While MAE-FR-LP exhibited diffuse and noisy attributions across broad spectral regions, fine-tuning progressively reduced gradient variance, yielding sharper and more interpretable feature importance profiles. This indicates that fine-tuning allows the pretrained prior to be refined toward trait-relevant spectral dependencies, transforming general correlations into targeted representations that drive improved predictive performance.

Other competing methods, such as SR-GAN and RTM-AE, provided modest gains over the supervised baseline. 
The corresponding scatter plot of the observed and predicted trait values from the different methods is presented in Fig.~\ref{fig:scatter} in the Appendix.

\section{Conclusions and perspectives}
\label{sec:conclusions}
In this study, we introduce \textbf{GreenHyperSpectra}, a large-scale cross-sensor and cross-ecosystem spectral dataset designed to train machine learning models for plant trait prediction from hyperspectral data. Leveraging GreenHyperSpectra as a pretraining resource, we demonstrated that models using MAE consistently outperformed all other benchmarked methods, including the fully supervised baseline, across a variety of settings. The adaptability of MAE models enables their application to multi-scale remote sensing platforms, including drone, airborne, and satellite imagery, paving the way to investigate how learned spectral features, derived from a heterogeneous spectral dataset, generalize across varying spatial resolutions. MAEs also serve as a strong foundation for advanced transfer learning architectures aimed at improving predictive performance.
While we explore default sensor configurations (VNIR+SWIR and VNIR), extending pretrained encoders to other spectral ranges remains open for future work.
The MAE architecture shows promise for cross-domain adaptation across heterogeneous sensing modalities through fine-tuning strategies. 
We contribute towards global pretraining datasets for spectral embeddings while highlighting critical biases affecting generalization. 
Despite improvements, run-to-run variance reveals challenges in learning stable representations from various ecological data distributions. 
Future research should expand multi-domain spectral datasets across biomes and sensing conditions to enhance transferability and address geographical and ecosystem-level biases in annotated data. 
Nevertheless, our pretrained models from GreenHyperSpectra will remain valuable as labeled data from underrepresented regions increases. This study confirms that semi- and self-supervised methods with large-scale pretraining are essential for advancing ecosystem monitoring.

\paragraph{{\textbf{Acknowledgments}}}
We thank all data owners for sharing the data either by request or through the public Ecological Spectral Information System (EcoSIS), Data Publisher for Earth and Environmental Science (PANGEA) and DRYAD platforms. 
EC, AO and DR acknowledge support for this work from IVADO and the Canada CIFAR AI Chairs program, and computational support from Mila – Quebec AI Institute, including in-kind support from Nvidia Corporation. EC and HF acknowledge the financial support by the Federal Ministry of Education and Research of Germany and by the Sächsische Staatsministerium für Wissenschaft Kultur und Tourismus in the program Center of Excellence for AI-research "Center for Scalable Data Analytics and Artificial Intelligence Dresden/Leipzig", project identification number: ScaDS.AI. 

\bibliographystyle{abbrv}
\bibliography{bib}

\medskip


\newpage
\section*{NeurIPS Paper Checklist}

\begin{enumerate}

\item {\bf Claims}
    \item[] Question: Do the main claims made in the abstract and introduction accurately reflect the paper's contributions and scope?
    \item[] Answer: \answerYes{} 
    \item[] Justification: we summarize four key contributions in Section \ref{sec:intro}, which are addressed through our experimental design and results.
    
\item {\bf Limitations}
    \item[] Question: Does the paper discuss the limitations of the work performed by the authors?
    \item[] Answer: \answerYes{} 
    \item[] Justification: See Section ~\ref{sec:conclusions}

\item {\bf Theory assumptions and proofs}
    \item[] Question: For each theoretical result, does the paper provide the full set of assumptions and a complete (and correct) proof?
    \item[] Answer: \answerNA{} 
    \item[] Justification: This work does not include theoretical analysis.

    \item {\bf Experimental result reproducibility}
    \item[] Question: Does the paper fully disclose all the information needed to reproduce the main experimental results of the paper to the extent that it affects the main claims and/or conclusions of the paper (regardless of whether the code and data are provided or not)?
    \item[] Answer: \answerYes{} 
    \item[] Justification: We share the data and code and provide more details on the experiments in Appendix \ref{sec:model_details}.

\item {\bf Open access to data and code}
    \item[] Question: Does the paper provide open access to the data and code, with sufficient instructions to faithfully reproduce the main experimental results, as described in supplemental material?
    \item[] Answer: \answerYes{} 
    \item[] Justification: We submit the code and data URL on the OpenReview platform.

\item {\bf Experimental setting/details}
    \item[] Question: Does the paper specify all the training and test details (e.g., data splits, hyperparameters, how they were chosen, type of optimizer, etc.) necessary to understand the results?
    \item[] Answer: \answerYes{} 
    \item[] Justification: Key training details, including model architecture, loss functions, data splits and major modifications, are summarized in the main text. Comprehensive descriptions of training procedures and hyperparameters are provided in Appendix ~\ref{sec:model_details}.
    

\item {\bf Experiment statistical significance}
    \item[] Question: Does the paper report error bars suitably and correctly defined or other appropriate information about the statistical significance of the experiments?
    \item[] Answer: \answerYes{} 
    \item[] Justification: 
    We evaluate all models using three random seeds and report the mean and standard deviation of the performance metrics (\ie $R^2$, nRMSE) to reflect variability due to random initialization. This is stated in the main text and reflected in relevant figures and tables (\eg Figures~\ref{fig:res_unlabeled} and ~\ref{fig:res_labeled}). For the cross-dataset OOD evaluation, we report results from a single run due to computational constraints; however, we attempt to account for variability through controlled random sampling, as detailed in Section~\ref{sec:expe_settings}.

\item {\bf Experiments compute resources}
    \item[] Question: For each experiment, does the paper provide sufficient information on the computer resources (type of compute workers, memory, time of execution) needed to reproduce the experiments?
    \item[] Answer: \answerYes{} 
    \item[] Justification: We provide detailed information on compute resources in Appendix~\ref{sec:model_details}, including the type of GPUs, model-specific runtime estimates, and the number of trainable parameters per method (see Table~\ref{tab:resReq}).
    

\item {\bf Code of ethics}
    \item[] Question: Does the research conducted in the paper conform, in every respect, with the NeurIPS Code of Ethics \url{https://neurips.cc/public/EthicsGuidelines}?
    \item[] Answer: \answerYes{} 
    \item[] Justification: We have reviewed the NeurIPS ethical guidelines and ensured that the paper complies with them.

\item {\bf Broader impacts}
    \item[] Question: Does the paper discuss both potential positive societal impacts and negative societal impacts of the work performed?
    \item[] Answer: \answerYes{} 
    \item[] Justification: We discuss the positive societal impacts of our work in Section ~\ref{sec:intro} and ~\ref{sec:conclusions}. While we did not identify any clear negative impacts, we remain open to addressing potential concerns if raised during the review process.

\item {\bf Safeguards}
    \item[] Question: Does the paper describe safeguards that have been put in place for responsible release of data or models that have a high risk for misuse (e.g., pretrained language models, image generators, or scraped datasets)?
    \item[] Answer: \answerNA{} 
    \item[] Justification: The data and models presented in this paper do not pose risks of misuse. All datasets are derived from publicly available, non-sensitive sources, and the models are focused on ecological applications.

\item {\bf Licenses for existing assets}
    \item[] Question: Are the creators or original owners of assets (e.g., code, data, models), used in the paper, properly credited and are the license and terms of use explicitly mentioned and properly respected?
    \item[] Answer: \answerYes{} 
    \item[] Justification: All datasets used in this work are publicly available and properly credited in the paper. For each source included in GreenHyperSpectra, we provide citation information, access links in Table~\ref{tab:datasets}.

\item {\bf New assets}
    \item[] Question: Are new assets introduced in the paper well documented and is the documentation provided alongside the assets?
    \item[] Answer: \answerYes{} 
    \item[] Justification: The GreenHyperSpectra dataset introduced in this paper is documented in detail in Appendix~\ref{sec:datasetinfo}. The appendix includes information about data sources, sensor specifications, geographic and temporal coverage, preprocessing steps, and licensing terms for each included dataset. The asset is composed entirely of publicly available data, contains no personal or sensitive information, and respects the original licenses of the sources.

\item {\bf Crowdsourcing and research with human subjects}
    \item[] Question: For crowdsourcing experiments and research with human subjects, does the paper include the full text of instructions given to participants and screenshots, if applicable, as well as details about compensation (if any)? 
    \item[] Answer: \answerNA{} 
    \item[] Justification: The paper does not involve any crowdsourcing or research with human subjects.

\item {\bf Institutional review board (IRB) approvals or equivalent for research with human subjects}
    \item[] Question: Does the paper describe potential risks incurred by study participants, whether such risks were disclosed to the subjects, and whether Institutional Review Board (IRB) approvals (or an equivalent approval/review based on the requirements of your country or institution) were obtained?
    \item[] Answer: \answerNA{} 
    \item[] Justification: The paper does not involve research with human subjects.

\item {\bf Declaration of LLM usage}
    \item[] Question: Does the paper describe the usage of LLMs if it is an important, original, or non-standard component of the core methods in this research? Note that if the LLM is used only for writing, editing, or formatting purposes and does not impact the core methodology, scientific rigorousness, or originality of the research, declaration is not required.
    \item[] Answer: \answerNA{} 
    \item[] Justification: The core methods and experiments presented in this paper do not involve the use of large language models (LLMs).

\end{enumerate}

\newpage
\appendix

\textbf{Table of figures:}
\begin{table}[h!]
\centering
\begin{tabular}{|c|p{12cm}|}
\hline
\textbf{Figure} & \textbf{Description} \\ \hline
\ref{fig:teaser} & Teaser illustration of the proposed semi-/self-supervised frameworks for multi-trait regression. \\ \hline
\ref{fig:coverage} & Comparison of GreenHyperSpectra and the labeled dataset, highlighting broader coverage in vegetation types and sensor diversity. \\ \hline
\ref{fig:archi_gan}, \ref{fig:archi_ae_rtm} and \ref{fig:archi_mae} & Overview of the SR-GAN, RTM-AE, and MAE architectures for trait prediction. \\ \hline
\ref{fig:res_labeled} & Effect of increasing labeled data volume on R\textsuperscript{2} performance. \\ \hline
\ref{fig:res_unlabeled} & Effect of increasing unlabeled pretraining data volume on R\textsuperscript{2} performance. \\ \hline
\ref{fig:spactral_range}, \ref{fig:histogram_data},\ref{fig:tsne_spectra} and \ref{fig:Appendix_subsamopUnl} & Dataset characteristics, focusing on spectral variability. \\ \hline
\ref{fig:figRes2_Labelled_n} and \ref{fig:figRes2_unLabelled_n} & Complementary results to Figs.~\ref{fig:res_labeled} and ~\ref{fig:res_unlabeled}: nRMSE performance trends with increasing labeled and unlabeled data. \\ \hline
\ref{fig:heat_r2} and \ref{fig:heat_nrmse} & Complementary results to Table~\ref{tab:hr_results}: Heatmaps of R\textsuperscript{2} and nRMSE across traits in the half-range input settings to show the performance of MAE vs baseline. \\ \hline
\ref{fig:scatter} & Complementary results to Table~\ref{tab:ood_results}: Observed vs. predicted plots, showing trait-wise calibration in the OOD setting. \\ \hline
\ref{fig:featureXai_MAE} & Feature importance of MAE-based downstream regression as a function of fine-tuning depth (linear probing, final block, and full fine-tuning). \\ \hline
\end{tabular}
\caption{Summary of figures and their descriptions.}
\label{tab:figures}
\end{table}

\textbf{Table of tables:}
\begin{table}[h!]
\centering
\begin{tabular}{|c|p{12cm}|}
\hline
\textbf{Table} & \textbf{Description} \\ \hline
\ref{tab:instrument-specs} & Summary of sensor and platform specifications in GreenHyperSpectra. \\ \hline
\ref{tab:fr_results} & Trait-wise performance (R\textsuperscript{2} and nRMSE) of all models under full-range input settings. \\ \hline
\ref{tab:hr_results} & Trait-wise performance (R\textsuperscript{2} and nRMSE) of all models under half-range (VNIR) input settings. \\ \hline
\ref{tab:ood_results} & Trait-wise performance (R\textsuperscript{2} and nRMSE) of all models under out-of-distribution (OOD) settings. \\ \hline
\ref{tab:datasets} and \ref{tab:trait_stats} & Dataset details: spectral data characteristics and trait distribution across sources. \\ \hline
\ref{tab:gan_architecture} and \ref{tab:gan_hyperparams} & Architecture and hyperparameters of SR-GAN. \\ \hline
\ref{tab:ae_rtm_architecture}, \ref{tab:aertm_hyperparams} and \ref{tab:aertm_RTM} & Architecture, hyperparameters, and RTM configuration of RTM-AE. \\ \hline
\ref{tab:mae_architecture} and \ref{tab:mae_hyperparams} & Architecture and hyperparameters of MAE. \\ \hline
\ref{tab:mae_ablation}, \ref{tab:mae_loss_ablation} and \ref{tab:mae_patchsize_ablation} & MAE ablation studies: effects of transformer depth, loss weighting, and token size on trait prediction (R\textsuperscript{2} and nRMSE). \\ \hline
\ref{tab:resReq} & Model size, runtime, and GPU usage across methods. \\ \hline
\ref{tab:tundra_results}, \ref{tab:forest_results}, \ref{tab:crops_results}, \ref{tab:shrubland_results}, \ref{tab:grassland_results} and \ref{tab:mix_results} & Complement to Table~\ref{tab:ood_results} and Fig.~\ref{fig:scatter}: OOD model performance when one vegetation class is excluded from the test set. \\ \hline
\ref{tab:noise_results_sup}, \ref{tab:gan_noise_results}, \ref{tab:aertm_noise_results}, \ref{tab:mae_lp_noise_results} and \ref{tab:mae_ft_noise_results} & Robustness evaluation under additive Gaussian noise during inference, reported for all models (R\textsuperscript{2} and nRMSE). \\ \hline
\end{tabular}
\caption{Summary of tables and their descriptions.}
\label{tab:tables}
\end{table}

\clearpage
\section{Details about the datasets}
\vspace*{-6em}
\label{sec:datasetinfo}
The data are publicly available \href{https://huggingface.co/datasets/Avatarr05/GreenHyperSpectra}{here}.
\paragraph{Spectral preprocessing.}
For standardized cross-instrument comparison, all reflectance spectra were resampled to a uniform wavelength grid spanning the 400-2500 nm solar-reflective range.
Spectral measurements were linearly interpolated to an interval of 1 nm, resulting in 2101 bands per sample. Regions of strong atmospheric water absorption, specifically 1351–1430 nm, 1801–2050 nm, and 2451–2500 nm, were removed to minimize noise and signal loss. The remaining bands were smoothed using a Savitzky-Golay filter with a 65 nm window \citep{savitzky1964smoothing}. 
After these steps, 1721 spectral bands were retained for analysis, providing a high-quality input space for training and evaluation.
\vspace*{-5em} 
\begin{figure}[H]
  \centering
  \vspace{-2ex} 
    \includegraphics[width=\linewidth, height=\textheight, keepaspectratio]{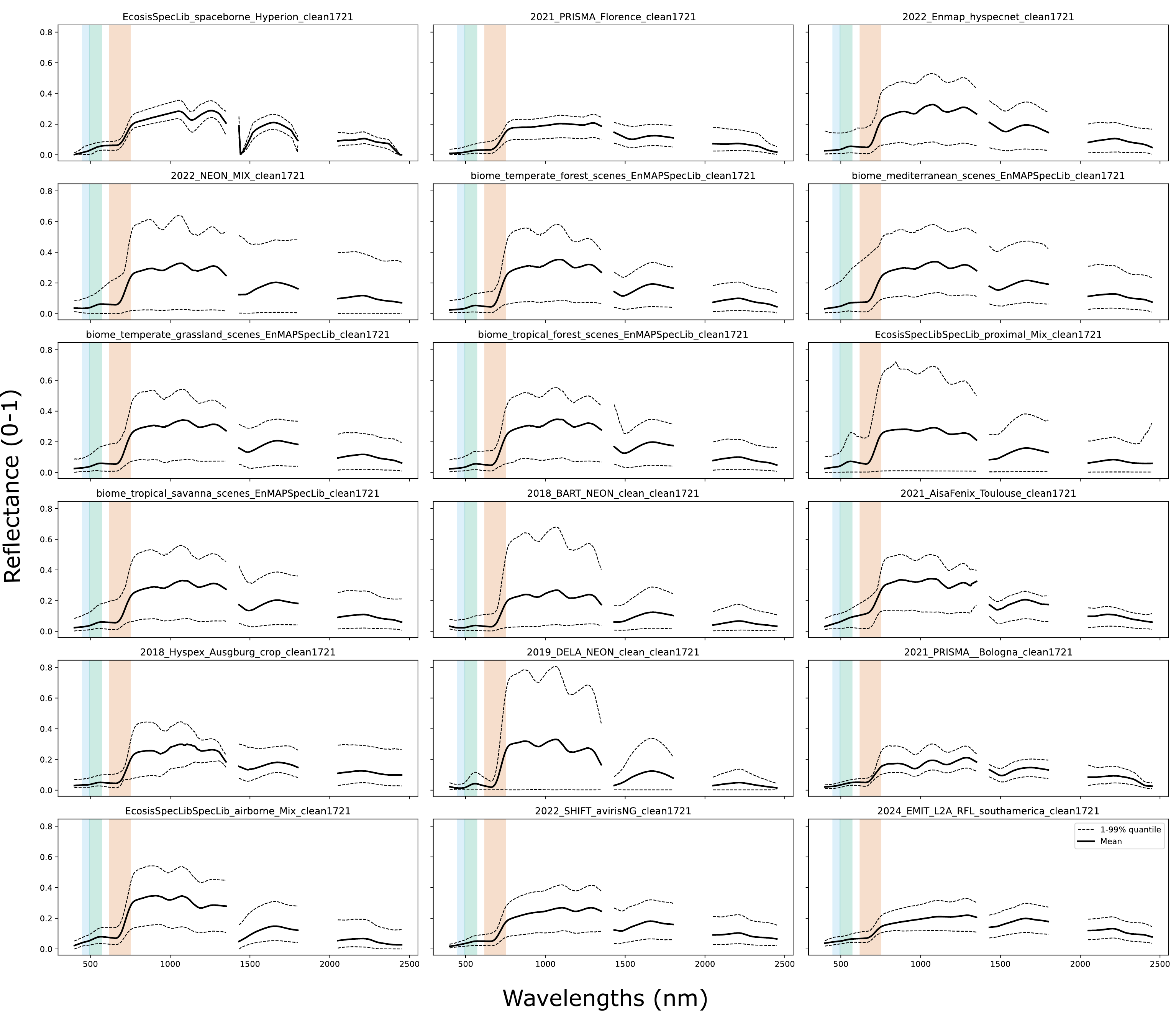}%
  \vspace{1ex} 
  \caption{\textbf{Spectral reflectance across wavelengths.} This plot shows the variation in canopy reflectance within GreenHyperSpectra across different data sources, highlighting differences due to acquisition conditions and sensor modalities. The colored ranges refer to the visible region.}
  \label{fig:spactral_range}
\end{figure}

\begin{figure}[H]
  \centering
    \includegraphics[width=\linewidth, height=0.5\textheight, keepaspectratio]{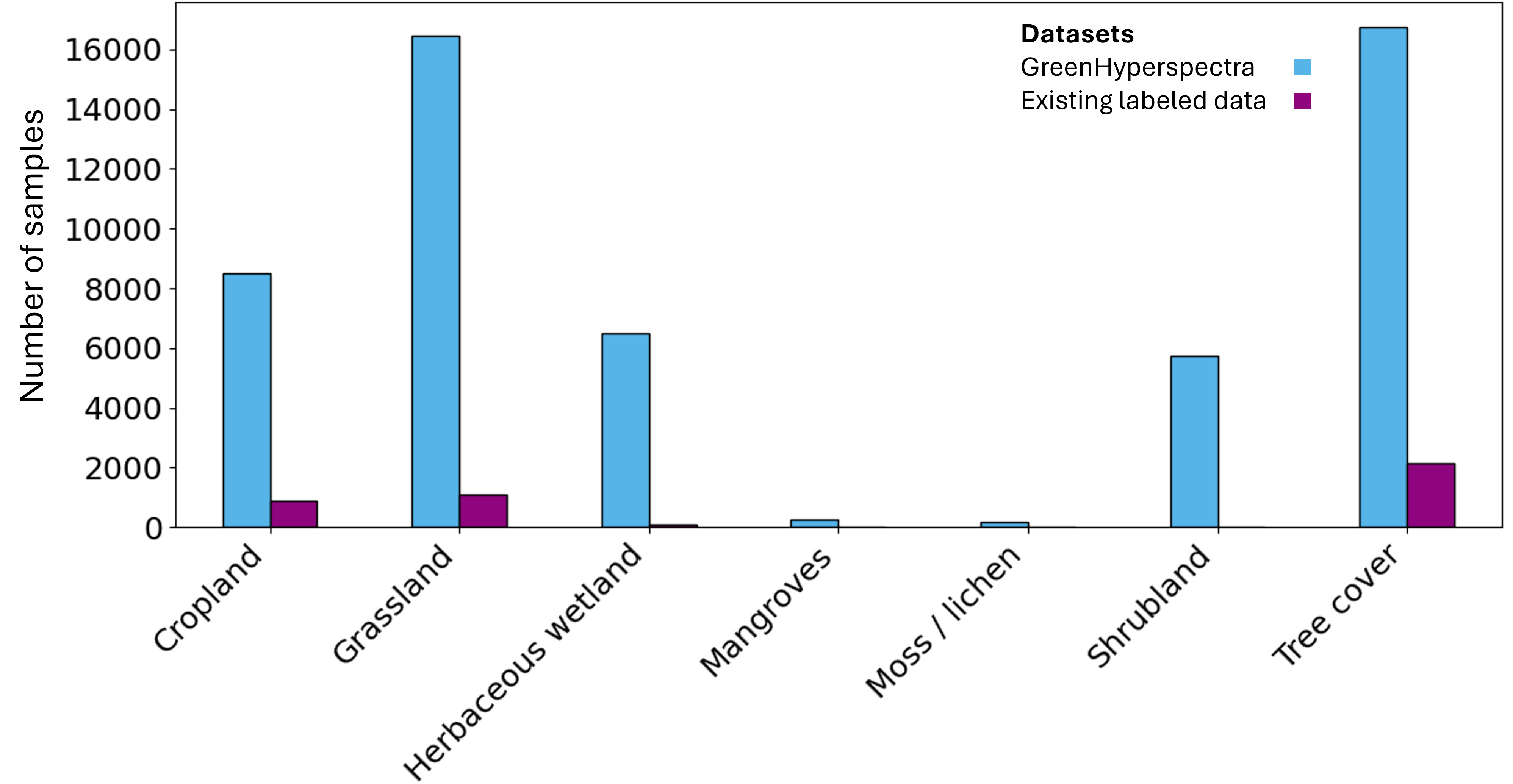}%
  \vspace{1ex} 

  \caption{\textbf{Sample distribution across vegetation types in GreenHyperSpectra.} The plot shows the number of samples in GreenHyperSpectra (blue) and the existing labeled (purple) for each vegetation type, highlighting class imbalance and the relative scarcity of labeled data in certain categories. The vegetation type information was retried from the \href{https://esa-worldcover.org/en}{ESA WORLDCOVER} product}
  \label{fig:histogram_data}
\end{figure}

\begin{table}[H]
\vspace{-1cm}
\centering
\tiny
\setlength{\tabcolsep}{2pt} 
\begin{tabularx}{\textwidth}{l l | l | l r r | l r | c c l}
\toprule

\textbf{Dataset} & \textbf{Platform} & \textbf{Sensor} & \textbf{GSD} & \textbf{\#Bands} & \textbf{Range (nm)} & \textbf{Year} & \textbf{\#Samples} & \textbf{Processing} & \textbf{Land Cover} & \textbf{Source} \\
\midrule

\multirow{2}{*}{DB1\citep{Dennison2018RangeCreek}} &  \multirow{2}{*}{proximal} &  \multirow{2}{*}{ASD FieldSpec Pro} &  \multirow{2}{*}{N/A} &  \multirow{2}{*}{2151} &  \multirow{2}{*}{350--2500} &  2007  & 31 &  \multirow{2}{*}{Reflectance spectra}  & \multirow{2}{*}{desert, shrubland} & \href{https://ecosis.org/package/range-creek-utah-species-spectra}{Link} \\
\cline{7-8}
 &   &   &   &   &   &   2009  & 49 &    &   &   \\
 \midrule
 
\multirow{3}{*}{DB2\citep{DennisonGardner2018Hawaii}} & \multirow{3}{*}{proximal} & \multirow{3}{*}{ASD FieldSpec FR} & \multirow{3}{*}{N/A} & \multirow{3}{*}{1063} & \multirow{3}{*}{352--2476} &  \multirow{3}{*}{2000}  & \multirow{3}{*}{792} &  \multirow{3}{*}{Reflectance spectra}  & \multirowcell{3}{forest, shrubland: \\ native-dominated \\ Hawaiian forest types} & \multirow{3}{*}{\href{https://data.ecosis.org/dataset/hawaii-2000-vegetation-species-spectra}{Link}} \\
& & & & & & & \\
& & & & & & & \\
 \midrule
 
\multirow{4}{*}{DB3\citep{DennisonRoberts2018SantaMonica}} & \multirow{4}{*}{proximal} & \multirow{4}{*}{ASD FieldSpec FR} & \multirow{4}{*}{N/A} & \multirow{4}{*}{1075} & \multirow{4}{*}{350--2498} & 1995  & 226 &  \multirow{4}{*}{Reflectance spectra}  & \multirow{4}{*}{shrubland} & \multirow{4}{*}{\href{https://data.ecosis.org/dataset/santa-monica-mountains-vegetation-species-spectra}{Link}} \\
\cline{7-8}
 &  &  &  &  &  & 1996  & 93 &    &  \\
 \cline{7-8}
 &  &  &  &  &  & 1997  & 132 &    &  \\
 \cline{7-8}
 &  &  &  &  &  & 1998  & 10 &    &  \\
 \midrule

\multirow{4}{*}{DB4\citep{Serbin2019UWBNL}} & \multirow{4}{*}{proximal} & SpecEvo PSM3500 & \multirow{4}{*}{N/A} & \multirow{4}{*}{2151} & \multirow{4}{*}{350--2500} &  2013  & 6 &  \multirow{4}{*}{Reflectance spectra}  & \multirowcell{4}{Forest, Gra4Sand, \\ Shrubland, Crops} & \multirow{4}{*}{\href{https://data.ecosis.org/dataset/uw-bnl-nasa-hyspiri-airborne-campaign-leaf-and-canopy-spectra-and-trait-data}{Link}} \\
 \cline{3-3}
 \cline{7-8}
 &  & ASD Fieldspec 3 &  &  &  & 2013  & 7   &   &   \\
\cline{3-3}
 \cline{7-8}
 &  & ASD Fieldspec 4 &  &  &  & 2013  & 49   &   &   \\
\cline{3-3}
 \cline{7-8}
 &  & SpecEvo\_PSM3500 &  &  &  & 2014  & 60   &   &   \\
\midrule

DB5\citep{Zesati2019ABoVE} & proximal & SVC HR-1024i & 1 m & 2178 & 338--2516 & 2018  & 112 &  Reflectance spectra  & tundra & \href{https://daac.ornl.gov/cgi-bin/dsviewer.pl?ds_id=1685}{Link} \\
\midrule

DB6\citep{Thompson2021EMITReflectance} & proximal & ASD FieldSpec 3 & N/A & 2151 & 350--2500 &  2001  & 112 &  Reflectance spectra  & urban vegetation & \href{https://ecosis.org/package/emit-manually-adjusted-surface-reflectance-spectra}{Link} \\
\midrule

\multirow{2}{*}{DB7\citep{Dennison2019FractionalCover}} & \multirow{2}{*}{proximal} & \multirow{2}{*}{ASD FieldSpec Pro} & \multirow{2}{*}{N/A} & \multirow{2}{*}{210} & \multirow{2}{*}{400--2490} &    & \multirow{2}{*}{715} & \multirow{2}{*}{Reflectance spectra}  & \multirowcell{2}{crops, gra4Sand, \\ forest, shrubland} & \href{https://ecosis.org/package/fractional-cover-simulated-vswir-dataset-version-2}{Link} \\
& & & & & & & & & & \\
\midrule

DB8\citep{Herold2004UrbanSB} & proximal & ASD FieldSpec 3 & N/A & 1075 & 350--2498 &  2001  & 37 &  Reflectance spectra  & urban vegetation & \href{https://ecosis.org/package/urban-reflectance-spectra-from-santa-barbara--ca}{Link} \\
\midrule

\multirow{2}{*}{DB9\citep{Kokaly2017USGSSpectralLib}} & \multirow{2}{*}{proximal} & \multirowcell{2}{ASD FieldSpec \\ 4 Hi-Res NG} & \multirow{2}{*}{N/A} & \multirow{2}{*}{2151} & \multirow{2}{*}{350--2500} &    & \multirow{2}{*}{87} & \multirow{2}{*}{Reflectance spectra}  & \multirow{2}{*}{--} & \multirow{2}{*}{\href{https://www.sciencebase.gov/catalog/item/5807a2a2e4b0841e59e3a18d}{Link}} \\
& & & & & & & & & & \\
\midrule

DB10\citep{Kokaly2017USGSSpectralLib} & proximal & Beckman 5270 & N/A & 480 & 205--2976 &    & 19 &  Reflectance spectra  & -- & \href{https://www.sciencebase.gov/catalog/item/5807a2a2e4b0841e59e3a18d}{Link} \\
\midrule

DB11\citep{Fang2023AquaticVeg} & proximal & SVC HR-1024i & 40 cm & 995 & 346--2499 &    & 133 &  Reflectance spectra  & aquatic vegetation & \href{https://ecosis.org/package/the-hyperspectra-dataset-for-typical-aquatic-vegetation}{Link} \\
\midrule

DB12\citep{Unger2021ArcticMoss} & proximal & SVC HR-1024i & 8 cm & 994 & 338--2515 &    & 34 &  Reflectance spectra  & tundra & \href{https://ecosis.org/package/arctic-moss-spectral-reflectance-desiccation-experiment}{Link} \\
\midrule

DB13\citep{Serbin2015NGEE} & proximal & SVC HR-1024i & N/A & 2150 & 350--2500 &  2015  & 44 &   Reflectance spectra & coastal, wetland & \href{https://ecosis.org/package/ngee-arctic-hr1024i-canopy-spectral-reflectance}{Link} \\
\midrule

DB14\citep{Roupioz2023CAMCATT} & proximal & ASD & N/A & 2101 & 400--2500 & 2021  & 45 &  Reflectance spectra & urban vegetation  & \href{https://camcatt.sedoo.fr/camcatt-product}{Link} \\
\midrule

DB15\citep{Dennison2018RushValley} & proximal & ADS FieldSpec Pro & N/A & 2151 & 350--2500 &  2005  & 82 &  Reflectance spectra  & shrubland, steppe & \href{https://ecosis.org/package/rush-valley-utah-sagebrush-time-series}{Link} \\
\midrule

DB16\citep{Zesati2019ABoVE} & proximal & SVC HR-1024i & 1 m & 994 & 338--2516 &  2017  & 1660 & Reflectance spectra   & tundra & \href{https://daac.ornl.gov/cgi-bin/dsviewer.pl?ds_id=1685}{Link} \\
\midrule

DB17\citep{Thompson2021EMITVeg} & proximal & ASD FieldSpec 3 & N/A & 2151 & 350--2500 &  2001  & 490 &  Reflectance spectra  & coastal, forest, shrubland & \href{https://ecosis.org/package/emit-manually-adjusted-vegetation-reflectance-spectra}{Link} \\
\midrule

\multirow{3}{*}{DB18\citep{Queally2018CAVeg}} & \multirow{3}{*}{airborne} & \multirow{3}{*}{AVIRIS Classic} & \multirow{3}{*}{17–20 m} & \multirow{3}{*}{244} & \multirow{3}{*}{365--2496} &  2013  & 341 &  \multirow{3}{*}{ACSR}  & \multirowcell{3}{Urban, chaparral, \\ oak woodland, conifer forest} & \multirow{3}{*}{\href{https://data.ecosis.org/dataset/california-vegetation-species-image-spectra}{Link}} \\
 \cline{7-8}
& & & & & & 2014 & 37 & & \\
 \cline{7-8}
& & & & & & 2016 & 31 & & \\
\midrule

DB19\citep{Kokaly2017USGSSpectralLib} & airborne & AVIRIS & 17–20 m & 224 & 365--2496 &    & 32 &    & -- & \href{https://www.sciencebase.gov/catalog/item/5807a2a2e4b0841e59e3a18d}{Link} \\
\midrule

\multirow{2}{*}{DB20\citep{PazKagan2015Multiscale}} & \multirow{2}{*}{airborne} & \multirow{2}{*}{AisaFenix} & \multirow{2}{*}{1 m} & \multirow{2}{*}{360} & \multirow{2}{*}{400--2400} &  \multirow{2}{*}{2014}  & \multirow{2}{*}{22889} &  ACSR  & \multirowcell{2}{Forest, Ecology, \\ Land Cover, Agriculture} & \multirow{2}{*}{\href{https://data.ecosis.org/dataset/multiscale-mapping-of-species-diversity}{Link}} \\
& & & & & & & & & & \\
\midrule

\multirow{2}{*}{DB21\citep{Campbell2016Mongu}} & \multirow{2}{*}{spaceborne} & \multirow{2}{*}{Hyperion} & \multirow{2}{*}{30 m} & \multirow{2}{*}{220} & \multirow{2}{*}{400--2500} & \multirowcell{2}{2008\\2009} & \multirow{2}{*}{25} &  \multirow{2}{*}{ACSR}  & \multirow{2}{*}{forest} & \multirow{2}{*}{\href{https://data.ecosis.org/dataset/eo-1-hyperion-for-vegetation-function-at-mongu}{Link}} \\
& & & & & & & & & & \\
\midrule

DB22\citep{Weinstein2022NeonTree} & airborne & NEON AOP & 1 m & 426 & 380--2510 &  2018  & 10322 & ACSR   & -- & \href{https://github.com/weecology/NeonTreeEvaluation}{Link} \\
\midrule

DB23\citep{hu2023mdas} & airborne & Hyspex & -- & 368 & 417--2484 &  2018  & 9993 &  ACSR  & crops & \href{https://mediatum.ub.tum.de/1657312}{Link} \\
\midrule

DB24\citep{Weinstein2022NeonTree} & airborne & NEON AOP & 1 m & 426 & 380--2510 &  2019  & 16373 &  ACSR  & -- & \href{https://github.com/weecology/NeonTreeEvaluation}{Link} \\
\midrule

DB25\citep{Vivone2022Fusion} & spaceborne & PRISMA & 30 m & 69 & 400--2500 &  2021  & 2155 &  ACSR  & -- & \href{https://openremotesensing.net/knowledgebase/prisma/}{Link} \\
\midrule

DB26\citep{Vivone2022Fusion} & spaceborne & PRISMA & 30 m & 63 & 400--2500 &  2021  & 10000 &  ACSR  & -- & -- \\
\midrule

DB27\citep{fuchs2023hyspecnet} & spaceborne & EnMAP & 30 m & 224 & 418--2445 &  2022  & 1890 &  ACSR  & -- & \href{https://gee-community-catalog.org/projects/hyspecnet/}{Link} \\
\midrule

DB28 & airborne & NEON AOP & 1 m & 426 & 380--2510 &  2022  & 3959 &  ACSR  & -- & \href{https://developers.google.com/earth-engine/datasets/catalog/projects_neon}{Link} \\
\midrule

DB29\citep{Roupioz2023CAMCATT} & airborne & AisaFenix & 1 m & 420 & 382--2499 &  2021  & 31811 &  ACSR  & urban vegetation & \href{https://www.toulouse-hyperspectral-data-set.com/}{Link} \\
\midrule

DB30\citep{Chadwick2025SHIFT} & airborne & Aviris NG & 20 m & 425 & 380--2510 &  2022  & 911 &  ACSR  & Mediterranean ecosystem & \href{https://avng.jpl.nasa.gov/pub/SHIFT/trait_data/}{Link} \\
\midrule

DB31 & spaceborne & EMIT & 60 m & 285 & 381--2492 &  2024  & 410 &  ACSR  & -- & \href{https://hypercoast.org/examples/emit/}{Link}\footnotemark[2] \\
\midrule

\multirow{2}{*}{DB32} & \multirow{2}{*}{spaceborne} & \multirow{2}{*}{EnMAP} & \multirow{2}{*}{30 m} & \multirow{2}{*}{224} & \multirow{2}{*}{418--2445} &  \multirowcell{2}{2022-\\2024}  & \multirow{2}{*}{6653} & \multirow{2}{*}{ACSR}  & \multirow{2}{*}{temperate forest} & \multirow{2}
{*}{\href{https://planning.enmap.org/}{EnMAP}\footnotemark[1]} \\
& & & & & & & & & & \\
\midrule

\multirow{2}{*}{DB33} & \multirow{2}{*}{spaceborne} & \multirow{2}{*}{EnMAP} & \multirow{2}{*}{30 m} & \multirow{2}{*}{224} & \multirow{2}{*}{418--2445} &  \multirowcell{2}{2022-\\2024}  & \multirow{2}{*}{1655} &  \multirow{2}{*}{ACSR}  & \multirow{2}{*}{Mediterranean ecosystem} & \multirow{2}{*}{\href{https://planning.enmap.org/}{EnMAP}\footnotemark[1]} \\
& & & & & & & & & & \\
\midrule

\multirow{2}{*}{DB34} & \multirow{2}{*}{spaceborne} & \multirow{2}{*}{EnMAP} & \multirow{2}{*}{30 m} & \multirow{2}{*}{224} & \multirow{2}{*}{418--2445} &  \multirowcell{2}{2022-\\2024}  & \multirow{2}{*}{2088} &  \multirow{2}{*}{ACSR}  & \multirow{2}{*}{temperate gra4Sand} & \multirow{2}{*}{\href{https://planning.enmap.org/}{EnMAP}\footnotemark[1]} \\
& & & & & & & & & & \\
\midrule

\multirow{2}{*}{DB35} & \multirow{2}{*}{spaceborne} & \multirow{2}{*}{EnMAP} & \multirow{2}{*}{30 m} & \multirow{2}{*}{224} & \multirow{2}{*}{418--2445} &  \multirowcell{2}{2022-\\2024}  & \multirow{2}{*}{4846} &  \multirow{2}{*}{ACSR}  & \multirow{2}{*}{tropical forest} & \multirow{2}{*}{\href{https://planning.enmap.org/}{EnMAP}\footnotemark[1]} \\
& & & & & & & & & & \\
\midrule

\multirow{2}{*}{DB36} & \multirow{2}{*}{spaceborne} & \multirow{2}{*}{EnMAP} & \multirow{2}{*}{30 m} & \multirow{2}{*}{224} & \multirow{2}{*}{418--2445} &  \multirowcell{2}{2022-\\2024}  & \multirow{2}{*}{6337} & \multirow{2}{*}{ACSR}  & \multirow{2}{*}{tropical savanna} & \multirow{2}{*}{\href{https://planning.enmap.org/}{EnMAP} \footnotemark[1]} \\
& & & & & & & & & & \\
\bottomrule
\end{tabularx}
\caption{\textbf{Summary of data sources of GreenHyperSpectra.} Technical detailed on the collected spectra and their corresponding sources. GSD = Ground Sampling Distance. ACSR = Atmospherically corrected surface reflectance.} 
\label{tab:datasets}
\vspace{1ex} 
\end{table}
\footnotetext[1]{Contains modified EnMAP data~\textcopyright~DLR [2024]. See \href{https://planning.enmap.org/}{EnMAP Portal}.}
\footnotetext[2]{Contains modified EMIT data, the original data used is licensed under the \href{http://www.apache.org/licenses/LICENSE-2.0}{Apache License, Version 2.0}.}


\begin{figure}[H]
  \centering
    \includegraphics[width=\linewidth, height=0.75\textheight, keepaspectratio]{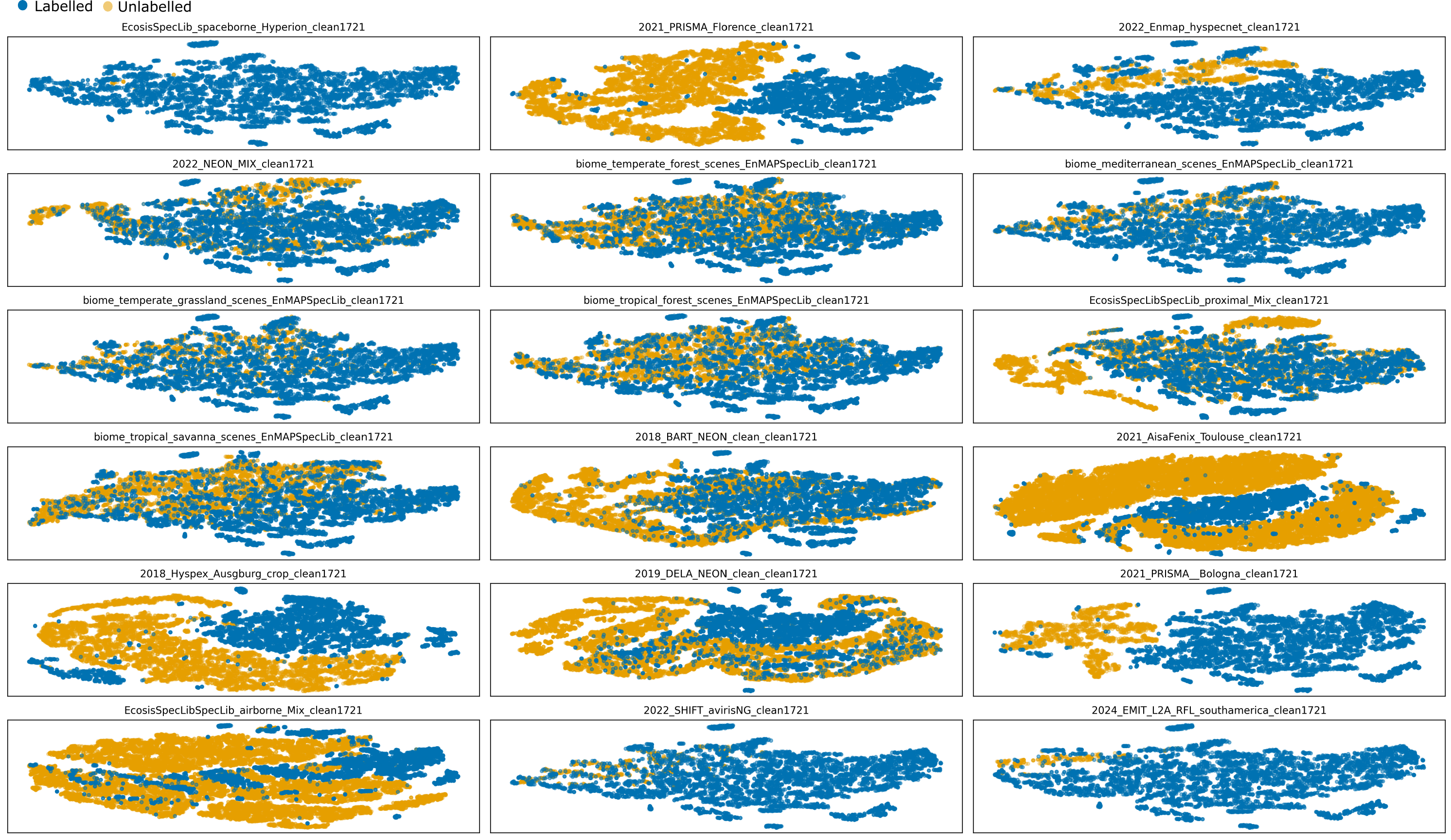}%

  \vspace{1ex} 
  
  \caption{\textbf{t-SNE projection of reflectance spectra.} Each subplot shows the projected spectral signatures from a specific data source in GreenHyperSpectra, illustrating variability driven by differences in sensors, biomes, or acquisition conditions. For each source, spectra from GreenHyperSpectra are compared to those in the aggregated annotated dataset. Orange points represent labeled spectra, and blue points denote unlabeled samples.}
  \label{fig:tsne_spectra}
\end{figure}

\vspace{-20ex}

\begin{figure}[H]
  \centering
    \includegraphics[width=\linewidth, height=0.75\textheight, keepaspectratio]{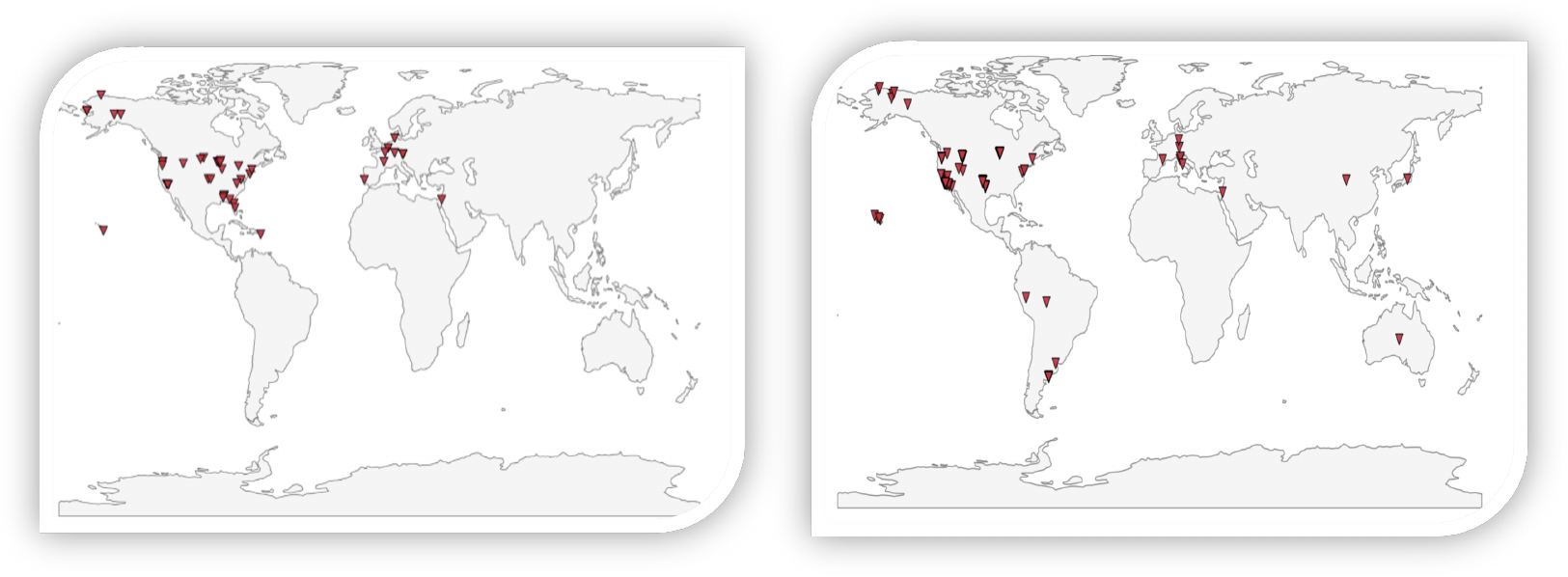}%

  \vspace{1ex} 
  
  \caption{\textbf{Spatial distribution of a subset from GreenHyperSpectra vs annotated data.} Points represent sample locations of the existing annotated data (left) and the GreenHyperSpectra subset (right). This subset, comprising 80,000 samples, was selected from the full dataset to ensure broad coverage of geographic regions and acquisition conditions. It is used for sample sensitivity analysis to assess the impact of data quantity on model performance, while enabling computationally efficient experimentation.}
  \label{fig:Appendix_subsamopUnl}
\end{figure}

\begin{table}[H]
  \caption{\textbf{Descriptive statistics for plant traits in the aggregated annotated dataset\citep{cherif2023spectra}.} List of  traits and units: leaf mass per area ($\text{g}/\text{cm}^2$) = Cm, leaf protein content ($\text{g}/\text{cm}^2$) = Cp, equivalent water thickness (cm) = Cw, leaf total chlorophyll content ($\mu\text{g}/\text{cm}^2$) = Cab, leaf carotenoid content ($\mu\text{g}/\text{cm}^2$)= Car, leaf anthocyanin content ($\mu\text{g}/\text{cm}^2$) = Anth, Leaf Area Index ($\text{m}^2/\text{m}^2$) = LAI and carbon-based constituents ($\text{g}/\text{cm}^2$)= cbc (Cm-Cp).
  }
  \vspace{1ex} 
  \label{tab:trait_stats}
  \centering
  \resizebox{\textwidth}{!}{%
  \begin{tabular}{lrrrrrrrr}
    \toprule
    Trait & Count & Mean & Std & Min & 25\% & 50\% & 75\% & Max \\
    \midrule
    cab   & 2593 & 39.1234 & 14.2312 & 4.4483 & 28.2500 & 38.0042 & 49.0675 & 229.4975 \\
    cw    & 2782 & 0.0160  & 0.0166  & 0.0000 & 0.0096  & 0.0130  & 0.0184  & 0.5138 \\
    cm    & 4062 & 0.0101  & 0.0083  & 0.0000 & 0.0051  & 0.0080  & 0.0117  & 0.0682 \\
    LAI   & 1656 & 3.4927  & 1.7178  & 0.0633 & 2.1944  & 3.4691  & 4.7743  & 8.7700 \\
    cp    & 3031 & 0.0009  & 0.0005  & 0.0000 & 0.0006  & 0.0008  & 0.0011  & 0.0050 \\
    cbc   & 3031 & 0.0104  & 0.0086  & 0.0000 & 0.0056  & 0.0078  & 0.0123  & 0.0671 \\
    car   & 1873 & 8.6925  & 2.8232  & 1.1826 & 6.9679  & 8.5176  & 10.2998 & 40.4432 \\
    anth  & 644  & 1.2730  & 0.4095  & 0.5610 & 0.9491  & 1.2345  & 1.5226  & 2.9811 \\
    \bottomrule
  \end{tabular}
  }
\end{table}

\newpage
\section{Details about Models}
\label{sec:model_details}
The code for accessing the dataset and benchmarking experiments can be found \href{https://github.com/echerif18/HyspectraSSL}{here}. The trained model objects are also available \href{https://huggingface.co/Avatarr05/Multi-trait_SSL}{here}

\subsection{Semi-supervised regression generative adversarial network (SR-GAN)}
\label{sec:GAN_details}
To enable trait prediction from unlabeled hyperspectral spectra, we adopt a semi-supervised GAN framework \citep{olmschenk2019generalizing}. The generator $G$ learns to synthesize spectra that are indistinguishable from real data by optimizing $\mathcal{L}_{\text{gen}}$, while the discriminator $D$ learns both to distinguish real from fake spectra and to regress plant trait values from real labeled data optimizing $\mathcal{L}_{disc}$ . We adopt this notation for $D$ outputs: $f$ is an intermediate feature extractor and $D \circ f$ is final layer trait prediction.
$\text{Dist}(\cdot,\cdot)$ denotes a distance metric (e.g., cosine or Euclidean).

\textbf{Notations}: \( x_{\text{fake}} \): generated fake spectra from the generator; \( x_{\text{unlb}} \): unlabeled sample from GreenHyperSpectra; \( x_{\text{lb}} \): spectra sample from the labeled data;

\paragraph{Generator Matching Loss.} The generator is trained to align the generated spectra with real spectra in the latent feature space:
\begin{equation}
\mathcal{L}_{\text{gen}} = \lambda_{\text{gen}} \cdot \text{Dist}\left(f(x_{\text{fake}}), f(x_{\text{unlabeled}})\right),
\end{equation}
where $\lambda_{\text{gen}}$ controls the influence of the generator loss.

\paragraph{Labeled Supervised Loss.} For labeled spectra $x_{\text{labeled}}$ with corresponding trait references $y$, we define a standard supervised regression loss:
\begin{equation}
\mathcal{L}_{\text{labeled}} = \lambda_{\text{labeled}} \cdot \text{MSE}(D \circ f(x_{\text{labeled}}), y),
\end{equation}
where $\lambda_{\text{labeled}}$ is a weighting coefficient, and $\text{MSE}$ denotes the mean squared error between predicted and true traits.

\paragraph{Unlabeled Matching Loss.} To regularize the feature space, we encourage the feature representations $f(\cdot)$ extracted by $D$ from labeled and unlabeled real spectra to be similar:
\begin{equation}
\mathcal{L}_{\text{unlabeled}} = \lambda_{\text{unlabeled}} \cdot \lambda_{\text{srgan}} \cdot \text{Dist}\left(f(x_{\text{labeled}}), f(x_{\text{unlabeled}})\right),
\end{equation}
where 
and $\lambda_{\text{unlabeled}}$ and $\lambda_{\text{srgan}}$ are scaling factors.

\paragraph{Fake Contrastive Loss.} The discriminator is further trained to push away fake spectra $x_{\text{fake}} = G(z)$ from real ones in the feature space:
\begin{equation}
\mathcal{L}_{\text{fake}} = \lambda_{\text{fake}} \cdot \lambda_{\text{srgan}} \cdot \text{Dist}\left(f(x_{\text{unlabeled}}), f(x_{\text{fake}})\right),
\end{equation}
where $\lambda_{\text{fake}}$ weights the contrastive term.

\newpage
\paragraph{Gradient Penalty.} A gradient penalty is used to enforce Lipschitz continuity, stabilizing the training of the discriminator:
\begin{equation}
\mathcal{L}_{\text{GP}} = \lambda_{\text{GP}} \cdot \mathbb{E}_{\hat{x}} \left[max(0, \left(\left\| \nabla_{\hat{x}} f(\hat{x}) \right\|_2 - 1 \right)^2) \right],
\end{equation}
where $\hat{x}$ is an interpolated sample between $x_{\text{fake}}$ and $x_{\text{unlabeled}}$, and $\lambda_{\text{GP}}$ is the penalty coefficient.

\paragraph{Total Losses.} The complete losses for the discriminator are defined as:
\begin{equation}
\mathcal{L}_{disc} = \mathcal{L}_{\text{labeled}} + \mathcal{L}_{\text{unlabeled}} + \mathcal{L}_{\text{fake}} + \mathcal{L}_{\text{GP}},
\end{equation}
\begin{table}[H]
  \caption{Architectural details of the convolutional GAN model used for spectral generation and trait regression.}
  \label{tab:gan_architecture}
  \vspace{1ex} 
  \centering
  \resizebox{\textwidth}{!}{%
  \begin{tabular}{lll}
    \toprule
    \textbf{Network} & \textbf{Layer} & \textbf{Description} \\
    \midrule
    \multirow{6}{*}{Generator} 
    & Input & Latent vector $z \in \mathbb{R}^{d}$ \\
    & Fully Connected & Linear: $d \rightarrow 64 \times \frac{S}{4}$, reshaped to $(64, \frac{S}{4})$ \\
    & Transposed Conv1D & $(64, \frac{S}{4}) \rightarrow (64, S)$, kernel=16, stride=4, pad=6 \\
    & Residual Stack & Three residual blocks (dilations=1, 3, 9), LeakyReLU, skip connections \\
    & Output Conv1D & Conv1D: $(64, S) \rightarrow (1, S)$, kernel=7, pad=3 \\
    & Output Activation & Tanh activation to constrain to $[-1, 1]$ \\
    \midrule
    \multirow{9}{*}{Discriminator} 
    & Input & Spectral input $x \in \mathbb{R}^{1 \times S}$ \\
    & Conv1D Layer 1 & SpectralNorm: $(1, S) \rightarrow (128, S/2)$, kernel=3, stride=2, pad=1 \\
    & BatchNorm + Activation & BatchNorm1D + LeakyReLU \\
    & Conv1D Layer 2 & SpectralNorm: $(128, S/2) \rightarrow (128, S/4)$ \\
    & Conv1D Layer 3 (output) & SpectralNorm: $(128, S/4) \rightarrow (128, S/8)$ \\
    & Adaptive Pooling & AdaptiveAvgPool1D (optional) \\
    & Flatten & Flatten to $(128 \times S/8)$ \\
    & Dropout & Dropout $p = 0.4$ \\
    & Fully Connected (output) & Linear: $128 \times S/8 \rightarrow n_{\text{traits}}$ \\
    \bottomrule
  \end{tabular}
  }
\end{table}

\begin{table}[H]
  \caption{\textbf{Training hyperparameters for the SR-GAN.} This table lists the default hyperparameters and optimization settings used during training for both generator and discriminator components.}
  \label{tab:gan_hyperparams}
  \vspace{1ex} 
  \centering
  \begin{tabular}{lll}
    \toprule
    \textbf{Parameter} & \textbf{Value} & \textbf{Description} \\
    \midrule
    input shape & 1720 or 500 & Number of spectral input bands \\
    latent dim & 100 & Generator latent vector size \\
    n\_lb & 8 & Number of predicted plant traits \\
    batch size & 128 & Samples per batch \\
    n\_epochs & 300 & Total training epochs \\
    learning rate G & 1e-4 & Generator optimizer learning rate \\
    learning rate D & 4*1e-4 & Discriminator optimizer learning rate \\
    optimizers & Adam (amsgrad=True) & Optimizer \\
    weight decay & 1e-4 & L2 regularization \\
    lambda\_fk & 1.0 & Generator adversarial loss weight \\
    lambda\_un & 10.0 & Unsupervised feature loss weight \\
    labeled\_loss\_multiplier & 1.0 & Supervised regression loss weight \\
    matching\_loss\_multiplier & 1.0 & Real/fake match loss weight\\
    contrasting\_loss\_multiplier & 1.0 & Contrastive loss weight \\
    srgan\_loss\_multiplier & 1.0 & Contrastive loss weight \\
    gradient penalty on & True & Enable gradient penalty \\
    gradient\_penalty\_multiplier & 10.0 & Weight for GP term \\
    augmentation & True & Data augmentation \\
    contrasting\_distance\_function & CosineEmbeddingLoss & Real/fake separation \\
    matching\_distance\_function & CosineEmbeddingLoss & Real-real alignment \\
    labeled\_loss\_function & Huber loss & Regression loss for traits \\
    \bottomrule
  \end{tabular}
\end{table}

\newpage
\subsection{Model Architecture Description for RTM-AE}
\label{sec:RTM_details}
\vspace{-23ex}

\begin{table}[H]
  \caption{\textbf{Architecture of the RTM-AE model.}}
  \label{tab:ae_rtm_architecture}
  \vspace{1ex} 
  \centering
  \begin{tabular}{lll}
    \toprule
    \textbf{Module} & \textbf{Layer} & \textbf{Description} \\
    \midrule
    \multirow{5}{*}{Encoder} 
    & Input & Spectral input $x \in \mathbb{R}^{1 \times S}$ \\
    & Fully Connected 1 & $S \rightarrow 64$, followed by LayerNorm and ReLU \\
    & Fully Connected 2 & $64 \rightarrow 32$, followed by LayerNorm and ReLU \\
    & Fully Connected 3 & $32 \rightarrow 16$, followed by LayerNorm and ReLU \\
    & Trait Output Layer & $16 \rightarrow n_{\text{traits}}$ \\
    \midrule
    RTM Decoder & Non-learnable Module & PROSAIL-PRO: $n_{\text{traits}} \rightarrow \tilde{x} \in \mathbb{R}^{1 \times 2101}$ \\
    \midrule
    \multirow{3}{*}{Correction Block}
    & Fully Connected 1 & $2101 \rightarrow 8404$, followed by ReLU \\
    & Fully Connected 2 & $8404 \rightarrow \hat{x} \in \mathbb{R}^{1 \times 2101}$ \\
    & Output & Corrected reflectance spectrum in $\mathbb{R}^{S}$ \\
    \bottomrule
  \end{tabular}
\end{table}

\begin{table}[H]
  \caption{\textbf{Training hyperparameters for the RTM-AE.} This table lists the default hyperparameters and optimization settings used during training.}
  \label{tab:aertm_hyperparams}
  \vspace{1ex} 
  \centering
  \begin{tabular}{lll}
    \toprule
    \textbf{Parameter} & \textbf{Value} & \textbf{Description} \\
    \midrule
    input shape & 1720 or 500 & Number of spectral input bands \\
    latent dimension & 8 & Number of biophysical traits (latent features) \\
    output spectrum length & 2101 & Number of bands in RTM-simulated output \\
    batch size & 128 & Number of samples per training batch \\
    training epochs & 300 & Set during experimental runs \\
    learning rate & 1e-4 & Learning rate used for the Adam optimizer \\
    weight decay & 1e-4 & L2 regularization term \\
    optimizer & Adam (amsgrad=True) & Optimizer used \\
    reconstruction loss & Cosine similarity + MAE & Match predicted vs. input spectra \\
    label loss & Huber loss & Trait prediction loss on labeled samples \\
    gradient stabilization & Enabled & Replace gradients when $\|\nabla\| < 10^{-5}$ \\
    RTM decoder & PROSAIL-PRO & Fixed physics-based decoder \\
    leaf model & PROSPECTPro & Leaf optical model \\
    canopy model & SAIL & Canopy RTM model \\
    \bottomrule
  \end{tabular}
\end{table}

\begin{table}[H]
\caption{\textbf{Parameter configuration for the PROSAIL-PRO model \citep{feret2021prospect}.} This table presents the default settings and corresponding notations for parameters used in the RTM block, which simulates leaf and canopy reflectance based on the PROSPECT-PRO \citep{jacquemoud1990prospect} and 4SAIL \citep{verhoef2007unified} models.}
\label{tab:aertm_RTM}
\vspace{1ex} 
\centering
\begin{tabular}{llll}
\toprule
\textbf{Model} & \textbf{Variable} & \textbf{Notation (unit)} & \textbf{Range} \\
\midrule
\multirow{8}{*}{PROSPECT-PRO} 
& Chlorophyll content & Cab ($\mu$g/cm$^2$) & Variable \\
& Carotenoid content & Car ($\mu$g/cm$^2$) & Variable \\
& Anthocyanin content & Anth ($\mu$g/cm$^2$) & Variable \\
& Water content & Cw (g/m$^2$) & Variable \\
& Protein content & Cp (g/m$^2$) & Variable \\
& Carbon-based constituents & CBC (g/m$^2$) = Cm - Cp & Variable \\
& Brown pigment content & Brown (--) & 0.25 \\
& Structural coefficient & Ns (--) & 1.5 \\
\midrule
\multirow{7}{*}{4SAIL}
& Leaf area index & LAI (m$^2$/m$^2$) & Variable \\
& Average leaf inclination angle & LIDF (Beta index) & 5 \\
& Fraction of dry soil & psoil & 0.8 \\
& Hotspot & hspot & 0.01 \\
& Viewing zenith angle & tto ($^\circ$) & 0 \\
& Solar zenith angle & tts ($^\circ$) & 30 \\
& Relative azimuth angle & psi ($^\circ$) & 0 \\
\bottomrule
\end{tabular}
\end{table}

\newpage
\subsection{Model Architecture Description for the 1D masked autoencoder (MAE)}
\label{sec:MAE_details}
\subsubsection{Training description}
\begin{table}[H]
  \caption{Architecture of the Masked Autoencoder (MAE)}
  \vspace{1ex} 
  \label{tab:mae_architecture}
  \centering
  \begin{tabular}{lll}
    \toprule
    \textbf{Module} & \textbf{Layer} & \textbf{Description} \\
    \midrule
    \multirow{4}{*}{Input} 
    & Input Spectrum & 1D spectral vector $x \in \mathbb{R}^{1 \times S}$ \\
    & Patch Embedding & Tokens of size $T$ via frozen Conv1D, $N = S/T$ \\
    & Positional Embedding & Fixed 1D sin-cos embeddings \\
    & Masking & Random masking of 75\% of tokens \\
    \midrule
    \multirow{3}{*}{Encoder} 
    & Transformer Blocks & $d$ blocks with $h$-head attention, MLP ($4 \times d$) \\
    & Normalization & LayerNorm \\
    & Output & Latent representation $\in \mathbb{R}^{N \times d}$ \\
    \midrule
    \multirow{4}{*}{Decoder} 
    & Linear Projection & Projects latent dim to decoder dimensions \\
    & Token Restoration & Restore with mask tokens using $ids\_restore$ \\
    & Transformer Blocks & $d'$ blocks with $h'$-head attention \\
    & Output Projection & Linear: decoder dim $\rightarrow$ token size $T$ \\
    \midrule
    Reconstruction & Spectrum Output & Reconstructed full spectrum $\in \mathbb{R}^{1 \times S}$ \\
    \bottomrule
  \end{tabular}

\end{table}

\begin{table}[H]
  \caption{\textbf{Training hyperparameters of the MAE.} This table lists the default hyperparameters and optimization settings used during training on the pretext task (spectra reconstruction).}
  \vspace{1ex} 
  \label{tab:mae_hyperparams}
  \centering
  \begin{tabular}{lll}
    \toprule
    \textbf{Parameter} & \textbf{Value} & \textbf{Description} \\
    \midrule
    Input dimension ($S$) & 1720 or 500 & Number of spectral bands \\
    Patch size ($T$) & 10, 20, 40, 430 & Token size (Ablation study) \\
    Embedding dimension ($d$) & 128 & Latent dimension \\
    Encoder depth ($d$) & 4, 6, 8 & Transformer blocks (Ablation study) \\
    Encoder heads ($h$) & 4, 8, 16 & Attention heads (Ablation study) \\
    Decoder depth ($d'$) & 4 & Decoder transformer depth \\
    Decoder heads ($h'$) & 4 & Decoder attention heads \\
    Mask ratio & 0.75 & Fraction of masked tokens \\
    Loss function & WLoss * Cosine + MSE & Shape and amplitude combined loss \\
    WLoss & 1, 0.1, 0.001, 0 & Weight for cosine term \\
    MLP ratio & 4.0 & MLP expansion factor \\
    Attention dropout & 0.0 & Dropout in attention \\
    Projection dropout & 0.0 & Dropout in projections \\
    Optimizer & AdamW & With AMSGrad \\
    Learning rate & 5e-4 & Initial learning rate \\
    Weight decay & 1e-4 & L2 regularization \\
    Batch size & 128 & Samples per batch \\
    Epochs & 500 & Total training epochs \\
    \bottomrule
  \end{tabular}
\end{table}

\subsubsection{Ablation study results}
\begin{table}[H]
  \caption{\textbf{Ablation study on the effect of transformer depth and attention heads in the MAE model.} This table reports the $R^2$ and nRMSE values for MAE models evaluated on the trait prediction task, with varying transformer depths and numbers of attention heads. The highest $R^2$ scores and lowest nRMSE values are highlighted.}
  \label{tab:mae_ablation}
  \vspace{1ex} 
  \centering
  \begin{tabular}{ccc ccc}
    \toprule
    \textbf{Depth} & \textbf{Heads} & \textbf{Final Val $R^2$} & \textbf{Final Val nRMSE} \\
    \midrule
    6  & 4  & 0.4492  & 15.73 \\
    6  & 8  & 0.4090  & 16.31 \\
    6  & 16 & 0.4327  & 15.96 \\
    8  & 4  & 0.4351  & 15.90 \\
    8  & 8  & 0.4356  & 15.90 \\
    8  & 16 & 0.3937  & 16.44 \\
    10 & 4  & 0.4060  & 16.30 \\
    10 & 8  & 0.4092  & 16.28 \\
    \textbf{10} & \textbf{16} & \textbf{0.4692}  & \textbf{15.43} \\
    \bottomrule
  \end{tabular}
\end{table}

\begin{table}[H]
  \caption{\textbf{Ablation study on the effect of cosine similarity loss weight in the MAE model.} This table presents the $R^2$ and nRMSE values for trait prediction as the cosine similarity loss weight ($w_\text{loss}$) is varied in the MAE objective. The highest $R^2$ scores and lowest nRMSE values are highlighted.}
  \label{tab:mae_loss_ablation}
  \vspace{1ex} 
  \centering
  \begin{tabular}{ccc}
    \toprule
    \textbf{$w_\text{loss}$} & \textbf{Final Val $R^2$} & \textbf{Final Val nRMSE} \\
    \midrule
    \textbf{1}     & \textbf{0.5018}  & \textbf{14.96} \\
    0.1   & 0.4907           & 15.15 \\
    0.01  & 0.4233           & 16.08 \\
    0.001 & 0.4627           & 15.55 \\
    0     & 0.4698           & 15.42 \\
    \bottomrule
  \end{tabular}
\end{table}

\begin{table}[H]
  \caption{\textbf{Ablation study on the effect of spectral token size in the MAE model.} This table reports the $R^2$ and nRMSE values for trait prediction as the spectral token (sequence) size is varied in the MAE architecture. The highest $R^2$ scores and lowest nRMSE values are highlighted.}
  \label{tab:mae_patchsize_ablation}
  \vspace{1ex} 
  \centering
  \begin{tabular}{ccc}
    \toprule
    \textbf{Token Size} & \textbf{Final Val $R^2$} & \textbf{Final Val nRMSE} \\
    \midrule
    10   & 0.4542  & 15.68 \\
    \textbf{20}   & \textbf{0.5018}  & \textbf{14.96} \\
    40   & 0.4744  & 15.35 \\
    430  & 0.2683  & 18.05 \\
    \bottomrule
  \end{tabular}
\end{table}

\newpage
\section{Resource requirements}
\begin{table}[H]
    \caption{\textbf{Comparison of methods by model size, runtime, and hardware usage.} This table summarizes the number of trainable parameters, average runtime, and GPU type used for each method, providing insights into their computational efficiency and resource requirements.}
    \vspace{1ex} 
    \centering
    \begin{tabular}{llll}
        \toprule
        \textbf{Method} & \textbf{\# Trainable Parameters} & \textbf{Run Time} & \textbf{GPU} \\
        \midrule
        Supervised (EffNetB0) & 6,998,280 & $\sim$11h & Quadro RTX 8000 \\
        MAE (pretext task) & Encoder: 2,006,912 & $\sim$20h & Quadro RTX 8000 \\
         & Decoder: 1,220,116 & & \\
         & Total: 3,227,028 & & \\
        MAE (downstream Linear Probing) & 1,288 & $\sim$15 min & Quadro RTX 8000 \\
        MAE (downstream Fine Tuning) & 1,607,196 & $\sim$15 min & Quadro RTX 8000 \\
        SR-GAN & Discriminator: 319,496 & $\sim$2.5d & Quadro RTX 8000 \\
         & Generator: 2,920,130 & & \\
         & Total: 3,239,626 & & \\
        RTM-AE & 35,437,289 & $\sim$16h & NVIDIA L40S \\
        \bottomrule
    \end{tabular}
    \label{tab:resReq}
\end{table}

\newpage
\section{complements to results}
\label{sec:resComplements}
\vspace{-1.5ex}

\begin{figure}[H]
  \centering
    \includegraphics[width=\linewidth, height=0.75\textheight, keepaspectratio]{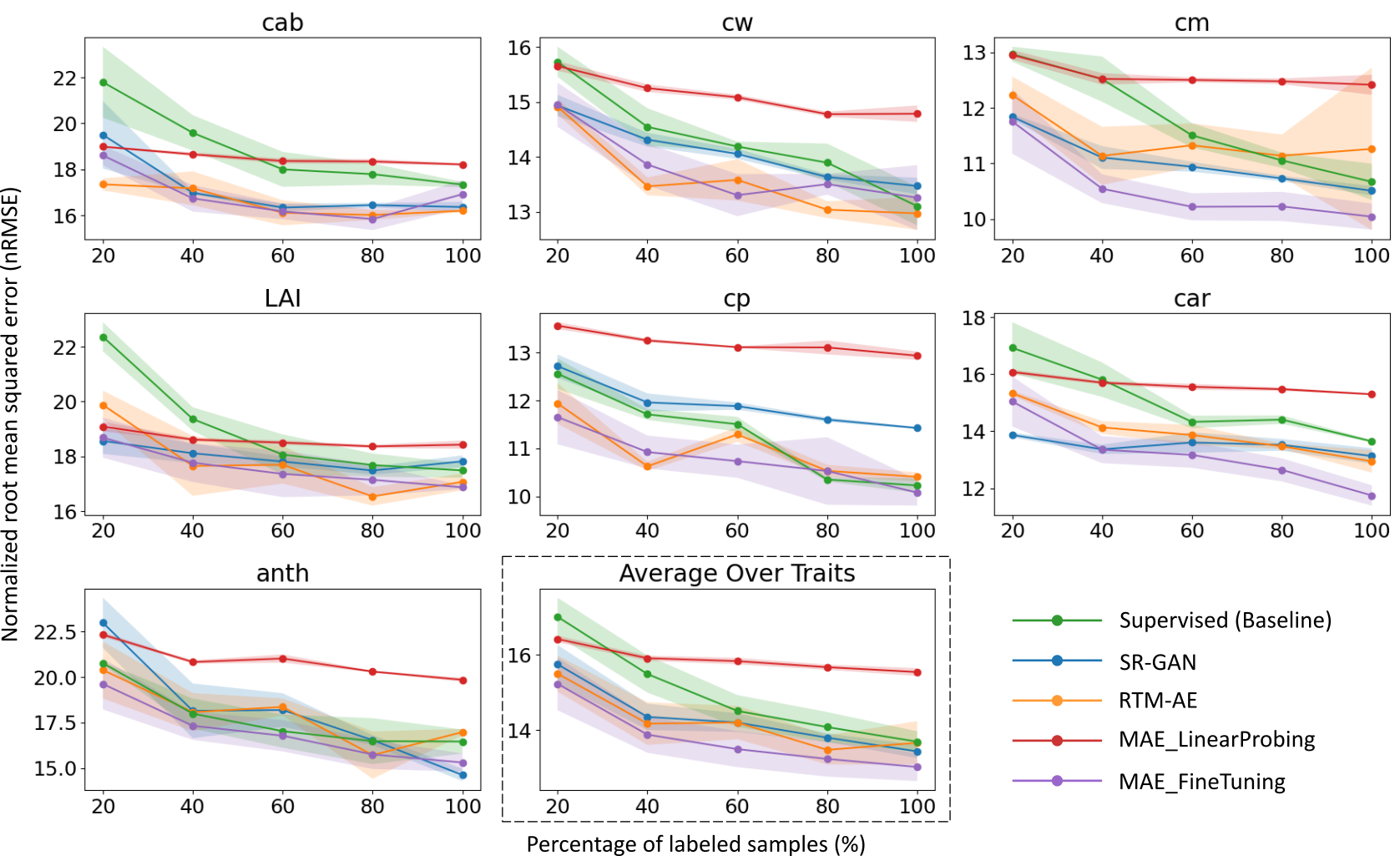}

  \vspace{1ex} 
  
  \caption{\textbf{Evaluation of trait prediction with variable-size labeled sets.} Validation performance (nRMSE) as a function of the percentage of labeled data used for training. The average nRMSE performance across all traits is indicated by the dashed box. Lower nRMSE values indicate better predictive performance.}
  \label{fig:figRes2_Labelled_n}
\end{figure}

\begin{figure}[H]
  \centering
    \includegraphics[width=\linewidth, height=0.75\textheight, keepaspectratio]{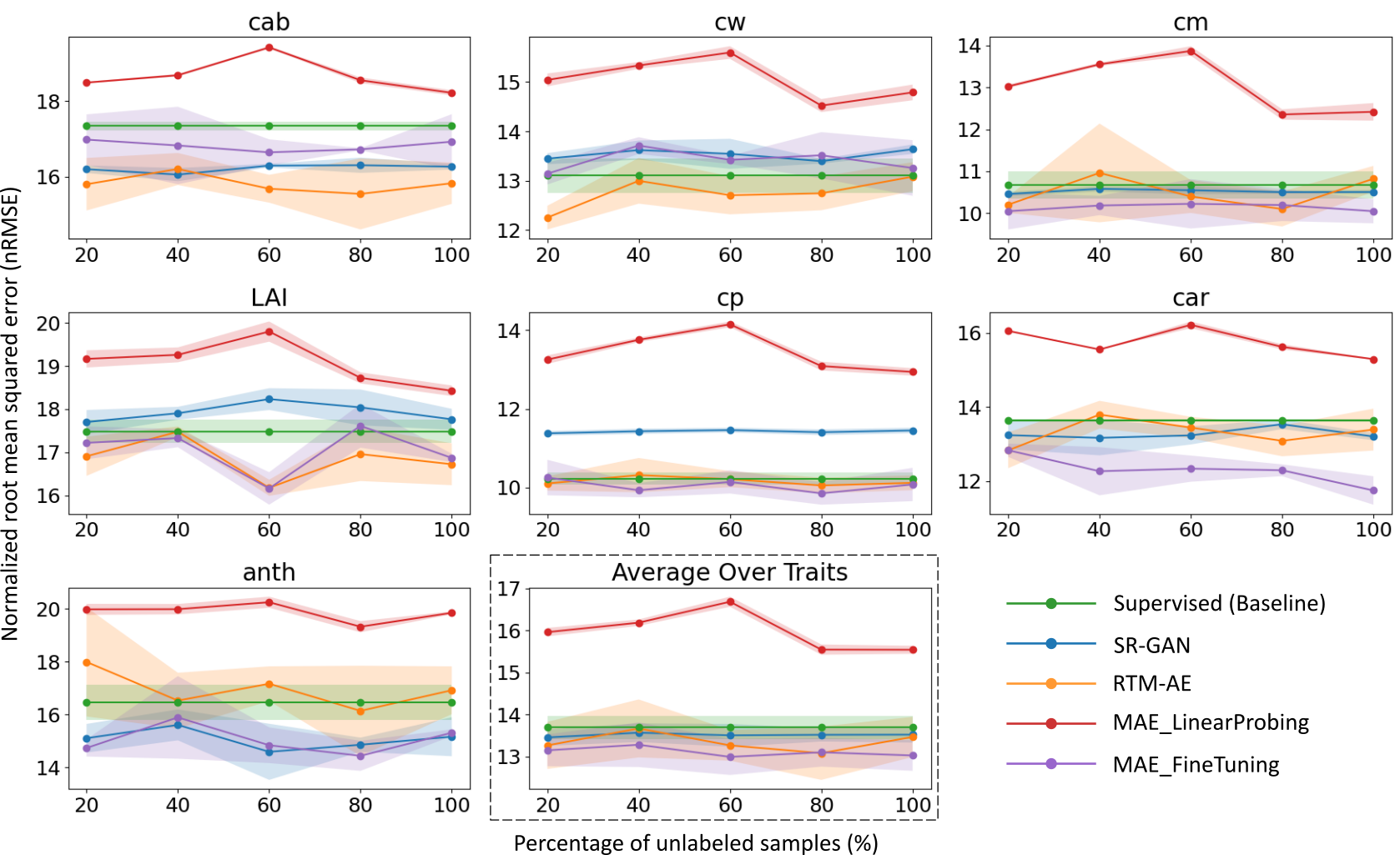}%

  \vspace{1ex} 
  
  \caption{\textbf{Evaluation of trait prediction with variable-size unlabeled sets.} Validation performance (nRMSE) as a function of the percentage of labeled data used for training. The average nRMSE performance across all traits is indicated by the dashed box. Lower nRMSE values indicate better predictive performance.}
  \label{fig:figRes2_unLabelled_n}
\end{figure}

\begin{figure}[H]
  \centering
  \includegraphics[width=\linewidth, height=0.75\textheight, keepaspectratio]{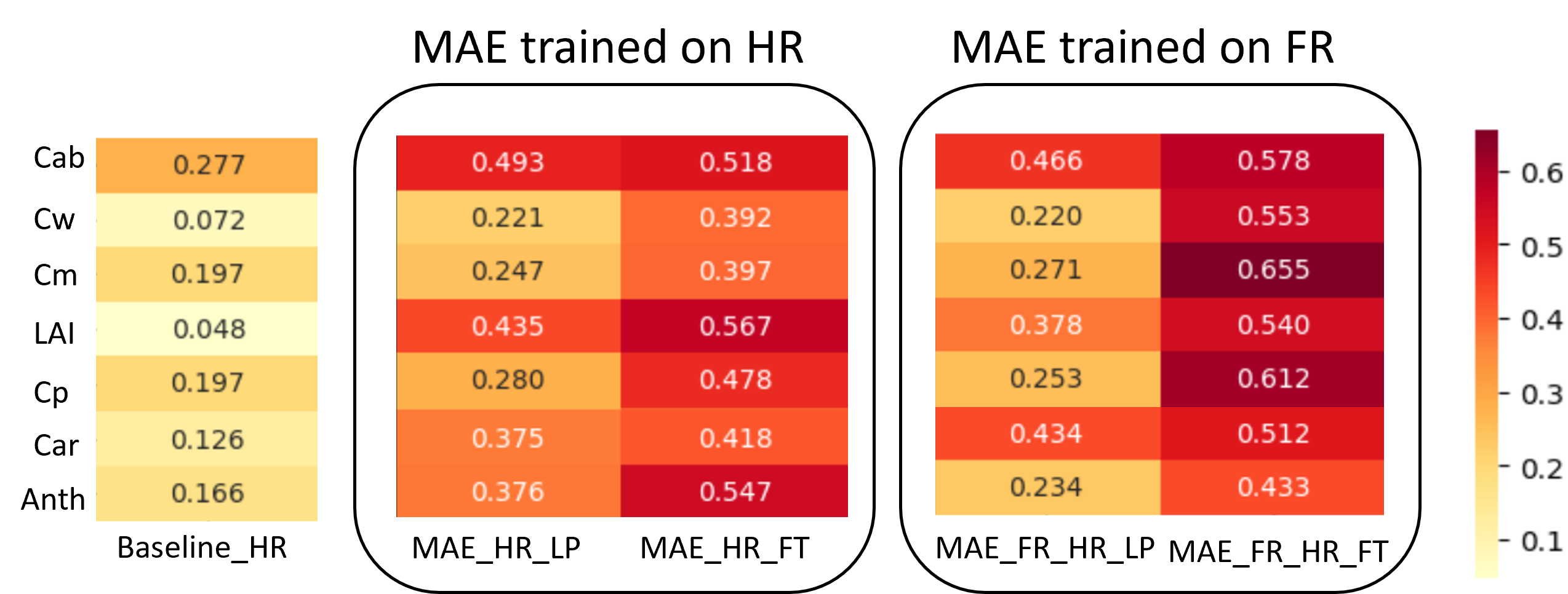}%

  \vspace{1ex} 
  \caption{\textbf{Trait-wise performance heatmaps in the half-range (HR) for MAE vs Baseline.} The heatmap displays the coefficient of determination ($R^2$; higher is better). Each cell represents the average performance across runs for a given trait-method combination.}
  \label{fig:heat_r2}
\end{figure}

\vspace{-2ex} 

\begin{figure}[H]
  \centering
  \includegraphics[width=\linewidth, height=0.75\textheight, keepaspectratio]{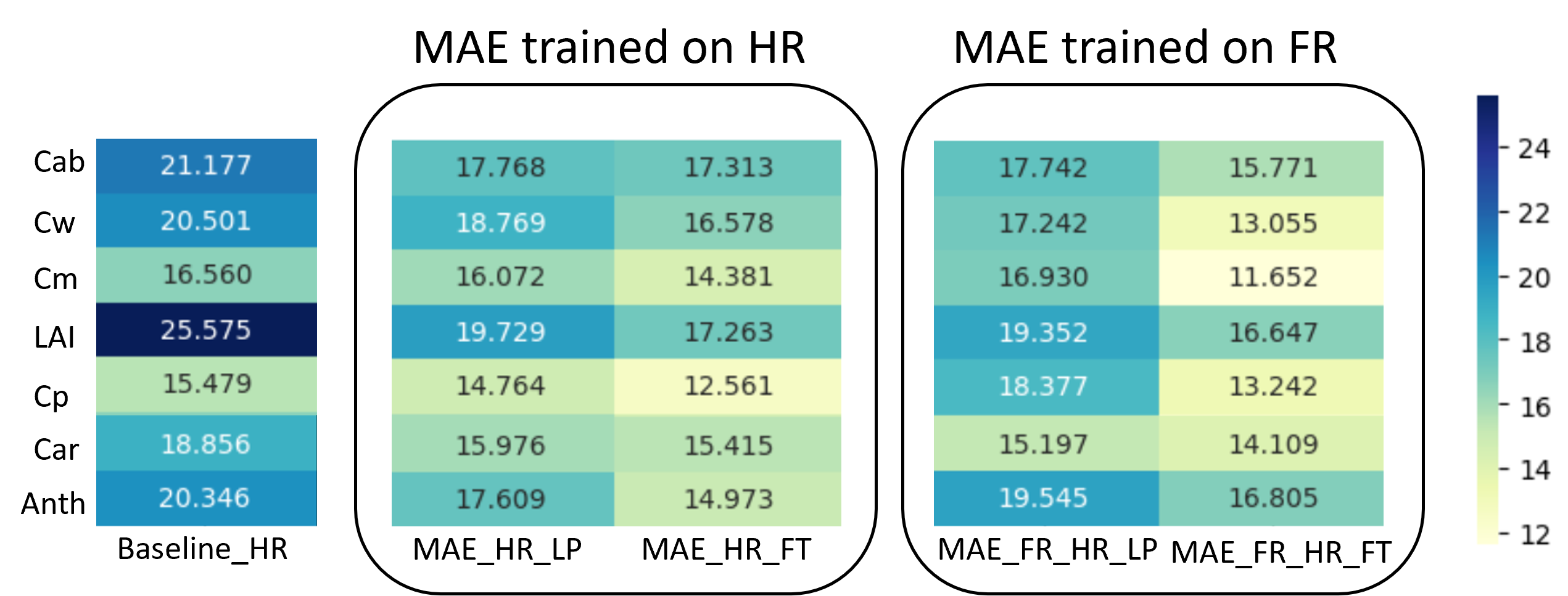}%

  \vspace{1ex} 
  \caption{\textbf{Trait-wise performance heatmaps in the half-range (HR) setting MAE vs Baseline.} The  heatmap displays the normalized root mean square error (nRMSE; lower is better). Each cell represents the average performance across runs for a given trait-method combination.}
  \label{fig:heat_nrmse}
\end{figure}

\clearpage

\begin{figure}[H]
  \centering
    \includegraphics[width=\linewidth, height=0.75\textheight, keepaspectratio]{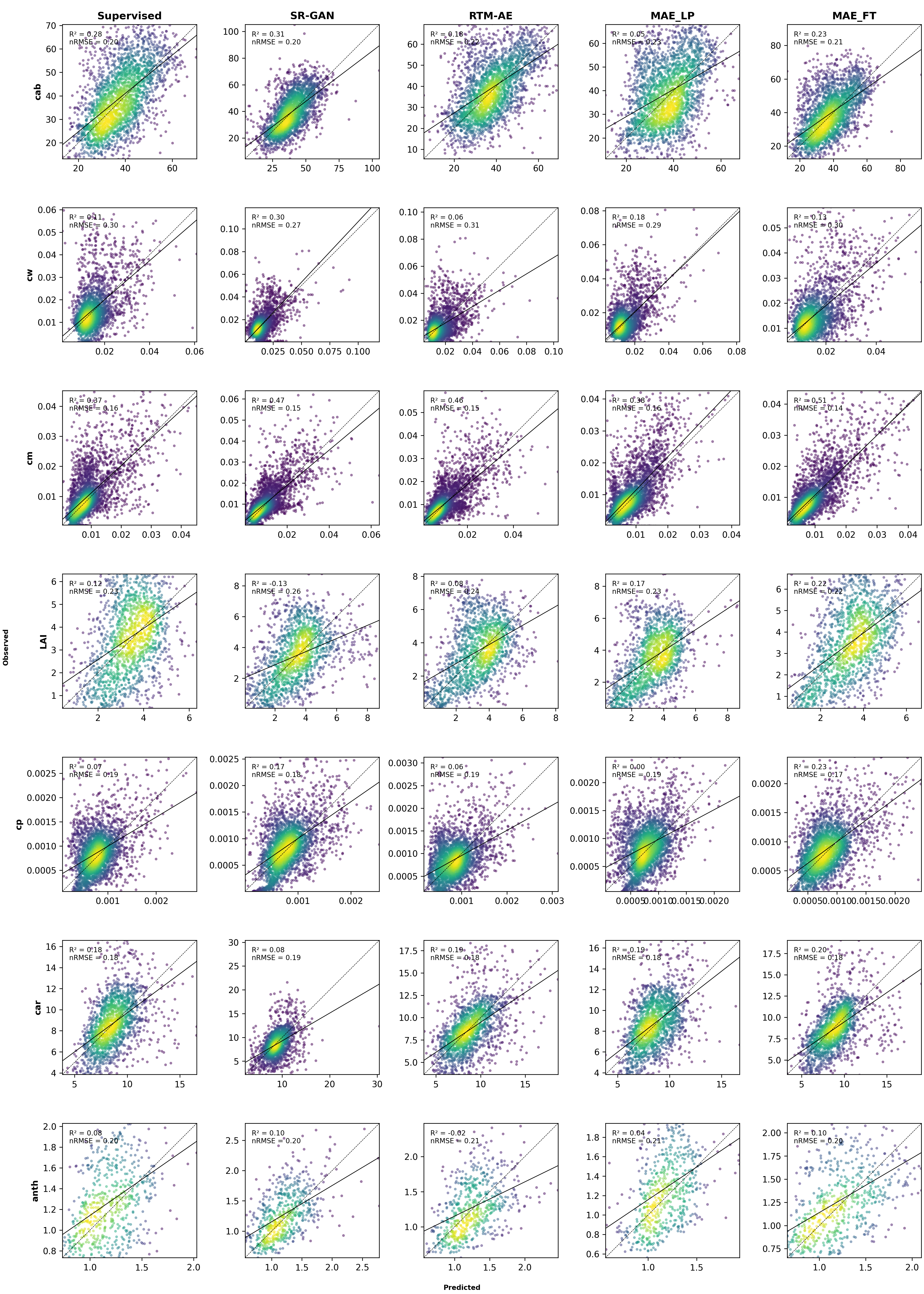}%

  \vspace{1ex} 
  \caption{\textbf{Observed vs. predicted trait values in the cross-dataset OOD setup.} Each subplot corresponds to a specific trait (rows) and method (columns), comparing predicted values to reference data. The black line indicates the 1:1 reference. $R^2$ and nRMSE values are reported in each plot to quantify predictive performance.}
  \label{fig:scatter}
\end{figure}

\begin{table}[t!]
  \centering
  \resizebox{\textwidth}{!}{%
  \begin{tabular}{lccccccccc}
    \toprule
    & cab & cw & cm & LAI & cp & cbc & car & anth & \lightgrey Average \\
    \midrule
    \multicolumn{10}{c}{$R^2$ $(\uparrow)$} \\
    \midrule
    Supervised & \underline{0.2836} & 0.1107 & 0.3728 & 0.1176 & 0.0795 & 0.3339 & 0.1784 & 0.0810 & \lightgrey 0.1947 \\
    SR\_GAN & \textbf{0.3108} & \textbf{0.3065} & \underline{0.4931} & -0.1324 & \underline{0.1924} & \underline{0.4810} & 0.0831 & \textbf{0.1022} & \lightgrey \underline{0.2296} \\
    RTM\_AE & 0.1846 & 0.0577 & 0.4681 & 0.0790 & 0.0704 & -0.1646 & \underline{0.1902} & -0.0186 & \lightgrey 0.1083 \\
    MAE\_FR\_LP & 0.0542 & \underline{0.1803} & 0.3857 & \underline{0.1673} & 0.0100 & 0.4034 & 0.1864 & 0.0442 & \lightgrey 0.1789 \\
    MAR\_FR\_FT & 0.2257 & 0.1337 & \textbf{0.5184} & \textbf{0.2239} & \textbf{0.2425} & \textbf{0.4837} & \textbf{0.2001} & \underline{0.0967} & \lightgrey \textbf{0.2656} \\
    \midrule
    \multicolumn{10}{c}{nRMSE $(\downarrow)$} \\
    \midrule
    Supervised & \underline{20.1711} & 30.4925 & 16.2647 & 23.4458 & 18.6372 & 17.3007 & 17.7155 & 20.3835 & \lightgrey 20.5514 \\
    SR\_GAN & \textbf{19.7855} & \textbf{26.9263} & \underline{14.6236} & 26.5084 & \underline{17.4572} & \underline{15.2720} & 18.7152 & \textbf{20.1473} & \lightgrey \underline{19.9294} \\
    RTM\_AE & 21.5192 & 31.3877 & 14.9780 & 23.9535 & 18.7291 & 22.8767 & \underline{17.5886} & 21.4592 & \lightgrey 21.5615 \\
    MAE\_FR\_LP & 23.1762 & \underline{29.2746} & 16.0959 & \underline{22.7759} & 19.3277 & 16.3739 & 17.6299 & 20.7873 & \lightgrey 20.6802 \\
    MAR\_FR\_FT & 20.9691 & 30.0953 & \textbf{14.2528} & \textbf{21.9882} & \textbf{16.9061} & \textbf{15.2323} & \textbf{17.4806} & \underline{20.2090} & \lightgrey \textbf{19.6417} \\
    \bottomrule
  \end{tabular}
  }
\vspace{1ex}
\caption{
\textbf{Cross-dataset generalization by vegetation type: Tundra.} 
Unlike the overall OOD results (Table~\ref{tab:ood_results}), here we exclude samples from tundra during evaluation to assess its individual impact on model generalization. 
Trait-wise performance is reported using $R^2$ $(\uparrow)$ and nRMSE $(\downarrow)$. 
We highlight the best and second-best scores in \textbf{bold} and \underline{underline}, respectively.
}
\label{tab:tundra_results}
\vspace{-3mm}
\end{table}

\begin{table}[t!]
  \centering
  \resizebox{\textwidth}{!}{%
  \begin{tabular}{lccccccccc}
    \toprule
    & cab & cw & cm & LAI & cp & cbc & car & anth & \lightgrey Average \\
    \midrule
    \multicolumn{10}{c}{$R^2$ $(\uparrow)$} \\
    \midrule
    Supervised & \underline{0.3586} & 0.0432 & 0.1471 & 0.1229 & -0.0927 & 0.1366 & \underline{0.3344} & 0.0810 & \lightgrey 0.1414 \\
    SR\_GAN & \textbf{0.3645} & \textbf{0.2524} & \underline{0.2372} & -0.1402 & \underline{0.0366} & \underline{0.2824} & -0.0012 & \textbf{0.1022} & \lightgrey \underline{0.1417} \\
    RTM\_AE & 0.1744 & -0.0311 & 0.1994 & 0.1343 & -0.0779 & -0.3144 & 0.2837 & -0.0186 & \lightgrey 0.0437 \\
    MAE\_FR\_LP & 0.0412 & \underline{0.1122} & 0.1686 & \underline{0.1624} & -0.2675 & 0.2253 & 0.2823 & 0.0442 & \lightgrey 0.0961 \\
    MAE\_FR\_FT & 0.2188 & 0.0491 & \textbf{0.3057} & \textbf{0.2248} & \textbf{0.0710} & \textbf{0.2927} & \textbf{0.3348} & \underline{0.0967} & \lightgrey \textbf{0.1992} \\
    \midrule
    \multicolumn{10}{c}{nRMSE $(\downarrow)$} \\
    \midrule
    Supervised & \underline{18.8612} & 37.3976 & 18.4678 & 23.8133 & 22.9505 & 19.2217 & \underline{17.0518} & 20.3835 & \lightgrey 22.2684 \\
    SR\_GAN & \textbf{18.7736} & \textbf{33.0584} & \underline{17.4651} & 27.1220 & \underline{21.5493} & \underline{17.5235} & 20.9137 & \textbf{20.1473} & \lightgrey \underline{22.0691} \\
    RTM\_AE & 21.3980 & 38.8223 & 17.8924 & 23.6594 & 22.7941 & 23.7159 & 17.6905 & 21.4592 & \lightgrey 23.4290 \\
    MAE\_FR\_LP & 23.0604 & \underline{36.0248} & 18.2334 & \underline{23.2713} & 24.7179 & 18.2073 & 17.7073 & 20.7873 & \lightgrey 22.7512 \\
    MAE\_FR\_FT & 20.8151 & 37.2836 & \textbf{16.6626} & \textbf{22.3877} & \textbf{21.1619} & \textbf{17.3969} & \textbf{17.0472} & \underline{20.2090} & \lightgrey \textbf{21.6205} \\
    \bottomrule
  \end{tabular}
  }
\vspace{1ex}
\caption{
\textbf{Cross-dataset generalization by vegetation type: Forest.} 
Unlike the overall OOD results (Table~\ref{tab:ood_results}), here we exclude samples from forest during evaluation to assess its individual impact on model generalization. 
Trait-wise performance is reported using $R^2$ $(\uparrow)$ and nRMSE $(\downarrow)$. 
We highlight the best and second-best scores in \textbf{bold} and \underline{underline}, respectively.
}
\label{tab:forest_results}
\vspace{-3mm}
\end{table}

\begin{table}[t!]
  \centering
  \resizebox{\textwidth}{!}{%
  \begin{tabular}{lccccccccc}
    \toprule
    & cab & cw & cm & LAI & cp & cbc & car & anth & \lightgrey Average \\
    \midrule
    \multicolumn{10}{c}{$R^2$ $(\uparrow)$} \\
    \midrule
    Supervised & 0.2522 & 0.1019 & 0.3391 & \underline{-0.0108} & -0.1300 & 0.2478 & \underline{0.1712} & 0.0810 & \lightgrey 0.1316 \\
    SR\_GAN & \underline{0.2811} & \textbf{0.3282} & \underline{0.4731} & -0.0505 & \underline{-0.0075} & \underline{0.3965} & 0.0441 & \textbf{0.1022} & \lightgrey \underline{0.1959} \\
    RTM\_AE & 0.2239 & 0.0859 & 0.4679 & -0.1631 & -0.1097 & -0.2146 & 0.1450 & -0.0186 & \lightgrey 0.0521 \\
    MAE\_FR\_LP & 0.0786 & \underline{0.2257} & 0.3792 & -0.0207 & -0.1404 & 0.3455 & 0.1641 & 0.0442 & \lightgrey 0.1345 \\
    MAR\_FR\_FT & \textbf{0.2897} & 0.1377 & \textbf{0.5074} & \textbf{0.0678} & \textbf{0.0769} & \textbf{0.4326} & \textbf{0.1835} & \underline{0.0967} & \lightgrey \textbf{0.2240} \\
    \midrule
    \multicolumn{10}{c}{nRMSE $(\downarrow)$} \\
    \midrule
    Supervised & 20.3244 & 34.6730 & 16.6769 & \underline{24.8992} & 20.4582 & 18.4717 & \underline{17.4666} & 20.3835 & \lightgrey 21.6692 \\
    SR\_GAN & \underline{19.9301} & \textbf{29.9879} & \underline{14.8919} & 25.3229 & \underline{19.3170} & \underline{16.5456} & 18.7583 & \textbf{20.1473} & \lightgrey \underline{20.6126} \\
    RTM\_AE & 20.7059 & 34.9801 & 14.9635 & 26.7101 & 20.2731 & 23.4729 & 17.7401 & 21.4592 & \lightgrey 22.5381 \\
    MAE\_FR\_LP & 22.5613 & \underline{32.1942} & 16.1629 & 25.0216 & 20.5521 & 17.2313 & 17.5413 & 20.7873 & \lightgrey 21.5065 \\
    MAR\_FR\_FT & \textbf{19.8086} & 33.9740 & \textbf{14.3981} & \textbf{23.9122} & \textbf{18.4904} & \textbf{16.0433} & \textbf{17.3367} & \underline{20.2090} & \lightgrey \textbf{20.5215} \\
    \bottomrule
  \end{tabular}
  }
\vspace{1ex}
\caption{
\textbf{Cross-dataset generalization by vegetation type: Crops.} 
Unlike the overall OOD results (Table~\ref{tab:ood_results}), here we exclude samples from crops during evaluation to assess its individual impact on model generalization. 
Trait-wise performance is reported using $R^2$ $(\uparrow)$ and nRMSE $(\downarrow)$. 
We highlight the best and second-best scores in \textbf{bold} and \underline{underline}, respectively.
}
\label{tab:crops_results}
\vspace{-3mm}
\end{table}

\begin{table}[t!]
  \centering
  \resizebox{\textwidth}{!}{%
  \begin{tabular}{lccccccccc}
    \toprule
    & cab & cw & cm & LAI & cp & cbc & car & anth & \lightgrey Average \\
    \midrule
    \multicolumn{10}{c}{$R^2$ $(\uparrow)$} \\
    \midrule
    Supervised & \underline{0.3099} & 0.1359 & 0.4383 & 0.1063 & 0.1549 & 0.4156 & 0.1811 & 0.0810 & \lightgrey 0.2279 \\
    SR\_GAN & \textbf{0.3318} & \textbf{0.3229} & \underline{0.5161} & -0.1941 & \underline{0.2496} & \underline{0.5148} & 0.0826 & \textbf{0.1022} & \lightgrey \underline{0.2408} \\
    RTM\_AE & 0.2156 & 0.0706 & 0.5020 & 0.0455 & 0.1366 & -0.1417 & \underline{0.1872} & -0.0186 & \lightgrey 0.1247 \\
    MAE\_FR\_LP & 0.0860 & \underline{0.2016} & 0.4433 & \underline{0.1270} & 0.0973 & 0.4713 & 0.1868 & 0.0442 & \lightgrey 0.2072 \\
    MAR\_FR\_FT & 0.2459 & 0.1419 & \textbf{0.5690} & \textbf{0.2074} & \textbf{0.3095} & \textbf{0.5471} & \textbf{0.2068} & \underline{0.0967} & \lightgrey \textbf{0.2905} \\
    \midrule
    \multicolumn{10}{c}{nRMSE $(\downarrow)$} \\
    \midrule
    Supervised & \underline{20.0381} & 31.1490 & 14.8665 & 23.4980 & 17.7968 & 15.6488 & 17.7263 & 20.3835 & \lightgrey 20.1384 \\
    SR\_GAN & \textbf{19.7169} & \textbf{27.5735} & \underline{13.7989} & 27.0932 & \underline{16.7698} & \underline{14.2586} & 18.7612 & \textbf{20.1473} & \lightgrey \underline{19.7649} \\
    RTM\_AE & 21.3638 & 32.3043 & 13.9977 & 24.2834 & 17.9889 & 21.8728 & 17.6592 & 21.4592 & \lightgrey 21.3661 \\
    MAE\_FR\_LP & 23.0615 & \underline{29.9417} & 14.7998 & \underline{23.2242} & 18.3933 & 14.8854 & \underline{17.6639} & 20.7873 & \lightgrey 20.3446 \\
    MAR\_FR\_FT & 20.9468 & 31.0410 & \textbf{13.0228} & \textbf{22.1287} & \textbf{16.0873} & \textbf{13.7758} & \textbf{17.4458} & \underline{20.2090} & \lightgrey \textbf{19.3322} \\
    \bottomrule
  \end{tabular}
  }
\vspace{1ex}
\caption{
\textbf{Cross-dataset generalization by vegetation type: Shrubland.} 
Unlike the overall OOD results (Table~\ref{tab:ood_results}), here we exclude samples from shrubland during evaluation to assess its individual impact on model generalization. 
Trait-wise performance is reported using $R^2$ $(\uparrow)$ and nRMSE $(\downarrow)$. 
We highlight the best and second-best scores in \textbf{bold} and \underline{underline}, respectively.
}
\label{tab:shrubland_results}
\vspace{-3mm}
\end{table}

\begin{table}[t!]
  \centering
  \resizebox{\textwidth}{!}{%
  \begin{tabular}{lccccccccc}
    \toprule
    & cab & cw & cm & LAI & cp & cbc & car & anth & \lightgrey Average \\
    \midrule
    \multicolumn{9}{c}{$R^2$ $(\uparrow)$} \\
    \midrule
    Supervised & 0.1565 & 0.2080 & 0.3163 & 0.2771 & 0.0615 & 0.3349 & 0.1825 & NaN & \lightgrey 0.2195 \\
    SR\_GAN & \textbf{0.1989} & \textbf{0.3005} & \underline{0.4212} & -0.1232 & \underline{0.1658} & \underline{0.4556} & 0.1364 & NaN & \lightgrey \underline{0.2222} \\
    RTM\_AE & 0.0098 & 0.0615 & 0.4207 & 0.3146 & 0.0452 & -0.1678 & \underline{0.2005} & NaN & \lightgrey 0.1264 \\
    MAE\_FR\_LP & -0.1716 & 0.1244 & 0.3338 & \textbf{0.4612} & -0.0172 & 0.3976 & \textbf{0.2543} & NaN & \lightgrey 0.1975 \\
    MAR\_FR\_FT & \underline{0.0393} & \underline{0.2813} & \textbf{0.4818} & \underline{0.4239} & \textbf{0.2123} & \textbf{0.4890} & 0.1801 & NaN & \lightgrey \textbf{0.3011} \\
    \midrule
    \multicolumn{9}{c}{nRMSE $(\downarrow)$} \\
    \midrule
    Supervised & \underline{21.9248} & 19.3998 & 16.7762 & 20.4917 & 18.6994 & 17.1365 & 20.1887 & NaN & \lightgrey 19.2310 \\
    SR\_GAN & \textbf{21.3651} & \textbf{18.2311} & \underline{15.4364} & 25.5063 & \underline{17.6296} & \underline{15.5046} & 20.7497 & NaN & \lightgrey \underline{19.2033} \\
    RTM\_AE & 23.7549 & 21.1173 & 15.4424 & 19.9577 & 18.8617 & 22.7073 & 19.9653 & NaN & \lightgrey 20.2581 \\
    MAE\_FR\_LP & 25.8392 & 20.3969 & 16.5597 & \textbf{17.6901} & 19.4682 & 16.3092 & \textbf{19.2815} & NaN & \lightgrey \underline{19.3636} \\
    MAR\_FR\_FT & 23.3975 & \underline{18.4795} & \textbf{14.6050} & \underline{18.2936} & \textbf{17.1319} & \textbf{15.0207} & 20.2180 & NaN & \lightgrey \textbf{18.1638} \\
    \bottomrule
  \end{tabular}
  }
\vspace{1ex}
\caption{
\textbf{Cross-dataset generalization by vegetation type: Grassland.} 
Unlike the overall OOD results (Table~\ref{tab:ood_results}), here we exclude samples from grassland during evaluation to assess its individual impact on model generalization. 
Trait-wise performance is reported using $R^2$ $(\uparrow)$ and nRMSE $(\downarrow)$. 
We highlight the best and second-best scores in \textbf{bold} and \underline{underline}, respectively.
}
\label{tab:grassland_results}
\vspace{-3mm}
\end{table}

\begin{table}[t!]
  \centering
  \resizebox{\textwidth}{!}{%
  \begin{tabular}{lccccccccc}
    \toprule
    & cab & cw & cm & LAI & cp & cbc & car & anth & \lightgrey Average \\
    \midrule
    \multicolumn{10}{c}{$R^2$ $(\uparrow)$} \\
    \midrule
    Supervised & \underline{0.2861} & 0.1110 & 0.3663 & 0.1023 & 0.0663 & 0.3336 & 0.1836 & 0.0810 & \lightgrey 0.1913 \\
    SR\_GAN & \textbf{0.3143} & \textbf{0.3069} & \underline{0.4736} & -0.1479 & \underline{0.1707} & \underline{0.4667} & 0.0895 & \textbf{0.1022} & \lightgrey \underline{0.2220} \\
    RTM\_AE & 0.1849 & 0.0567 & 0.4620 & 0.0720 & 0.0592 & -0.1583 & \underline{0.1924} & -0.0186 & \lightgrey 0.1063 \\
    MAE\_FR\_LP & 0.0545 & \underline{0.1804} & 0.3809 & \underline{0.1577} & -0.0022 & 0.4005 & 0.1858 & 0.0442 & \lightgrey 0.1752 \\
    MAR\_FR\_FT & 0.2283 & 0.1339 & \textbf{0.5162} & \textbf{0.2179} & \textbf{0.2266} & \textbf{0.4849} & \textbf{0.2077} & \underline{0.0967} & \lightgrey \textbf{0.2640} \\
    \midrule
    \multicolumn{10}{c}{nRMSE $(\downarrow)$} \\
    \midrule
    Supervised & \underline{20.1602} & 30.5654 & 16.0148 & 23.5725 & 18.9082 & 16.9249 & 17.6727 & 20.3835 & \lightgrey 20.5253 \\
    SR\_GAN & \textbf{19.7600} & \textbf{26.9896} & \underline{14.5977} & 26.6337 & \underline{17.8197} & \underline{15.1402} & 18.6633 & \textbf{20.1473} & \lightgrey \underline{19.9689} \\
    RTM\_AE & 21.5421 & 31.4855 & 14.7563 & 23.9671 & 18.9805 & 22.3141 & \underline{17.5763} & 21.4592 & \lightgrey 21.5101 \\
    MAE\_FR\_LP & 23.2023 & \underline{29.3485} & 15.8300 & \underline{22.8332} & 19.5902 & 16.0532 & 17.6486 & 20.7873 & \lightgrey 20.6617 \\
    MAR\_FR\_FT & 20.9645 & 30.1712 & \textbf{13.9845} & \textbf{22.0142} & \textbf{17.2109} & \textbf{14.8817} & \textbf{17.4092} & \underline{20.2090} & \lightgrey \textbf{19.6056} \\
    \bottomrule
  \end{tabular}
  }
\vspace{1ex}
\caption{
\textbf{Cross-dataset generalization by vegetation type: Mix.} 
Unlike the overall OOD results (Table~\ref{tab:ood_results}), here we exclude samples from mix during evaluation to assess its individual impact on model generalization. 
Trait-wise performance is reported using $R^2$ $(\uparrow)$ and nRMSE $(\downarrow)$. 
We highlight the best and second-best scores in \textbf{bold} and \underline{underline}, respectively.
}
\label{tab:mix_results}
\vspace{-3mm}
\end{table}

\begin{table}[t!]
  \centering
  \resizebox{\textwidth}{!}{%
  \begin{tabular}{lccccccccc}
    \toprule
    Noise & cab & cw & cm & LAI & cp & cbc & car & anth & \lightgrey Average \\
    \midrule
    \multicolumn{10}{c}{$R^2$ $(\uparrow)$} \\
    \midrule
    0.01 & 0.552 $\pm$ 0.010 & 0.602 $\pm$ 0.021 & 0.677 $\pm$ 0.006 & 0.564 $\pm$ 0.030 & 0.659 $\pm$ 0.008 & 0.679 $\pm$ 0.008 & 0.588 $\pm$ 0.006 & 0.422 $\pm$ 0.062 & \lightgrey 0.593 $\pm$ 0.019 \\
    0.03 & 0.392 $\pm$ 0.072 & 0.387 $\pm$ 0.095 & 0.426 $\pm$ 0.059 & 0.285 $\pm$ 0.094 & 0.404 $\pm$ 0.118 & 0.448 $\pm$ 0.065 & 0.374 $\pm$ 0.101 & 0.298 $\pm$ 0.032 & \lightgrey 0.377 $\pm$ 0.079 \\
    0.05 & -0.031 $\pm$ 0.214 & -0.123 $\pm$ 0.212 & -0.011 $\pm$ 0.060 & -0.181 $\pm$ 0.230 & -0.083 $\pm$ 0.173 & 0.009 $\pm$ 0.059 & -0.127 $\pm$ 0.220 & 0.028 $\pm$ 0.045 & \lightgrey -0.065 $\pm$ 0.152 \\
    \midrule
    \multicolumn{10}{c}{nRMSE $(\downarrow)$} \\
    \midrule
    0.01 & 16.704 $\pm$ 0.181 & 13.409 $\pm$ 0.354 & 10.534 $\pm$ 0.100 & 17.332 $\pm$ 0.620 & 10.165 $\pm$ 0.121 & 10.674 $\pm$ 0.127 & 12.975 $\pm$ 0.093 & 16.935 $\pm$ 0.906 & \lightgrey 13.591 $\pm$ 0.313 \\
    0.03 & 19.436 $\pm$ 1.164 & 16.617 $\pm$ 1.326 & 14.021 $\pm$ 0.727 & 22.154 $\pm$ 1.436 & 13.392 $\pm$ 1.361 & 13.983 $\pm$ 0.834 & 15.952 $\pm$ 1.323 & 18.669 $\pm$ 0.422 & \lightgrey 16.778 $\pm$ 1.074 \\
    0.05 & 25.244 $\pm$ 2.650 & 22.482 $\pm$ 2.104 & 18.624 $\pm$ 0.555 & 28.434 $\pm$ 2.725 & 18.078 $\pm$ 1.438 & 18.753 $\pm$ 0.563 & 21.385 $\pm$ 2.152 & 21.969 $\pm$ 0.507 & \lightgrey 21.871 $\pm$ 1.587 \\
    \bottomrule
  \end{tabular}
  }
\vspace{1ex}
\caption{
\textbf{Supervised: Noise robustness analysis.} 
Model performance under different noise intensities (0.01, 0.03, 0.05). 
Trait-wise $R^2$ (higher is better) and nRMSE (lower is better) are reported as mean $\pm$ standard deviation.
}
\label{tab:noise_results_sup}
\vspace{-3mm}
\end{table}

\begin{table}[t!]
  \centering
  \resizebox{\textwidth}{!}{%
  \begin{tabular}{lccccccccc}
    \toprule
    Noise & cab & cw & cm & LAI & cp & cbc & car & anth & \lightgrey Average \\
    \midrule
    \multicolumn{10}{c}{$R^2$ $(\uparrow)$} \\
    \midrule
    0.01 & 0.539 $\pm$ 0.018 & 0.544 $\pm$ 0.006 & 0.630 $\pm$ 0.010 & 0.522 $\pm$ 0.012 & 0.528 $\pm$ 0.017 & 0.659 $\pm$ 0.007 & 0.533 $\pm$ 0.009 & 0.497 $\pm$ 0.012 & \lightgrey 0.556 $\pm$ 0.011 \\
    0.03 & 0.329 $\pm$ 0.047 & 0.292 $\pm$ 0.031 & 0.318 $\pm$ 0.060 & 0.375 $\pm$ 0.025 & 0.264 $\pm$ 0.047 & 0.348 $\pm$ 0.096 & 0.283 $\pm$ 0.054 & 0.108 $\pm$ 0.118 & \lightgrey 0.290 $\pm$ 0.060 \\
    0.05 & -0.038 $\pm$ 0.164 & -0.071 $\pm$ 0.122 & -0.195 $\pm$ 0.300 & 0.065 $\pm$ 0.212 & -0.246 $\pm$ 0.280 & -0.196 $\pm$ 0.364 & -0.080 $\pm$ 0.103 & -0.468 $\pm$ 0.317 & \lightgrey -0.154 $\pm$ 0.233 \\
    \midrule
    \multicolumn{10}{c}{nRMSE $(\downarrow)$} \\
    \midrule
    0.01 & 16.943 $\pm$ 0.323 & 14.367 $\pm$ 0.097 & 11.272 $\pm$ 0.147 & 18.143 $\pm$ 0.236 & 11.958 $\pm$ 0.218 & 11.008 $\pm$ 0.115 & 13.810 $\pm$ 0.133 & 15.811 $\pm$ 0.185 & \lightgrey 14.164 $\pm$ 0.182 \\
    0.03 & 20.421 $\pm$ 0.708 & 17.887 $\pm$ 0.390 & 15.294 $\pm$ 0.664 & 20.773 $\pm$ 0.442 & 14.897 $\pm$ 0.478 & 15.175 $\pm$ 1.130 & 17.108 $\pm$ 0.656 & 21.016 $\pm$ 1.414 & \lightgrey 17.821 $\pm$ 0.735 \\
    0.05 & 25.367 $\pm$ 1.969 & 21.970 $\pm$ 1.256 & 20.147 $\pm$ 2.468 & 25.295 $\pm$ 2.775 & 19.220 $\pm$ 2.021 & 20.449 $\pm$ 3.018 & 20.982 $\pm$ 1.001 & 26.895 $\pm$ 3.012 & \lightgrey 22.541 $\pm$ 2.190 \\
    \bottomrule
  \end{tabular}
  }
\vspace{1ex}
\caption{
\textbf{SR\_GAN: Noise robustness analysis.} 
Model performance under different noise intensities (0.01, 0.03, 0.05). 
Trait-wise $R^2$ (higher is better) and nRMSE (lower is better) are reported as mean $\pm$ standard deviation.
}
\label{tab:gan_noise_results}
\vspace{-3mm}
\end{table}

\begin{table}[t!]
  \centering
  \resizebox{\textwidth}{!}{%
  \begin{tabular}{lccccccccc}
    \toprule
    Noise & cab & cw & cm & LAI & cp & cbc & car & anth & \lightgrey Average \\
    \midrule
    \multicolumn{10}{c}{$R^2$ $(\uparrow)$} \\
    \midrule
    0.01 & 0.576 $\pm$ 0.028 & 0.633 $\pm$ 0.020 & 0.659 $\pm$ 0.018 & 0.549 $\pm$ 0.029 & 0.654 $\pm$ 0.012 & 0.666 $\pm$ 0.022 & 0.551 $\pm$ 0.040 & 0.312 $\pm$ 0.140 & \lightgrey 0.575 $\pm$ 0.038 \\
    0.03 & 0.527 $\pm$ 0.037 & 0.595 $\pm$ 0.004 & 0.620 $\pm$ 0.032 & 0.515 $\pm$ 0.043 & 0.602 $\pm$ 0.015 & 0.636 $\pm$ 0.029 & 0.499 $\pm$ 0.056 & 0.219 $\pm$ 0.164 & \lightgrey 0.527 $\pm$ 0.048 \\
    0.05 & 0.377 $\pm$ 0.023 & 0.439 $\pm$ 0.082 & 0.456 $\pm$ 0.054 & 0.457 $\pm$ 0.042 & 0.456 $\pm$ 0.068 & 0.456 $\pm$ 0.066 & 0.350 $\pm$ 0.055 & -0.258 $\pm$ 0.880 & \lightgrey 0.342 $\pm$ 0.159 \\
    \midrule
    \multicolumn{10}{c}{nRMSE $(\downarrow)$} \\
    \midrule
    0.01 & 16.245 $\pm$ 0.531 & 12.892 $\pm$ 0.346 & 10.817 $\pm$ 0.281 & 17.615 $\pm$ 0.567 & 10.246 $\pm$ 0.172 & 10.879 $\pm$ 0.350 & 13.536 $\pm$ 0.616 & 18.420 $\pm$ 1.940 & \lightgrey 13.831 $\pm$ 0.600 \\
    0.03 & 17.153 $\pm$ 0.667 & 13.543 $\pm$ 0.066 & 11.411 $\pm$ 0.486 & 18.302 $\pm$ 0.782 & 10.987 $\pm$ 0.214 & 11.356 $\pm$ 0.451 & 14.282 $\pm$ 0.813 & 19.619 $\pm$ 2.134 & \lightgrey 14.582 $\pm$ 0.702 \\
    0.05 & 19.690 $\pm$ 0.362 & 15.899 $\pm$ 1.198 & 13.647 $\pm$ 0.686 & 19.385 $\pm$ 0.771 & 12.828 $\pm$ 0.818 & 13.884 $\pm$ 0.863 & 16.284 $\pm$ 0.692 & 24.058 $\pm$ 8.329 & \lightgrey 16.959 $\pm$ 1.715 \\
    \bottomrule
  \end{tabular}
  }
\vspace{1ex}
\caption{
\textbf{RTM\_AE: Noise robustness analysis.} 
Model performance under different noise intensities (0.01, 0.03, 0.05). 
Trait-wise $R^2$ (higher is better) and nRMSE (lower is better) are reported as mean $\pm$ standard deviation.
}
\label{tab:aertm_noise_results}
\vspace{-3mm}
\end{table}

\begin{table}[t!]
  \centering
  \resizebox{\textwidth}{!}{%
  \begin{tabular}{lccccccccc}
    \toprule
    Noise & cab & cw & cm & LAI & cp & cbc & car & anth & \lightgrey Average \\
    \midrule
    \multicolumn{10}{c}{$R^2$ $(\uparrow)$} \\
    \midrule
    0.01 & 0.518 $\pm$ 0.003 & 0.432 $\pm$ 0.005 & 0.587 $\pm$ 0.005 & 0.416 $\pm$ 0.006 & 0.438 $\pm$ 0.009 & 0.591 $\pm$ 0.006 & 0.443 $\pm$ 0.008 & 0.170 $\pm$ 0.017 & \lightgrey 0.449 $\pm$ 0.007 \\
    0.03 & 0.153 $\pm$ 0.029 & 0.357 $\pm$ 0.026 & 0.486 $\pm$ 0.004 & 0.395 $\pm$ 0.011 & 0.253 $\pm$ 0.028 & 0.476 $\pm$ 0.017 & 0.250 $\pm$ 0.009 & 0.100 $\pm$ 0.025 & \lightgrey 0.309 $\pm$ 0.019 \\
    0.05 & -0.677 $\pm$ 0.039 & 0.084 $\pm$ 0.066 & 0.387 $\pm$ 0.002 & 0.359 $\pm$ 0.018 & 0.043 $\pm$ 0.031 & 0.372 $\pm$ 0.018 & -0.215 $\pm$ 0.008 & 0.001 $\pm$ 0.030 & \lightgrey 0.044 $\pm$ 0.027 \\
    \midrule
    \multicolumn{10}{c}{nRMSE $(\downarrow)$} \\
    \midrule
    0.01 & 16.858 $\pm$ 0.048 & 14.712 $\pm$ 0.058 & 12.747 $\pm$ 0.069 & 18.752 $\pm$ 0.092 & 15.937 $\pm$ 0.133 & 12.861 $\pm$ 0.094 & 15.070 $\pm$ 0.107 & 20.346 $\pm$ 0.203 & \lightgrey 15.910 $\pm$ 0.101 \\
    0.03 & 22.348 $\pm$ 0.391 & 15.658 $\pm$ 0.318 & 14.222 $\pm$ 0.053 & 19.085 $\pm$ 0.182 & 18.367 $\pm$ 0.349 & 14.551 $\pm$ 0.230 & 17.494 $\pm$ 0.107 & 21.187 $\pm$ 0.292 & \lightgrey 17.864 $\pm$ 0.240 \\
    0.05 & 31.452 $\pm$ 0.367 & 18.672 $\pm$ 0.675 & 15.524 $\pm$ 0.029 & 19.645 $\pm$ 0.280 & 20.783 $\pm$ 0.352 & 15.928 $\pm$ 0.236 & 22.266 $\pm$ 0.076 & 22.317 $\pm$ 0.338 & \lightgrey 20.823 $\pm$ 0.294 \\
    \bottomrule
  \end{tabular}
  }
\vspace{1ex}
\caption{
\textbf{MAE\_FR\_LP: Noise robustness analysis.} 
Model performance under different noise intensities (0.01, 0.03, 0.05). 
Trait-wise $R^2$ (higher is better) and nRMSE (lower is better) are reported as mean $\pm$ standard deviation.
}
\label{tab:mae_lp_noise_results}
\vspace{-3mm}
\end{table}

\begin{table}[t!]
  \centering
  \resizebox{\textwidth}{!}{%
  \begin{tabular}{lccccccccc}
    \toprule
    Noise & cab & cw & cm & LAI & cp & cbc & car & anth & \lightgrey Average \\
    \midrule
    \multicolumn{10}{c}{$R^2$ $(\uparrow)$} \\
    \midrule
    0.01 & 0.583 $\pm$ 0.012 & 0.645 $\pm$ 0.024 & 0.787 $\pm$ 0.006 & 0.648 $\pm$ 0.014 & 0.667 $\pm$ 0.026 & 0.781 $\pm$ 0.007 & 0.584 $\pm$ 0.037 & 0.447 $\pm$ 0.031 & \lightgrey 0.643 $\pm$ 0.020 \\
    0.03 & 0.440 $\pm$ 0.028 & 0.536 $\pm$ 0.007 & 0.659 $\pm$ 0.009 & 0.530 $\pm$ 0.021 & 0.496 $\pm$ 0.035 & 0.647 $\pm$ 0.012 & 0.461 $\pm$ 0.029 & 0.264 $\pm$ 0.029 & \lightgrey 0.504 $\pm$ 0.021 \\
    0.05 & 0.242 $\pm$ 0.045 & 0.388 $\pm$ 0.023 & 0.477 $\pm$ 0.008 & 0.411 $\pm$ 0.014 & 0.276 $\pm$ 0.060 & 0.460 $\pm$ 0.009 & 0.258 $\pm$ 0.027 & 0.136 $\pm$ 0.073 & \lightgrey 0.331 $\pm$ 0.032 \\
    \midrule
    \multicolumn{10}{c}{nRMSE $(\downarrow)$} \\
    \midrule
    0.01 & 15.677 $\pm$ 0.227 & 11.628 $\pm$ 0.388 & 9.141 $\pm$ 0.121 & 14.564 $\pm$ 0.290 & 12.265 $\pm$ 0.478 & 9.401 $\pm$ 0.147 & 13.025 $\pm$ 0.575 & 16.594 $\pm$ 0.462 & \lightgrey 12.787 $\pm$ 0.336 \\
    0.03 & 18.163 $\pm$ 0.461 & 13.302 $\pm$ 0.099 & 11.571 $\pm$ 0.146 & 16.823 $\pm$ 0.378 & 15.083 $\pm$ 0.526 & 11.947 $\pm$ 0.200 & 14.830 $\pm$ 0.406 & 19.154 $\pm$ 0.380 & \lightgrey 15.109 $\pm$ 0.325 \\
    0.05 & 21.131 $\pm$ 0.621 & 15.270 $\pm$ 0.291 & 14.346 $\pm$ 0.105 & 18.836 $\pm$ 0.219 & 18.078 $\pm$ 0.748 & 14.772 $\pm$ 0.123 & 17.400 $\pm$ 0.320 & 20.738 $\pm$ 0.879 & \lightgrey 17.571 $\pm$ 0.413 \\
    \bottomrule
  \end{tabular}
  }
\vspace{1ex}
\caption{
\textbf{MAE\_FR\_FT: Noise robustness analysis.} 
Model performance under different noise intensities (0.01, 0.03, 0.05). 
Trait-wise $R^2$ (higher is better) and nRMSE (lower is better) are reported as mean $\pm$ standard deviation.
}
\label{tab:mae_ft_noise_results}
\vspace{-3mm}
\end{table}

\clearpage
\begin{figure}[H]
  \centering
  \includegraphics[width=\linewidth, height=0.75\textheight, keepaspectratio]{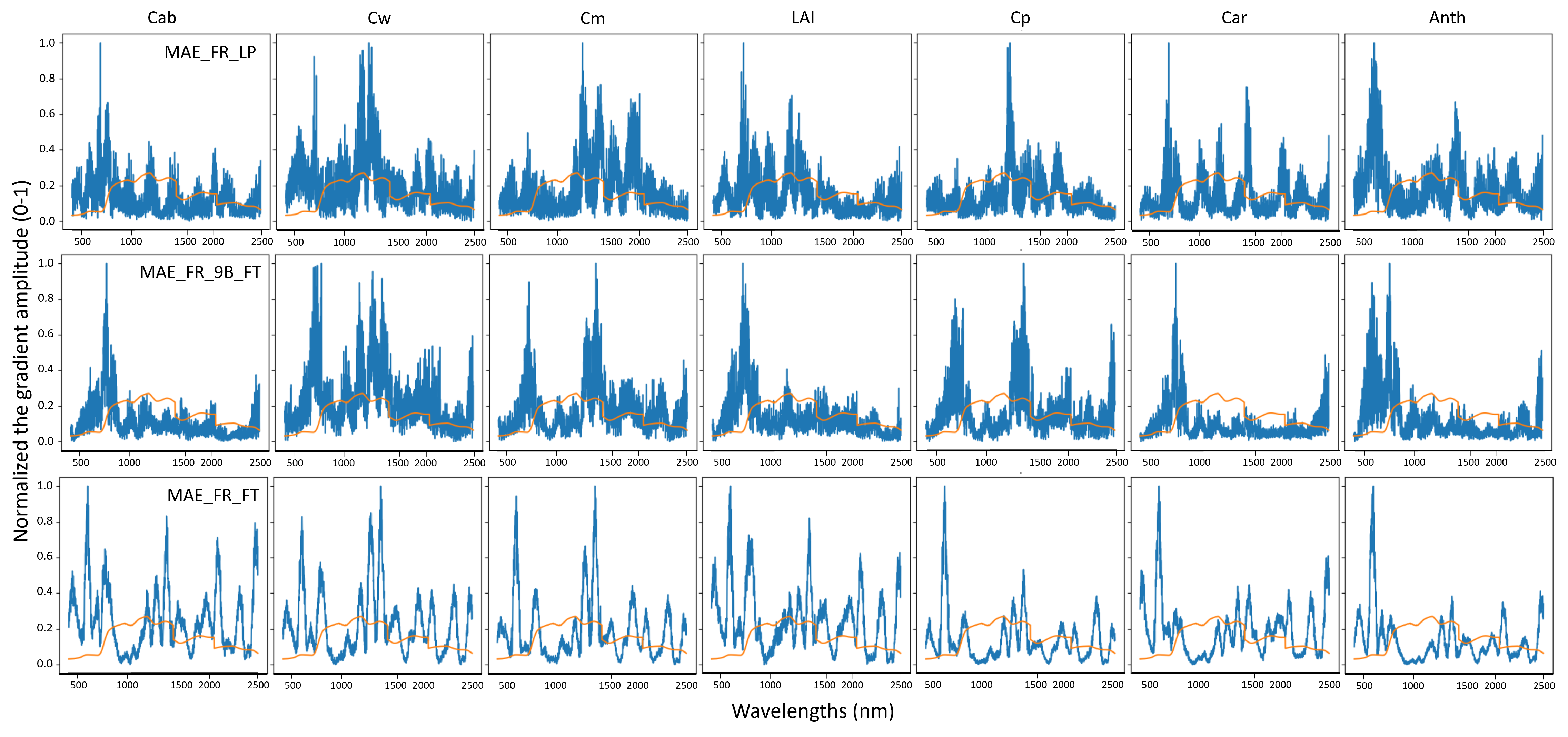}%

  \vspace{1ex} 
  
  \caption{
    \textbf{Feature importance of MAE-based downstream regression.} Results are shown for (top) linear probing ($MAE\_FR\_LP$), (middle) fine-tuning the last block ($MAE\_FR\_9B\_FT$), and (bottom) full fine-tuning ($MAE\_FR\_FT$). The blue lines indicate the importance scores across spectral bands, while the orange line shows a reference of a vegetation spectra.
  }

  \label{fig:featureXai_MAE}
\end{figure}

\end{document}